\pdfoutput=1
\documentclass{article} 

\usepackage{merlin_preprint,times}
\usepackage{colortbl}
\definecolor{t1fill}{RGB}{226,231,247}
\definecolor{t3fill}{RGB}{244,222,231}
\definecolor{t1}{RGB}{103,119,163}
\definecolor{t2}{RGB}{130,110,142}
\definecolor{t3}{RGB}{157,101,121}

\usepackage{booktabs}

\usepackage{amsmath,amsfonts,bm}

\def\eqref#1{equation~\ref{#1}}

\def\1{\bm{1}}

\DeclareMathAlphabet{\mathsfit}{\encodingdefault}{\sfdefault}{m}{sl}
\SetMathAlphabet{\mathsfit}{bold}{\encodingdefault}{\sfdefault}{bx}{n}

\usepackage{amssymb}
\usepackage{longtable,array,tabularx}
\usepackage{float,needspace}
\usepackage{xcolor,xurl,fvextra}
\usepackage{tikz}
\usetikzlibrary{arrows.meta,calc,fit,positioning,shapes.geometric}
\newtheorem{lemma}{Lemma}[section]
\newcommand{\iso}{\cong_{\mathrm{col}}}
\newcommand{\F}{\mathbb F}

\newcommand{\code}[1]{\texttt{\detokenize{#1}}}
\newcounter{prompt}[section]
\renewcommand{\theprompt}{\thesection.\arabic{prompt}}
\newcommand{\promptfile}[3]{%
  \par\Needspace{5\baselineskip}%
  \refstepcounter{prompt}\label{#1}%
  \noindent\textbf{Prompt \theprompt. #2}\par\smallskip
  \VerbatimInput[fontsize=\footnotesize,breaklines=true,breakanywhere=true,
    breakautoindent=true,breaksymbolleft={},breaksymbolright={},
    xleftmargin=1em,xrightmargin=1em,tabsize=2]{#3}\par}
\newcommand{\shortpromptfile}[3]{%
  \par\noindent\begin{minipage}{\linewidth}
  \promptfile{#1}{#2}{#3}
  \end{minipage}\par\smallskip}
\tikzset{
  graph vertex/.style={circle,draw,fill=white,inner sep=0pt,minimum size=4.8pt},
  outer vertex/.style={circle,draw,fill=white,inner sep=0pt,minimum size=5.8pt},
  middle vertex/.style={circle,draw,fill=gray!25,inner sep=0pt,minimum size=6.2pt},
  sum variable/.style={circle,draw,fill=white,inner sep=0pt,minimum size=6mm,font=\scriptsize},
  sum factor/.style={rectangle,draw,fill=gray!15,rounded corners=1pt,minimum height=5.5mm,minimum width=8mm,font=\scriptsize},
  circuit gate/.style={rectangle,draw,fill=gray!12,minimum height=5.5mm,minimum width=10mm,font=\scriptsize},
  diagram arrow/.style={-{Latex[length=2mm]},thin},
  diagram label/.style={font=\small},diagram edge/.style={thin}}

\usepackage{tcolorbox}
\tcbuselibrary{skins,breakable,listings}

\definecolor{promptPaper}{RGB}{251,245,227}
\definecolor{promptBorder}{RGB}{190,150,84}
\definecolor{promptInk}{RGB}{45,48,56}
\definecolor{promptCode}{RGB}{83,104,151}
\definecolor{promptVariable}{RGB}{130,94,139}
\definecolor{promptStatus}{RGB}{153,86,106}

\lstdefinestyle{MerlinPrompt}{
  language={},
  basicstyle=\fontencoding{T1}\fontfamily{lmtt}\fontsize{8}{9.4}\selectfont\color{promptInk},
  columns=fullflexible,keepspaces=true,
  showspaces=false,showtabs=false,showstringspaces=false,
  upquote=true,
  tabsize=2,breaklines=true,breakatwhitespace=false,
  breakautoindent=true,breakindent=0pt,
  aboveskip=0pt,belowskip=0pt,
  numbers=none,frame=none,
  mathescape=false,texcl=false,
  literate={-}{{\char45}}1 {*}{{\char42}}1,
  alsoletter={_},
  morecomment=[l][\color{promptCode}\bfseries]{\#},
  moredelim=**[s][\color{promptCode}]{`}{`},
  moredelim=**[s][\color{promptVariable}\bfseries]{\{\{}{\}\}},
  emph={ACCEPT,REJECT,TIMEOUT,BUDGET_EXCEEDED,READY_FOR_EVALUATION},
  emphstyle=\color{promptStatus}\bfseries
}
\tcbset{
  MerlinPromptBox/.style={
    enhanced,breakable,
    colback=promptPaper!55,colframe=promptBorder!40,
    colbacktitle=promptPaper,coltitle=promptInk,
    fonttitle=\small\bfseries,
    boxrule=.35pt,arc=1mm,
    boxsep=0pt,left=5pt,right=5pt,top=4pt,bottom=4pt,
    toptitle=3pt,bottomtitle=3pt,
    before skip=6pt,after skip=6pt,
    pad at break*=2pt,
    listing only,listing options={style=MerlinPrompt}
  }
}
\renewcommand{\promptfile}[3]{%
  \par\Needspace{5\baselineskip}%
  \refstepcounter{prompt}\label{#1}%
  \tcbinputlisting{MerlinPromptBox,
    title={Prompt \theprompt. #2},
    title after break={Prompt \theprompt. #2\ \textnormal{(continued)}},
    listing file={#3}}%
}
\renewcommand{\shortpromptfile}[3]{%
  \par\addvspace{6pt}\noindent
  \begin{minipage}{\linewidth}%
  \tcbset{MerlinPromptBox/.append style={
    unbreakable,before skip=0pt,after skip=0pt}}%
  \promptfile{#1}{#2}{#3}%
  \end{minipage}\par\addvspace{6pt}%
}

\newsavebox{\MBfindingcontent}
\newcounter{MBfindingitem}
\definecolor{findingwarmbg}{RGB}{251,245,227}
\definecolor{findingwarmedge}{RGB}{190,150,84}
\tikzset{
  finding/bluebar/.style={fill=t1!7,draw=none},
  finding/lavender/.style={fill=t2!6,draw=t2!65,line width=.35pt,
    rounded corners=1pt},
  finding/roseband/.style={fill=t3!6,draw=none},
  finding/warm/.style={fill=findingwarmbg!55,rounded corners=1.5pt}
}

\usepackage{hyperref}
\hypersetup{
    pdftitle={Interactive Proofs as a Unified Framework for Evaluating and Leveraging Verifiable Proof Generation by LLM Agents},
    pdfauthor={Yankai Shu, Xingyu Zhao, Chuan Mao, Congchi Yin},
    colorlinks=true,
    citecolor=t1, 
    linkcolor=t3, 
    urlcolor=black   
}
\usepackage{url}

\usepackage{graphicx}
\usepackage{wrapfig}

\title{Interactive Proofs as a Unified Framework for Evaluating and Leveraging Verifiable Proof Generation by LLM Agents}

\author{%
\textbf{Yankai Shu}\textsuperscript{1,*,$\dagger$}\quad
\textbf{Xingyu Zhao}\textsuperscript{1,*,$\dagger$}\quad
\textbf{Chuan Mao}\textsuperscript{1,*}\quad
\textbf{Congchi Yin}\textsuperscript{2}
}

\newsavebox{\MBcomparisonbox}
\begin{document}

\maketitle
\authornotes{%
\textsuperscript{1}School of Electronics Engineering and Computer Science, Peking University.\par
\textsuperscript{2}College of Artificial Intelligence, Nanjing University of Aeronautics and Astronautics.\par
\textsuperscript{*}Equal contribution.\par
\textsuperscript{$\dagger$}\mbox{Correspondence to:
Yankai Shu \textless\href{mailto:syksykccc@stu.pku.edu.cn}{syksykccc@stu.pku.edu.cn}\textgreater,
Xingyu Zhao \textless\href{mailto:stellary@stu.pku.edu.cn}{stellary@stu.pku.edu.cn}\textgreater.}}

\begin{abstract}
Recent advances in LLM-assisted theorem proving and program verification have attracted widespread attention. Beyond determining whether a proposition is true, LLM agents must produce evidence that can be independently checked by humans or formal verification tools. To evaluate the underlying, protocol-agnostic ability of LLM agents to construct verifiable proofs—rather than their proficiency with any particular proof language or verification tool—we introduce \(\textsc{Merlin-Bench}\). Drawing on interactive proof systems from computational complexity theory, we cast an LLM agent as the prover: under a designated protocol, the agent interacts with a formal verifier to substantiate its claimed answer to a problem instance. We develop an automated framework for constructing test cases and use it to create 100 cases across 15 problem families, each formulated within this prover-verifier framework. Experiments with 10 frontier LLMs across multiple agent harnesses show that reliably producing verifiable proofs in accordance with specified protocols remains a substantial challenge. Our case studies further provide actionable insights into improving agents’ ability to generate verifier-accepted proofs: tool use is essential for completing practical proof tasks; different forms of guidance, ranging from engineering instructions to problem-specific knowledge, substantially affect success rates on interactive-proof tasks; and proof-solving experience distilled from abstract tasks can transfer to concrete, real-world verification tasks.
\end{abstract}

\section{Introduction} \label{sec:introduction}

Recently, LLM agents have demonstrated strong capabilities in theorem proving. A new paradigm of human–AI collaboration has emerged in which LLM agents search for proofs or counter-examples and the resulting outputs are subsequently verified by human experts or formal verification tools. For example, in frontier mathematics, LLM agents discover proofs and produce proof texts (\cite{openai2026planar, openai2026finite}), structured lemmas (\cite{ke2026towards}) or formalizations in Lean, which are then checked by human experts (\cite{alon2026remarksdisproofunitdistance}) or verified by Lean (\cite{openai2026navierstokeslean}). In software verification, LLM agents discover annotations such as loop invariants (\cite{yang2025autoverus, mugnier2025laurel}), whose correctness is then verified by tools such as Dafny (\cite{leino2010dafny}). Overall, this emerging paradigm requires LLM agents not only to find evidence, but also to complete proofs through interacting with a specified framework. Much of the prior work has focused on various specific tasks. In this paper, we seek to characterize:

\textit{How can we develop a high-level, abstract model of this underlying \textbf{unified proving capability}: an LLM agent proposes a claim and, following certain mutually agreed-upon protocol, conveys a message rigorously to a human?}

A large body of prior work has investigated the ability of LLMs to construct certificates and prove propositions. For example, NLGraph and NPEval (\cite{fan-etal-2024-nphardeval, wang2023graph}) ask LLMs to solve graph-theoretic problems and small-scale NP-hard problems, respectively. However, these benchmarks are not \textit{agent-oriented}, because an agent can solve such problems simply by writing code. SATBench (\cite{wei-etal-2025-satbench}) evaluates LLMs on automatically generated SAT instances. However, it uses  another LLM to audit the output of the LLM being benchmarked; therefore, the \emph{certificate} is not \emph{rigorous}. MiniF2F (\cite{zheng2022minif2f}) evaluates LLMs' capability of generating formal proofs, with questions drawn from Olympiad mathematics, making it difficult to \textit{scale up}. Verus-Bench and DafnyBench (\cite{yang2025autoverus, mugnier2025laurel}) require LLMs to complete tasks using specific protocol (in particular the formal verification tool). Consequently, the capabilities they evaluate are closely tied to specific tools rather than reflecting an \textit{unified proof ability} at a general, abstract level. We detail the specification of the four merits mentioned above in Appendix \ref{app:merits}. 

\begin{table}[htbp]
    \centering
    \def\MBpairscale{1.15}

    \sbox{\MBcomparisonbox}{%
        \scalebox{\MBpairscale}{%
            \begin{minipage}{64mm}
\begingroup
\fontsize{6.8}{7.5}\selectfont
\color[rgb]{0.149,0.192,0.243}
\setlength{\tabcolsep}{1.1pt}
\setlength{\aboverulesep}{1.2pt}
\setlength{\belowrulesep}{1.2pt}
\renewcommand{\arraystretch}{0.96}
\newcommand{\MBAY}{%
  \textcolor[rgb]{0.380,0.490,0.431}{%
    {\fontsize{7.5}{7.5}\usefont{U}{pzd}{m}{n}\char51}}}
\newcommand{\MBAN}{%
  \textcolor[rgb]{0.616,0.396,0.475}{%
    {\fontsize{7.5}{7.5}\usefont{U}{pzd}{m}{n}\char55}}}
\newcommand{\MBAhead}[1]{%
  {\fontsize{6}{6.5}\selectfont\bfseries\begin{tabular}[c]{@{}c@{}}#1\end{tabular}}}
\noindent\begin{tabular*}{64mm}{
  @{\extracolsep{\fill}}l
  *{4}{>{\centering\arraybackslash}p{10.2mm}}@{}
}
  \toprule[0.45pt]
  \MBAhead{Benchmark}
  & \MBAhead{Agent-\\oriented}
  & \MBAhead{Rigorous\\certificate}
  & \MBAhead{Scalable\\synthesis}
  & \MBAhead{Unified\\proof}\\
  \midrule[0.28pt]
  NLGraph     & \MBAN & \MBAN & \MBAY & \MBAN\\
  NPHardEval  & \MBAN & \MBAN & \MBAY & \MBAY\\
  SATBench    & \MBAN & \MBAN & \MBAY & \MBAY\\
  miniF2F     & \MBAN & \MBAY & \MBAN & \MBAN\\
  Verus-Bench & \MBAY & \MBAY & \MBAN & \MBAN\\
  DafnyGym    & \MBAY & \MBAY & \MBAN & \MBAN\\
  \midrule[0.28pt]
  \textbf{Merlin-Bench} & \MBAY & \MBAY & \MBAY & \MBAY\\
  \bottomrule[0.45pt]
\end{tabular*}
\endgroup

            \end{minipage}%
        }%
    }

    \begin{minipage}[t]{0.49\linewidth}
        \vspace{0pt}
        \centering
        \raisebox{\depth}{\usebox{\MBcomparisonbox}}
        \captionsetup{skip=3pt}
        \captionof{table}{Comparison with related benchmarks.}
        \label{tab:benchmark-comparison}
    \end{minipage}\hfill
    \begin{minipage}[t]{0.49\linewidth}
        \vspace{0pt}
        \centering
        \includegraphics[
            height=\dimexpr
                \ht\MBcomparisonbox+\dp\MBcomparisonbox
            \relax,
            width=\linewidth,
            keepaspectratio
        ]{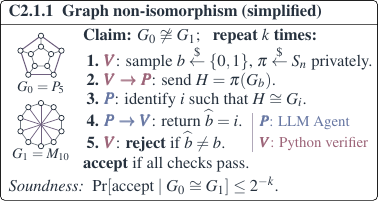}
        \captionsetup{skip=3pt}
        \captionof{figure}{Example of Test Case 2.1.1}
        \label{fig:c211-example}
    \end{minipage}
\end{table}










This paper introduces \(\textsc{Merlin-Bench}\), addressing all four limitations of prior work described above. Regardless of the formal verification tool employed or the proposition being proved, these proof tasks fall within the framework of interactive proof systems from computational complexity theory \cite{Babai1985}. Accordingly, in each test case of \(\textsc{Merlin-Bench}\), the LLM agent is required to complete the following task: given a instance, which is dynamically generated by a generator $\mathcal{G}$, of an decision problem equipped with an interactive protocol \(\Pi\) and a formal verifier \(V\), an LLM agent is supposed to complete an interactive proof under protocol $\Pi$ with \(V\), as illustrated in Figure \ref{fig:c211-example}. On the one hand, this task formulation naturally aligns with the emerging paradigm of human–AI collaboration in theorem proving and formal verification. On the other hand, it raises the problem complexity to \(\mathsf{IP}=\mathsf{PSPACE}\) (\cite{Shamir1992IP}), thereby preserving substantial headroom in difficulty (see Appendix~\ref{app:extensibility}), even for agents equipped with tool-use and code-generation capabilities.

We construct 100 test cases from 15 \(\mathsf{IP}\) problems, requiring models to complete interactive proofs according to specified protocols. To construct these test cases efficiently while ensuring their diversity, feasibility, and difficulty, we develop a test-case construction framework: a construction agent automatically executes an iterative cycle of literature review, generator implementation, and feasibility and difficulty checks to generate challenging but tractable test cases at scale. The correctness of all 100 test cases generated in this work has been verified by human reviewers, demonstrating the effectiveness of this framework. On this benchmark, we evaluate 10 frontier LLMs from major model families. The results show that, beyond the ability to find answers, completing proofs according to specified requirements remains a significant bottleneck for current LLMs. We further conduct several case studies and identify promising directions for improving models’ proof capabilities. In summary, our main technical contributions are:
\begin{enumerate}
    \item \textbf{A new evaluation paradigm based on $\mathsf{IP}$.} We propose an evaluation paradigm based on interactive proof systems that is agent-oriented, evaluates models’ unified proof capabilities through strict formal verification, and can be scaled up.
    \item \textbf{A benchmark and a test case generation framework.} We construct \textsc{Merlin-Bench}, which includes 100 test cases spanning 15 \(\mathsf{IP}\) problems to dynamically generate evaluation instances to evaluate LLM agents’ proof capabilities. Additionally, we propose a agentic framework that reliably generates practical but discriminative test cases at scale.
    \item \textbf{Empirical observations on improving models’ proof capabilities.} We find that tool use and user prompts at three different levels have significant effects on models’ proof capabilities. We further show that proof experience acquired from the abstract \(\mathsf{IP}\) problems in our benchmark can improve performance on real-world formal verification tasks.
\end{enumerate}

For reproducible code and datasets for \textsc{Merlin-Bench}, see \url{https://github.com/syksykCCC/Merlin-Bench}.

\section{Preliminaries}

This section introduces the theoretical foundations of computation underlying \textsc{Merlin-Bench}. The problems in \textsc{Merlin-Bench} are derived from decision problems in computational theory. Intuitively, a decision problem asks whether an input \(x\) satisfies a particular property. By encoding each input as a binary string, any decision problem can be characterized by a subset \(L \subseteq \{0,1\}^*\), where \(\{0,1\}^*\) denotes the set of all finite-length binary strings and \(x\in L\) if and only if it encodes an input which satisfies this property.

\paragraph{Nondeterministic Polynomial ($\textsf{NP}$) Problems.} For any decision problem \(L\), we say that \(L\) is solvable in nondeterministic polynomial time (or write \(L \in \mathsf{NP}\)) if and only if there exists a polynomial-time algorithm \(\mathcal{V}\) and a polynomial $q(x)$ such that
$$
x\in L \quad \Leftrightarrow \quad \exists w : |w| \leq q(|x|) \wedge \mathcal{V}(x, w)\rightarrow 1,
$$
where \(\mathcal{V}(x,w)\rightarrow 1\) means that the algorithm \(\mathcal{V}\) takes \(x\) and \(w\) as input and outputs \(1\). \(w\), the second input to the algorithm \(\mathcal{V}\), is called the \emph{certificate} for this problem.


\paragraph{Interactive Proof (\textsf{IP}) Problems.} Some decision problems can be verified by interactive proof systems. Intuitively, an interactive proof is carried out jointly by two algorithms, called the prover \(P\) and the verifier \(V\). The prover \(P\) is computationally unbounded, whereas the verifier \(V\) is restricted to polynomial-time randomized computation. The prover and the verifier communicate for a polynomial number of rounds, after which the verifier decides whether to accept or reject the prover’s proof based on the communication transcript. Formally, a multi-round interactive protocol is defined as follows:

\begin{definition}[Interactive proof protocol]
A multi-round interactive protocol \(\Pi\) consists of the following data: the number of interaction rounds \(n\), the format of messages, and the speaking order \(r\in\{\mathsf{P},\mathsf{V}\}^n\), where \(n\) is polynomial in the input length \(|x|\).
\end{definition}

The interaction between the prover and the verifier under the interactive protocol \(\Pi\) is formally defined as follows. Let \(h_0=\varepsilon\) denote the empty sequence, and let \(h_i=(m_1,\ldots,m_i)\) denote the prefix of the interaction transcript between the two parties, and
$$
m_i = \begin{cases}
    P(x, h_{i - 1}), & r_i = \textsf{P}, \\
    V(x, h_{i - 1}), & r_i = \textsf{V},
\end{cases}, \quad D_\Pi(P, x) = V_{\mathrm{out}}(x, h_n).
$$

A decision problem is said to admit an interactive proof (or belong to \(\textsf{IP}\)) if and only if there exists a interactive proof protocol $\Pi$ and a verifier $V$ such that the following two properties hold:

\begin{enumerate}
    \item \emph{Completeness}. There exists a sufficiently powerful prover that, for every valid input, can produce a proof that causes the verifier to accept with high probability:
    \[
    \exists P,\ \forall x\in L,\quad \Pr[D_\Pi(P,x)\rightarrow 1]\geq 1-\varepsilon_c.
    \]
    \item \emph{Soundness}.  No prover can convince the verifier to accept an invalid input with high probability:
    \[
    \forall P,\ \forall x\notin L,\quad \Pr[D_\Pi(P,x)\rightarrow 1]\leq \varepsilon_s.
    \]
\end{enumerate}

where $\varepsilon_s, \varepsilon_c$ are constants with $\varepsilon_s, \varepsilon_c\leq 1/3$. We refer to \(P\) and \(V\) as \textit{Merlin} and \textit{Arthur}, respectively. Note that in this paper, Merlin’s input does not include Arthur’s random seed. 


\begin{example}[Graph Nonisomorphism, GNI]
GNI asks whether two graphs $G_0, G_1$ satisfy $G_0\not\cong G_1$. Its membership in $\mathsf{NP}$ remains open, as no general efficiently checkable certificate is known.
However, we can construct an $\mathsf{IP}$ protocol (as shown in Figure~\ref{fig:c211-example}): asking $P$ to identify the source of a randomly relabeled input graph. An unbounded $P$ can always identify the source when $G_0\not\cong G_1$, but can only guess when $G_0\cong G_1$.
Independent repetition gives arbitrarily small soundness error.
\end{example}

\begin{lemma}
[$\mathsf{NP}\subseteq\mathsf{IP}$]
$\mathsf{NP}$ can be viewed as a `single round' interactive proof: 
$P$ sends the certificate $w$ to $V$, and $V$ executes $\mathcal V(x, w)$ in polynomial time and decides to accept or reject.
\end{lemma}


\section{Merlin-Bench}

\begin{figure}[htbp]
    \centering
    \def\cropTop{1.5mm}
    \def\cropBottom{2mm}

    \includegraphics[
        width=\linewidth,
        trim={0mm \cropBottom{} 0mm \cropTop{}},
        clip
    ]{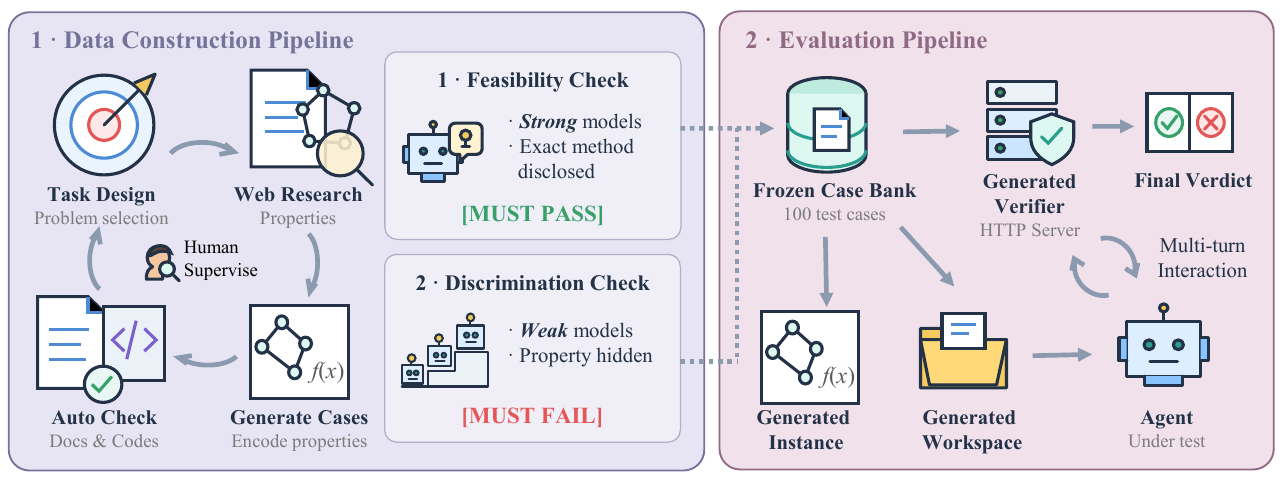}
    
    \caption{Overview of the \textsc{Merlin-Bench} construction and evaluation pipeline. (Left) Test cases are developed through problem selection, literature research, generator implementation, and automatic checks under human supervision, followed by feasibility and discrimination checks. (Right) The frozen case bank generates an instance, an agent workspace, and a verifier service for each evaluation. The agent interacts with the verifier according to the protocol to obtain a final verdict.}
    \label{fig:pipeline}
\end{figure}

In this section, we introduce the key design in \textsc{Merlin-Bench}. The complete \textsc{Merlin-Bench} pipeline, including test set construction and evaluation, is illustrated in Figure \ref{fig:pipeline}.

\subsection{Basic Settings}

\paragraph{Formulation of Test Cases.} The basic unit of \textsc{Merlin-Bench} is a \textit{test case}. Each test case is associated with a decision problem in \(\textsf{IP}\) and specifies a rule for constructing problem instances, together with constraints on the resources available for carrying out the interactive proof. Formally, a test case associated with $L\in \textsf{IP}$ is formulated by a $5$-tuple
$c = (D_c, \mathcal{G}_c, \Pi_c, \mathcal{V}_c, B_c)$.
Here, the \emph{description} $D_c$ is a string describing problem $L$, the \emph{generator} \(\mathcal{G}_c\) is a randomized algorithm that takes a random seed \(r \sim \mathcal{R}\) as input and randomly constructs an instance \(z\in L\) according to a manually specified rule; different random seeds can be viewed as equivalent transformations of the problem to prevent data leakage, having nearly no impact on the instance's difficulty; the \emph{interactive proof protocol} \(\Pi_c\) with completeness factor $\varepsilon_c \le 10^{-10}$ and soundness $\varepsilon_s \leq 2^{-23}$ (See Appendix~\ref{app:perfect-completeness}), together with the \emph{verifier} \(\mathcal{V}_c\), constitutes a complete and sound interactive proof system for \(L\); the \emph{resource constraints} \(B_c\) include limits on computation time, permitted tools, and other resources. The score of the agent \(A_\theta\) on test case $c$ is defined as follows:
$$
\mathrm{Score}_c(A_\theta) = \mathbb{I}\left[\mathcal{V}_c(z, h_n)\rightarrow 1 \wedge \mathrm{Valid}_{B_c}(\tau) \mid r\leftarrow \mathcal{R}, z\leftarrow \mathcal{G}_c(r)\right],
$$
where $h_n$ denotes the interaction history of $\mathcal{V}$ and $A_\theta$ under protocol $\Pi$, and \(\tau\) denotes the agent’s complete execution trajectory, including its tool-call history and interactions under the protocol.

\paragraph{Evaluation Procedure.} The entire evaluation process is managed by a controller program called the \emph{evaluator}, as described below. For a given test case \(c\), the evaluator samples a random seed \(r\) and invokes \(\mathcal{G}_c(r)\) to generate an instance \(z\). The evaluator then creates a workspace \(W_c\) for the agent under evaluation, \(A_\theta\), containing the instance \(z\), the description $D_c$ of the decision problem \(L\), and the message format specified by \(\Pi_c\). In addition, the evaluator launches a local HTTP server whose backend runs the verifier \(\mathcal{V}_c\), and the agent interacts with it through a client.

If the verifier ultimately outputs \(1\), the evaluation result is recorded as a \emph{pass}. The evaluator also retains the complete execution trajectory of \(A_\theta\) and checks whether the agent’s runtime and resource usage satisfy the requirements specified by \(B_c\). If all checks pass, the evaluation result is recorded as a \emph{strict pass}. Finally, the agent’s score is calculated based on its strict-pass rate.

\subsection{Dataset Construction}

\textsc{Merlin-Bench} contains 100 test cases spanning 15 problems. This section describes the hierarchical organization of these test cases, as well as the methods used to collect and filter them.

\paragraph{Problem Tier Classification.} To evaluate models more precisely, we classify the problems into three tiers rather than treating them all uniformly, as described below.

\begin{description}
    \item[\color{t1} Tier-1] \emph{Deterministic certificate.} This tier contains 36 test cases covering six \(\mathsf{NP}\) problems. After searching for an answer and a certificate, the agent engages in a single round of interaction with the verifier to submit the certificate, which the verifier only needs to check for validity.
    \item[\color{t2} Tier-2] \emph{Multi-round interactions.} This tier contains \(40\) test cases covering six \(\mathsf{IP}\) problems. In this tier, the agent must convince Arthur, through a protocol-compliant multi-round interaction, that the input is indeed an instance satisfying the required property. Some test cases in this tier are fundamentally more difficult than those in Tier 1: certain problems are not currently known to belong to \(\mathsf{NP}\), and the agent must provide a consistent interactive proof without relying on a fixed certificate.
    \item[\color{t3} Tier-3] \emph{Real-world tasks.} This tier uses three real-world protocols, Lean, Dafny, and DRAT, as verifiers, with eight test cases for each. Through this tier, we demonstrate that an agent’s ability to perform these three types of formal verification is fundamentally aligned with the capabilities evaluated by \(\textsc{Merlin-Bench}\).
\end{description}

\begin{figure}[htbp]
    \centering
    \def\cropTop{2mm}
    \def\cropBottom{2mm}

    \includegraphics[
        width=\linewidth,
        trim={0mm \cropBottom{} 0mm \cropTop{}},
        clip
    ]{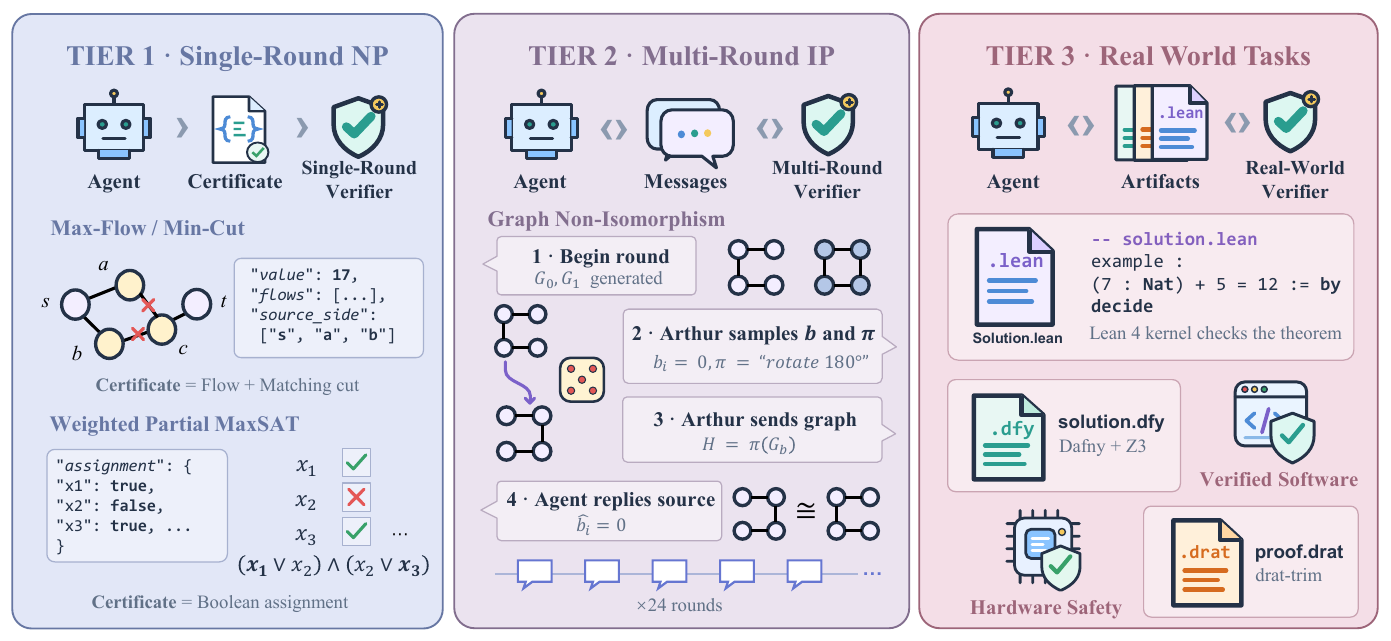}
    \caption{Representative tasks and verification processes across the three tiers of \textsc{Merlin-Bench}. (Left) Tier 1 is deterministic single round verification. (Middle) Tier 2 is probabilistic multi-round verification. (Right) Tier 3 gives a demonstration of this framework in real-world tasks.}
    \label{fig:tasks-tier}
\end{figure}

Figure \ref{fig:tasks-tier} presents representative problems from the three tiers to illustrate their interaction processes. The complete list and detailed descriptions are provided in Appendices~\ref{app:tier1}--\ref{app:tier3}.

\paragraph{Test Case Construction.} We developed a construction agent to generate test cases at scale. Its workflow is described below. Details on the construction process are provided in Appendix~\ref{app:construction-evaluation}.

\textit{Literature review.} Given a decision problem \(L\), the construction agent conducts a literature review to identify structural restrictions that admit specialized solutions. For example, general \text{SAT} is \textsf{NP-hard}, whereas \text{2-SAT} can be solved in linear time. The identified structure is then used to construct the generator \(\mathcal{G}_c\) and the reference algorithm \(\mathcal{A}_c\). This structure and both algorithms are subsequently hidden from the agent under evaluation.

\textit{Generator implementation and automatic check.} Based on the problem and the structure identified in the previous step, the construction agent implements the generator \(\mathcal{G}_c\), the reference algorithm \(\mathcal{A}_c\), and the verifier \(\mathcal{V}_c\). The construction agent then writes the description \(D_c\), ensuring that its specifications of the interactive protocol and input format are consistent with \(\Pi_c\) and \(\mathcal{V}_c\). Next, the construction agent automatically checks the correctness of the test cases. It runs the reference algorithm \(\mathcal{A}_c\) to ensure that it is accepted and submits some invalid proofs to check that they are rejected.

\textit{Feasibility and difficulty check.} Next, the construction agent performs a feasibility check on the test case \(c\). In this check, the original test case data \(c\), together with the special property and the reference solution, are provided to a model widely recognized as highly capable (GPT-5.6 Sol in particular for our experiment). The test case passes the feasibility check if the model completes the proof within the specified time limit, so the construction of test cases depends directly on the pre-set time limits (600s in our experiment).

The construction agent then performs a difficulty check on the test case \(c\). The test case data \(c\) are provided to a sequence of less capable model, and the test case passes the difficulty check if at least one of the models fails to complete the proof. 

\textit{Human check.} Once a test case has passed all the checks above, it undergoes a final human review to verify the correctness of the reference solution and execution trace, as well as the validity of the preceding checks. Test cases that pass the human review are included in the benchmark.

\section{Experiments}

\subsection{Experimental Setting}

\paragraph{Models.} We evaluated 10 frontier LLMs that span the major model families: GPT-5.6-Luna, GPT-5.6-Terra, GPT-5.6-Sol, Claude-5-Sonnet, Claude-5-Opus, DeepSeek-V4-Flash, DeepSeek-V4-Pro, GLM-5.3-Flash, GLM-5.3, Kimi-K3.

\paragraph{Harness.} To construct complete LLM agent systems, we pair DeepSeek-V4-Pro with DeepSeek Harness, Claude-5-Sonnet and Claude-5-Opus with Claude Code, and other models with Codex. All reasoning effort settings have been set to \textit{high}.

\subsection{Basic Results}

In this section, we present the complete evaluation results for the systems described above and provide a preliminary analysis of the causes of failure in unsuccessful test cases.

\paragraph{Full Results.} The complete evaluation results for all models are presented in Figure \ref{fig:main-results}, and a detailed scoring table can be found in Appendix \ref{app:detailed-benchmark-scores}. Note that the score is an evaluation of the agent as a whole, rather than the model itself. As shown, our benchmark—particularly Tier 2—exhibits strong discriminative power across agents. Moreover, across the 11 model–harness configurations in Table \ref{tab:full-benchmark-scores}, the combined number of tasks solved in Tiers 1 and 2 is strongly positively correlated with the number solved in Tier 3 (Spearman’s \(\rho = 0.834\)).

\begin{figure}[htbp]
    \centering
    \centering
    \def\cropTop{2.5mm}
    \def\cropBottom{1mm}

    \includegraphics[
        width=\linewidth,
        trim={0mm \cropBottom{} 0mm \cropTop{}},
        clip
    ]{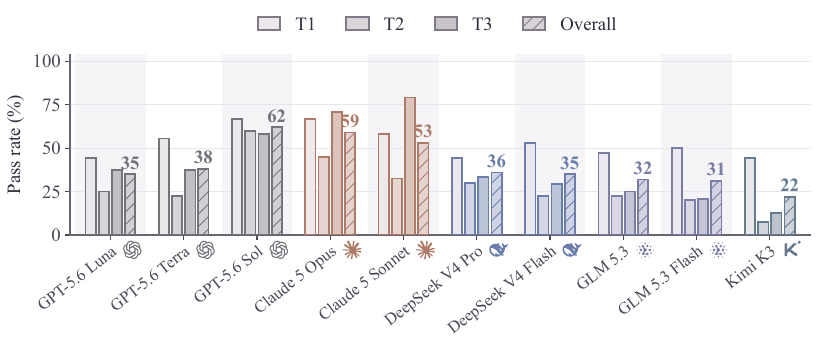}
    \caption{Benchmark scores normalized by total cases in Tier 1, Tier 2, Tier 3, and full benchmark.}
    \label{fig:main-results}
\end{figure}

\paragraph{Failure Analysis.} Using GPT-5.6-Sol and GPT-5.6-Terra as examples, we analyze the causes of failure in their unsuccessful test cases. To examine these causes more closely, we further count errors at both the \textit{method level} and the \textit{practice level}, the latter involving the interaction protocol and tool resources, and report the execution times of both models. Detailed statistics are provided in Figures \ref{fig:failure-analysis} (a) and (b); the labeling criteria appear in Appendix~\ref{app:failure-analysis}. 

\begin{figure}[htbp]
    \centering
    \def\cropTop{4mm}
    \def\cropBottom{3mm}

    \includegraphics[
        width=\linewidth,
        trim={0mm \cropBottom{} 0mm \cropTop{}},
        clip
    ]{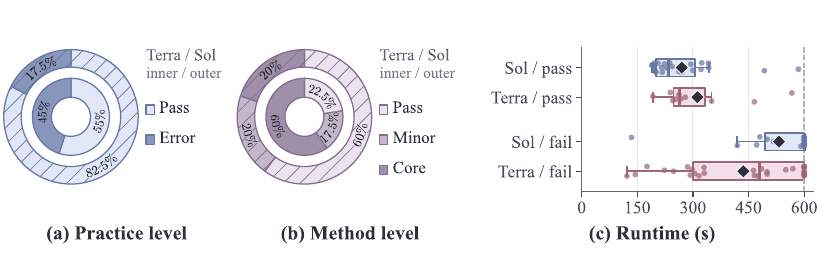}
    \caption{Failure analysis for GPT-5.6-Terra and GPT-5.6-Sol on 40 Tier-2 cases. (a) Proportions of runs with and without practice-level errors. (b) Method-level outcomes classified as Pass, Minor, or Core. Inner and outer rings represent Terra and Sol, respectively. (c) Runtime distributions for successful and failed runs. Dots represent individual runs and diamonds indicate means.}
    \label{fig:failure-analysis}
    \label{fig:time-distribution}
\end{figure}

We obtain the following two empirical observations:

\begin{itemize}
    \item \textbf{Practice level issues remain a major bottleneck in the proof process.} Across all 40 test tasks, practice level errors account for $17.5\%$ of Sol’s results and $45.0\%$ of Terra’s. Among failed runs only, the proportions involving verofocation level issues are $43.8\%$ and $58.1\%$, respectively. Moreover, the two models identified the correct methodology in 8 and 7 test cases, respectively, but failed to pass them for other reasons.
    \item \textbf{Sol completes successful tasks more quickly but consumes more of its budget on failed tasks.} In Figure \ref{fig:time-distribution}, the average runtimes of Sol and Terra on successful tasks are $269.$ and $312.3$ seconds, respectively, compared with $532.2$ and $436.1$ seconds on failed tasks. Practice level errors may cause runs to terminate early, whereas Sol’s greater number of full-budget failures indicates that its difficult tasks often continue until the time limit is reached.
\end{itemize}

\begin{table*}[htbp]
    \begin{minipage}[t]{0.38\textwidth}
        \vspace{0pt}
        \paragraph{Benchmark Robustness.}
        To assess the reliability of the benchmark scores, we evaluated
        GPT-5.6-Sol-High and GPT-5.6-Terra-High on the benchmark over three
        independent runs. Table~\ref{tab:3-shot-test} reports the mean and
        standard deviation of the resulting scores. The low variance across
        runs indicates that the evaluation results are highly reliable.
    \end{minipage}
    \hfill
    \begin{minipage}[t]{0.59\textwidth}
        \vspace{0pt}
        \centering
        
        \setlength{\tabcolsep}{6pt}
        \begin{tabular}{@{}lcc@{}}
            \toprule
            \textbf{Tier}
            & \textbf{GPT-5.6-Sol-High}
            & \textbf{GPT-5.6-Terra-High} \\
            \midrule
            Tier 1 (36) & $25.0 \pm 1.0$ & $19.7 \pm 0.6$ \\
            Tier 2 (40) & $21.0 \pm 2.6$ & $9.3 \pm 0.6$  \\
            Tier 3 (24) & $16.0 \pm 2.0$ & $12.3 \pm 3.1$ \\
            \midrule
            Total (100) & $62.0 \pm 2.0$ & $41.3 \pm 2.9$ \\
            \bottomrule
        \end{tabular}
        \caption{Results over three independent evaluation runs. Entries report the mean and standard deviation.}
        \label{tab:3-shot-test}
    \end{minipage}
\end{table*}

\section{Case Studies}
 
In this section, we conduct several case studies on \textsc{Merlin-Bench}, aimed at better leveraging models’ ability to produce verifiable proofs in practice.

\subsection{Case Study 1: Necessities of Tool Use and Problem Complexity}
\label{subsec:case-study-1}

The first case study investigates the effect of tool use on LLMs’ ability to complete interactive proofs. Furthermore, this experiment explains why problems of this level of complexity are chosen to evaluate the agent's proof process.

\paragraph{Method.} This case study includes two sets of ablation experiments. We choose three problems in Tier 1—Max SAT, Min-Cut/Max-Flow (MCMF) and Set Cover. On the one hand, we construct a temporary \emph{Tier 0} using three corresponding weakened problems in \(\mathsf{P}\)—2-SAT, MCMF (which is already in $\mathsf{P}$), and Minimum Vertex Cover in Bipartite Graphs (BGMC)—with six test cases generated for each problem using our method. We evaluate GPT-5.6-Luna, GPT-5.6-Terra, and GPT-5.6-Sol on this tier. On the other hand, after removing from \(B_c\) all permissions for \emph{tool calls} other than interaction with Arthur, we evaluate the base GPT-5.6-Terra and GPT-5.6-Sol models as provers for the three Tier-1 problems. Appendix~\ref{app:tool-use} provides task definitions and prompts.

\paragraph{Results.} The complete results of the two ablation experiments are shown in Tables \ref{tab:p-w-tool}.

\begin{table}[htbp]
    \centering
    \includegraphics[width=\linewidth]{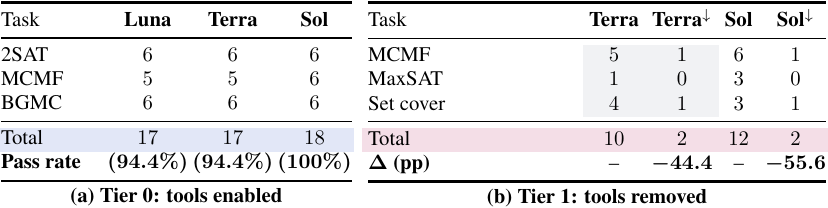}
    \caption{Ablations of tool-use and problem complexity. (a) Pass counts and overall pass rates for Luna, Terra, and Sol on 18 auxiliary Tier-0 cases with tools enabled. (b) Pass counts for Terra and Sol with and without tools on 18 selected Tier-1 cases. Downward arrows denote the no-tools condition, and $\Delta$ reports the change in overall pass rate in percentage points after removing tools.}
    \label{tab:p-w-tool}
    \label{tab:ip-wo-tool}
\end{table}

These results indicates the following empirical conclusions:

\begin{enumerate}
    \item \textbf{Tool use is essential for LLM agents to produce practical results, enabling them
to solve problems that base models cannot solve on their own.} As shown in Table \ref{tab:ip-wo-tool}, when no additional tool calls are permitted, both Sol and Terra score only 2 points, obtained from two small-scale test cases involving Set Cover and Min-Cut/Max-Flow, respectively. When tool calls are enabled, all frontier models achieve substantial scores.
    \item \textbf{Once tool use is taken into account, benchmarks constructed from problems in \(\mathsf{P}\) with known \emph{low degree polynomial time} algorithms  will almost inevitably become saturated.} As shown in Table \ref{tab:p-w-tool}, when tool use is permitted and the test-set difficulty is restricted to \(\mathsf{P}\), all three models in the GPT family achieve $90\%+$ scores. This result supports our motivation for raising the complexity of the benchmark problems from \(\mathsf{P}\) to \(\mathsf{IP}\).
\end{enumerate}

\subsection{Case Study 2: Influence of Expertise in User Prompt}

In practice, users can provide user prompts at different levels depending on their familiarity with the research problem. This case study investigates how user prompts at different levels affect models’ ability to complete interactive proofs.

\paragraph{Method.} We divide user prompts into three levels to simulate how three types of users affect the model’s proof capabilities:
\begin{description}
    \item[\color{t1} Engineer] The Engineer provides guidance on carrying out the interactive proof, including practice level suggestions such as recording Arthur’s messages and reserving sufficient time for the final submission.
    \item[\color{t2} Student] The Student provides general directions for an interactive-proof task but offers no test-case-specific guidance.
    \item[\color{t3} Expert] The Expert describes how the current instance is constructed and provides a strategy for solving it. Note that this guidance remains at the strategic level and does not provide a concrete solution method, making it fundamentally different from the feasibility check used during test-set construction.
\end{description}

For example, for the GNI task, the Student prompt provides the general direction of using an “invariant,” while the Expert prompt further identifies the local structure to be recovered and the corresponding parity computation, giving the model more concrete steps for making the required distinction. The complete prompt can be found in Appendix~\ref{app:user-expertise}. We fix the model under evaluation to GPT-5.6-Terra. 

\paragraph{Results.} 
From the results summarized in Figure \ref{fig:user-prompt}, we obtain the following empirical observations:

\begin{figure}[htbp]
    \centering
    \def\cropTop{4mm}
    \def\cropBottom{6mm}

    \includegraphics[
        width=\linewidth,
        trim={0mm \cropBottom{} 0mm \cropTop{}},
        clip
    ]{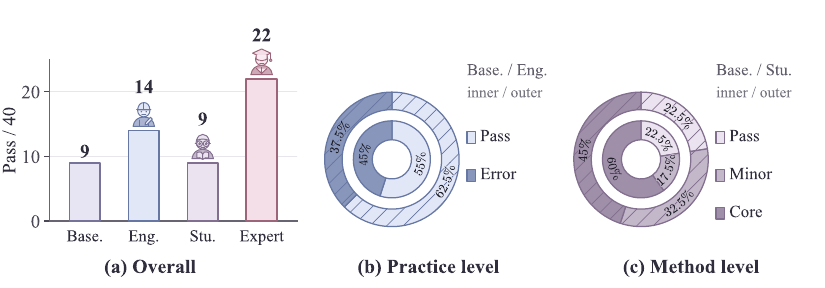}

    \caption{Effects of prompt expertise on GPT-5.6-Terra across 40 Tier-2 cases. (a) Numbers of cases passed under the Baseline, Engineer, Student, and Expert conditions. (b) Proportions of runs with and without practice-level errors under Baseline and Engineer prompts. (c) Method-level outcomes under Baseline and Student prompts, distinguishing Pass, Minor, and Core categories.}
    \label{fig:user-prompt}
    \label{fig:per-task}
    \label{fig:overall-scores}
    \label{fig:engineer-verification}
\end{figure}

\begin{enumerate}
    \item \textbf{Engineer prompts help reduce execution failures.} Figure \ref{fig:overall-scores} shows that the numbers of passed tasks under the Baseline, Engineer, Student, and Expert conditions are 9, 14, 9, and 22, respectively. Under the Engineer condition, the practice level error rate decreases from $45.0\%$ to $37.5\%$ (Figure \ref{fig:engineer-verification} (b)). 
    \item \textbf{General methodological guidance must still be translated into reliable execution.} From Figure \ref{fig:engineer-verification} (c), under the Student condition, the number of tasks with an inadequate core method decreases from 24/40 to 18/40, whereas the number of tasks with a broadly correct approach that nevertheless fail increases from 7/40 to 13/40. The total number of passed tasks remains 9/40. This indicates that even after learning the general algorithm for a task family, the model still struggles implement it as a complete proof process.
    \item \textbf{Domain-specific knowledge expands the range of tasks that the model can solve.} Under the Expert condition, 22/40 tasks pass, improved substantially compared to the baseline (9/40). 
\end{enumerate}

\subsection{Case Study 3: Transfer Experience on Synthetic Data to Real Tasks}

This case study investigates whether LLM agents' experience on a synthetic interactive proof dataset can be used to improve their performance on real-world verification tasks.

\paragraph{Method.} We ask the model to distill lessons from its Tier 1 and Tier 2 solution traces into a text, which we call a \(\textit{study guide}\). In this experiment, the study guide is provided to the model as a prompt. Specifically, we let agent \(B\) distill a study guide from the Tier 1 and Tier 2 solution trajectories. We then instantiate a new agent \(A\) and allow it to reason over the study guide for up to \(300\) seconds (sufficiently large in our experiment settings). Afterward, we provide the Tier 3 tasks to agent \(A\) and ask it to solve them. To emphasize the robustness of the results, we used our dataset construction pipeline to double the number of Tier 3 cases and ran three independent evaluations on the full set of 48. We denote the transfer setting by $A^{+B}$. Moreover, we introduce a blank control condition in which an agent generates a study guide \textit{without proof trajectories}. For an evaluated model \(A\), we denote this condition by \(A^{+0}\). We selected GPT-5.6-Sol and GPT-5.6-Terra as agents. The study guides, complete prompt and more explanations are provided in Appendix~\ref{app:transfer}. We also perform the procedure on two external benchmarks, AlgoVeri (\cite{zhao2026algoveri}) and CombiBench (\cite{liu2025combibench}) to test its capability of generalization.

\paragraph{Results.} Table \ref{tab:transfer} compares the baseline with the three transfer settings.

\begin{table}[htbp]
    \centering
    
    \includegraphics[width=\linewidth]{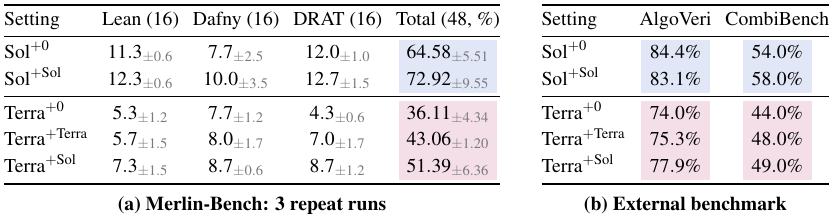}
    \caption{Results of the transferability experiment. (a) Results on 48 Tier-3 cases, reporting mean and standard deviation over three runs. (b) Pass rates on AlgoVeri (Dafny task) and CombiBench. The setting $A^{+B}$ evaluates model A using a study guide distilled from model B's Tier-1 and Tier-2 trajectories, while $A^{+0}$ uses a guide generated without proof trajectories.}
    \label{tab:transfer}
\end{table}

These results support the following empirical conclusion:

\begin{enumerate}
    \item \textbf{Synthesized experience can improve models’ ability to solve real-world tasks.} After incorporating the study guides, all three configurations achieve substantially higher scores on Tier 3. The same observation holds for the experiments on two external benchmarks except a subtle decrease of score in AlgoVeri ($\text{Sol}^{\text{+Sol}}$). Empirically, the gains are associated with the capabilities of both models involved. Comparing $\mathrm{Terra}^{+\mathrm{Sol}}$ with $\mathrm{Terra}^{+\mathrm{Terra}}$ shows that a stronger teacher for distilling the study guide leads to better downstream performance. Comparing $\mathrm{Sol}^{+\mathrm{Sol}}$ with $\mathrm{Terra}^{+\mathrm{Sol}}$ shows that a stronger learner for solving the tasks likewise yields better final performance.
    \item \textbf{Post-training on synthesized experience could potentially improve real-world performance.} In this experiment, we provide the study guide to the model solely through prompting, leaving the parameters unchanged. A promising direction for future work is to convert the Tier 1 and Tier 2 solution trajectories into training data and use them for post-training, with the goal of improving models’ ability to construct proofs using real-world verifiers.
\end{enumerate}
\section{Conclusion}

In summary, we introduce \(\textsc{Merlin-Bench}\), a benchmark that evaluates models’ ability to construct unified proofs according to specified requirements by placing LLM agents in the role of the prover within an interactive proof system. Our evaluation of 10 frontier LLMs shows that, beyond the ability to find certificates, constructing proofs according to the required specifications remains a significant capability bottleneck for LLMs. Our further case studies shows:

\begin{enumerate}
    \item Tool use is essential for LLM agents to produce practical results and must therefore be taken into account when constructing evaluation datasets.
    \item User prompts have a direct impact on the capabilities of LLM agents. Both guidance on formal proof construction and domain-specific expertise can improve model performance.
    \item Test-time scaling or post-training based on synthetic interactive proof datasets may improve models’ formal proof capabilities.
\end{enumerate}

\section*{AI use statement}

In this work, we used generative AI tools for generating synthetic data sets, assisting in the writing of proofs, implementing methods, assisting with translation and cleaning and reformatting dataset.
We have not used generative AI tools for helping develop theoretical models or conceptual frameworks, formulating mathematical claims, providing critical ingredients for proving mathematical claims, designing or providing feedback on research  methodology or experiments, supporting qualitative and thematic data analysis and interpreting results.
and proposing or refining hypotheses are not applicable to this work.
We reviewed all AI-assisted work. We
checked LLM-generated code and data sets. All the text in the paper has undergone rigorous human review. We take responsibility for the final content of this work,
including text, claims or artifacts produced with the aid of generative AI.

\bibliography{merlin_main,appendix_references}
\bibliographystyle{merlin_preprint}

\clearpage
\appendix
\numberwithin{equation}{section}
\section{Comparison to Related Works}

\subsection{More Related Works}

In this section, we review additional literature related to the topics of this paper.

\paragraph{Interactive proof in machine learning.} Interactive-proof complexity provides a formal prover–verifier framework for machine learning, with explicit guarantees concerning completeness, soundness, round complexity, communication complexity, sample complexity, and verification time. \cite{amit2025theory} introduce Self-Proving models that generate an answer and then interact with a fixed verifier to establish its per-instance correctness under a worst-case soundness guarantee. However, they treat the verifier as a reward model and do not use the properties $\mathsf{IP}$ to characterize the workflow of agent systems. \cite{goldwasser_et_al:LIPIcs.ITCS.2021.41} formulate IP-PAC, in which a resource-constrained verifier interacts with a prover to certify the approximate correctness of a learned hypothesis. \cite{gur2024iplearning} develop doubly efficient interactive proofs for agnostic learning tasks, including the recovery of heavy Fourier coefficients and the learning of $\textsf{AC}^0[2]$ functions and juntas. \cite{chiesa_et_al:LIPIcs.ITCS.2018.53} introduce IP distribution testers and show that interaction with an untrusted prover can exponentially reduce the sample complexity of distribution testing.

\paragraph{Agent-oriented benchmarks.} Several benchmarks have been developed specifically to evaluate different capabilities of LLM agents. \cite{yin2026investigating} investigates the advanced reasoning capabilities of models through interaction with black-box environments. \cite{liu2024agentbench} assess general-purpose agent capabilities across heterogeneous environments, including operating systems, databases, knowledge graphs, card games, and embodied household tasks. \cite{yao2025tau, boisvert2024workarena} evaluate agents on user-facing business operations and enterprise knowledge work, using policy-constrained airline and retail support tasks and compositional ServiceNow workflows, respectively. \cite{zhou2024web, xie2024osworld} assess agents’ ability to operate digital interfaces, focusing on functional websites and cross-application tasks in real operating systems, respectively.  \cite{jimenez2024swe} evaluate software-engineering agents by requiring them to resolve real GitHub issues through repository-level code modifications. \cite{siegel2025core, chen2025sciagent} evaluate scientific research agents on computational reproduction of published studies and the implementation of data-driven scientific analyses, respectively.

\paragraph{Lean, Dafny, and LLM-assisted formal verification.} Formal verification systems such as Lean \cite{deMoura2015Lean}, Dafny \cite{leino2010dafny}, and DRAT \cite{Wetzler2014DRATTrim} provide machine-checkable mechanisms for validating mathematical proofs, program correctness, and SAT certificates. \cite{zheng2022minif2f} introduce miniF2F, a cross-system benchmark of formalized Olympiad-level problems that using Lean verification. \cite{jiang2023draftsketchproveguiding} use informal proofs to guide the generation and completion of formal proof sketches. \cite{yang2023leandojo} introduce LeanDojo, an environment and benchmark for retrieval-augmented theorem proving in Lean. \cite{azerbayev2024llema} develop Llemma, a mathematics-focused language model that supports formal theorem proving in Lean. \cite{xin2025dsprover} use Lean’s verification feedback as a reinforcement-learning signal and combine it with Monte-Carlo tree search. \cite{kamran2024proof} propose proof-carrying code completions in which language models generate Dafny code together with machine-checkable safety proofs.

\subsection{Detailed Specification of the Four Merits} \label{app:merits}

In this section, we detail the specification of the four expected merits proposed in Section \ref{sec:introduction}.

\paragraph{Unified proof ability.} By unified proof ability, we mean the ability of an LLM agent to convince a human verifier, through a specified protocol and with high confidence, that a claim is true when the verifier does not know the answer in advance. In practice, the human verifier and the LLM agent agree on a protocol $\pi$ and interact according to it. If the interaction meets the agreed conditions, the verifier accepts the claim. The protocol can cover rigorous mathematical proofs, formal verification code, and tasks that allow small completeness and soundness errors (see our GNI task). Bounding the number of rounds and the verification cost of $\pi$ by low-degree polynomials gives the setting of our benchmark.

\paragraph{Agent oriented evaluation.} We evaluate the complete agent system, including the LLM, its harness, and its tools. We therefore want the benchmark to have two properties. (1) A bare LLM, without a harness or tools, has difficulty solving the benchmark problems. (2) When LLM agents can write code and search the web, the benchmark still distinguishes their performance. Case Study 1 (Section 5.1) was designed to demonstrate these properties.

\paragraph{Rigorous certification.} We want the procedure for checking a claim to have explicit completeness and soundness bounds. We want to be able to check claims rigorously even on sizable instances of problems that lack a low-degree polynomial-time decision procedure. Methods such as using an LLM as a judge offer no theoretical guarantee of correctness. Appendix E gives proofs of the completeness and soundness bounds for every protocol in our benchmark.

\paragraph{Scalable synthesis.} We mean that we can quickly generate test cases at scale that meet the three conditions above. Scale refers both to the number of test cases and to the size of the individual instances. One important difference from prior work is that, by building \textsc{Merlin-Bench} around interactive proofs (IP), we can evaluate an LLM agent even when the benchmark designer does not know whether a claim is true (see Appendix \ref{app:extensibility}). This makes it straightforward to include a wide range of decision problems in \textsf{IP}. Although our test cases span only 15 families, our dataset construction agent can identify a variety of special properties within these families and use them to generate diverse test cases.

\section{Supplementary Definitions and Evaluation Details}
\label{app:main-supplements}

\subsection{Analysis of Completeness and Soundness}
\label{app:perfect-completeness}

An interactive verifier uses randomness to test a proof. Its two possible
statistical errors have different consequences for evaluation. A
\emph{completeness error} rejects an honest proof of a true claim; a
\emph{soundness error} accepts a false claim, even when the prover chooses its
strategy adversarially. Small soundness error makes acceptance reliable,
whereas small completeness error prevents verifier randomness alone from
rejecting a correct proof strategy.

Let $\operatorname{Acc}(P,x;r)$ be the acceptance indicator for prover
strategy $P$, public instance $x$, and verifier randomness $r$. Writing
$\mathcal{Y}$ and $\mathcal{N}$ for the true and false claims covered by a
task's promise, perfect completeness means
\begin{equation}
 \forall x\in\mathcal{Y}\quad \exists P_x\quad
 \Pr_r[\operatorname{Acc}(P_x,x;r)=1]=1.
 \label{eq:perfect-completeness}
\end{equation}
Soundness error at most $\varepsilon$ means
\begin{equation}
 \forall x\in\mathcal{N}\quad \forall P^*\quad
 \Pr_r[\operatorname{Acc}(P^*,x;r)=1]\leq\varepsilon.
 \label{eq:soundness}
\end{equation}
For a task that checks a numerical output, the claim includes the proposed
output: the soundness statement concerns an incorrect value. Statistical
completeness replaces the right-hand side of
Equation~\ref{eq:perfect-completeness} by a lower bound of $1-\delta$.

This distinction matters in a one-shot benchmark. If completeness is
perfect, an honest strategy that follows the protocol succeeds for every
verifier challenge. A rejected execution cannot be attributed to an
unavoidable statistical rejection of that strategy. This makes an individual
failure easier to interpret and avoids introducing a completeness-related
reason to repeat an otherwise correct attempt. At the same time, the
soundness guarantee prevents a prover from obtaining credit for a false
claim except with the stated small probability.

\paragraph{Tier-2 guarantees.}
Every Tier-2 family has soundness error strictly below $2^{-23}$ under the
conditions of its proof. Five families have perfect completeness. The
Boolean-Circuit Image-Size Lower Bound protocol instead has statistical
completeness with $\delta<10^{-10}$. Table~\ref{tab:tier2-guarantees} lists
the guarantees for the published parameter ranges and links to the detailed
proofs in Section~\ref{app:tier2}.

\begin{table}[htbp]
\centering\small
\caption{Completeness error $\delta$ and soundness error $\varepsilon$ for
the six Tier-2 families. Each soundness bound is strictly below $2^{-23}$.}
\label{tab:tier2-guarantees}
\begin{tabularx}{\linewidth}{@{}>{\raggedright\arraybackslash}p{0.22\linewidth}
 >{\raggedright\arraybackslash}p{0.13\linewidth}
 >{\raggedright\arraybackslash}p{0.36\linewidth}X@{}}
\toprule
Task family & $\delta$ & Upper bound on $\varepsilon$ & Proof \\
\midrule
Graph Non-Isomorphism & $0$ & $2^{-24}$ & Lemma~\ref{lem:tier2-gni} \\
\addlinespace
Boolean-Hypercube Sum-Check & $0$ &
 $\displaystyle\frac{5m}{p}\leq\frac{2000}{p}<2^{-50}$ & Lemma~\ref{lem:tier2-sumcheck} \\
\addlinespace
GKR Circuit Evaluation & $0$ &
 $\displaystyle\frac{5\sum_{i=0}^{L-1}s_{i+1}}{p}<2^{-49}$ & Lemma~\ref{lem:tier2-gkr} \\
\addlinespace
Circuit Image-Size Lower Bound & $<10^{-10}$ & $e^{-16}<2^{-23}$ & Lemma~\ref{lem:tier2-image} \\
\addlinespace
Shamir-Style TQBF & $0$ &
 $\displaystyle\frac{\sum_q d_q}{p}\leq\frac{2060}{p}<2^{-49}$ & Lemma~\ref{lem:tier2-tqbf} \\
\addlinespace
Hamiltonicity Bind--Challenge--Open & $0$ & $2^{-24}$ & Lemma~\ref{lem:tier2-hamilton} \\
\bottomrule
\end{tabularx}
\end{table}

For the algebraic protocols, $p=2^{61}-1$. Sum-Check has $m\leq400$ rounds
with per-round degree at most five. In GKR, $s_i$ is the base-two logarithm
of layer $i$'s width and $\sum_i s_{i+1}<256$. In TQBF, $d_q$ is the degree
bound for round $q$, and the largest total over the six profiles is 2060.
The GNI and Hamiltonicity protocols use 24 independent challenge lanes.
For Image-Size, the 128 independent lanes and the 64-success acceptance
threshold give the soundness bound $e^{-16}$ and the completeness-error
bound $e^{-23.04}<10^{-10}$.

\subsection{Evaluation environment and success criterion}
For each evaluation, the evaluator samples a seed, generates the instance,
and creates a public workspace for a fresh agent. The workspace contains the
task description, public instance, protocol, schemas and evaluation client.
Reference solutions and private generation information remain with the
evaluator. Arthur runs as a local service; the client queries the public
session state and sends the messages prescribed by the task.

The 600-second wall-clock budget starts when Arthur loads the runtime.
Reading the task, local computation, proof construction, artifact upload and
Arthur's final verification all count toward this limit. A session is
one-shot: a rejection or timeout ends the attempt. The model must preserve
the interaction order and use the challenges actually returned by Arthur.

An accepted proof is recorded as a pass. The final score uses a strict pass,
which additionally requires a valid independent transcript audit and
compliance with the execution and resource constraints. Writing $z$ for
the public instance, $\tau$ for
the execution trajectory and $h_n$ for the protocol transcript, the scored
indicator is
\begin{equation}
 I_c=\mathbb{I}\!\left[
 V_c(z,h_n)=1\ \land\ \operatorname{Audit}(h_n)=\mathrm{VALID}
 \ \land\ \operatorname{Valid}_{B_c}(\tau)\right].
\end{equation}
The score for a fixed set of $N$ cases is $N^{-1}\sum_c I_c$.

\subsection{Base evaluation prompt}
Prompt~\ref{pr:base} supplies the common task instructions. The evaluator
instantiates the task title, task identifier and case identifier, and supplies
the corresponding public files. Braced uppercase fields in the prompt
listings denote such substitutions; they do not represent additional advice.
The public task specification fixes the mathematical problem and exact
message schemas, so it is read together with this prompt.

\promptfile{pr:base}{Base task instructions}{prompts/publication/base_task.txt}

\Needspace{18\baselineskip}
\subsection{Detailed benchmark scores}
\label{app:detailed-benchmark-scores}
Table~\ref{tab:full-benchmark-scores} reports the pass counts for each
model--harness configuration. The Codex + DeepSeek V4 Pro configuration is included here but
omitted from the main-paper figure.

\begin{table}[H]
\centering
\caption{Per-model pass counts across 36 Tier-1, 40 Tier-2, and 24 Tier-3 cases. Overall is the sum across all 100 cases.}
\label{tab:full-benchmark-scores}
\small
\begin{tabular*}{\linewidth}{@{\extracolsep{\fill}}llrrrr@{}}
\toprule[1.2pt]
Harness & Model & Tier 1 (36) & Tier 2 (40) & Tier 3 (24) & Overall (100) \\
\midrule[0.6pt]
Codex & GPT-5.6-Luna & 16 & 10 & 9 & 35 \\
Codex & GPT-5.6-Terra & 20 & 9 & 9 & 38 \\
Codex & GPT-5.6-Sol & 24 & 24 & 14 & 62 \\
Claude Code & Claude 5 Opus & 24 & 18 & 17 & 59 \\
Claude Code & Claude 5 Sonnet & 21 & 13 & 19 & 53 \\
DSH & DeepSeek V4 Pro & 16 & 12 & 8 & 36 \\
Codex & DeepSeek V4 Pro & 18 & 7 & 7 & 32 \\
Codex & DeepSeek V4 Flash & 19 & 9 & 7 & 35 \\
Codex & GLM 5.3 & 17 & 9 & 6 & 32 \\
Codex & GLM 5.3 Flash & 18 & 8 & 5 & 31 \\
Codex & Kimi K3 & 16 & 3 & 3 & 22 \\
\bottomrule[1.2pt]
\end{tabular*}
\end{table}

\subsection{Failure analysis and runtime measurements}
\label{app:failure-analysis}
The failure analysis separates methodology from practice. At the methodology
level, \emph{Core Methodology Missing} (CM) means that the trajectory does not
establish an adequate core approach, including failure to identify the
relevant structure or incorrect modeling of the central object.
\emph{Near Correct} (NC) means that an appropriate route is evidenced, but a
local computational, implementation, state-management or completion defect
prevents a strict pass. Evidence can include accepted protocol rounds,
independent exact checks or an almost complete proof artifact.

At the practice level, \emph{Practice Error} (PE) records an execution fault,
such as a malformed message, lost or repeated protocol state, an invalid
artifact, unauthorized workspace access, or violation of an explicit process
or resource boundary. A normal timeout caused solely by an insufficient
mathematical method is not automatically a practice error. CM and NC are
mutually exclusive among failures; PE is assessed separately and may coexist
with either. The main-paper percentages use all 40 Tier-2 cases as the
denominator, while a failure-only percentage uses the number of failed runs.

\begin{table}[htbp]
\centering
\caption{Failure labels used in the Tier-2 analysis.}
\label{tab:failure-labels}
\begin{tabularx}{\linewidth}{@{}llX@{}}
\toprule
Axis & Label & Evidence criterion \\
\midrule
Methodology & CM & No adequate core method is established. \\
Methodology & NC & A correct core route is evidenced but the proof is not completed successfully. \\
Practice & PE & An observable execution or interface fault affects validity or completion. \\
\bottomrule
\end{tabularx}
\end{table}

Elapsed time is measured from Arthur's start to the recorded outcome, using
the same 600-second budget. Successful and failed runs are summarized
separately, since early rejection and budget exhaustion describe different
execution behavior.

\clearpage
\section{Benchmark Construction}
\label{app:construction-evaluation}

\subsection{Construction procedure}
The benchmark contains 100 test cases in 15 task families: 36 deterministic
certificate tasks in Tier-1, 40 interactive tasks in Tier-2, and 24 native
verification tasks in Tier-3. A case specifies a public description $D_c$,
an instance generator $G_c$, an interactive protocol $\Pi_c$, a verifier $V_c$,
and resource constraints $B_c$. Different cases within a family share the
verification interface while varying the construction or its parameters.

Construction proceeds through literature review, implementation and automatic
checking, model-based feasibility and difficulty checks, and human review.
The construction agent identifies a structural restriction admitting a
specialized solution, implements a seeded generator and reference algorithm,
and writes the public description and verifier. The reference algorithm must
produce accepted evidence; invalid submissions must be rejected. Feasibility
is checked with a capable model given the structural explanation and reference
solution, whereas difficulty is checked with a less capable model given only
the ordinary public task. Human review checks the reference solution,
execution trace, and validity of these checks before inclusion.

In our experiments, since the expected solution has already passed formal verification, human review focuses primarily on the validity of the automated validation process, and the meaningfulness and reasonableness of the test cases, rather than on specific implementation details. More than $90\%$ of the generated test cases are ready for immediate use, while the remaining $10\%$ are replaced with more diverse alternatives to ensure comprehensive testing. The entire benchmark generation process takes approximately 4–5 days, with human review accounting for about 1–2 days of that time.

The following prompt gives the construction agent its instructions. It ends
with the production of a review package for the human reviewers.

\promptfile{pr:construction}{Instructions to the benchmark construction agent}{prompts/publication/construction.txt}

\subsection{On the Extensibility and Resistance to Saturation}
\label{app:extensibility}

The current release of \textsc{Merlin-Bench} evaluates a single true claim for each generated instance. Its instance-construction procedure searches the literature for special structures admitting low-degree polynomial-time algorithms and applies both feasibility and difficulty checks. This design is intended to produce meaningful performance differences at the current capability frontier; it is not intended to imply that the fixed set of instances can never become saturated. In this section, we describe a prospective bidirectional evaluation mode that can provide additional headroom once agents reliably solve the current benchmark. \emph{This mode is not used in any score reported in this paper.}

\paragraph{Existence of complementary language in $\textsf{IP}$.} Let \(L\) be a language, and let
\[
    L_1 = L
    \qquad\text{and}\qquad
    L_0 = \overline{L}.
\]
Since \(\mathsf{IP}=\mathsf{PSPACE}\) and \(\mathsf{PSPACE}\) is closed under complement, if \(L\in\mathsf{IP}\), both \(L_1\) and \(L_0\) admit interactive-proof systems. For several problem families considered in our benchmark, engineering-feasible protocols and verifiers are already available for both the problem and its complement. For example, the complement of Graph Non-Isomorphism is Graph Isomorphism, for which a particularly
simple certificate-based protocol suffices: Merlin supplies an isomorphism between the two graphs, and the verifier checks it in polynomial time. TQBF is likewise closed under complementation in a constructive sense. Let \((\Pi_1,V_1)\) be an interactive proof system for TQBF. To prove that a closed QBF \(\psi\) belongs to $\overline{\text{TQBF}}$, the protocol \((\Pi_0,V_0)\) first computes \(\neg\psi\), normalizing it into the required QBF representation if necessary, and then invokes \((\Pi_1,V_1)\) to prove that \(\neg\psi\) is true. Since \(\psi\) is false if and only if \(\neg\psi\) is true, this construction inherits the completeness and soundness guarantees of the original TQBF protocol.

\paragraph{The bidirectional evalutation mode.} For the same input \(x\), the evaluator opens one independent session for each side, using fresh and hidden verifier randomness. The two sessions are evaluated under a pre-specified pair-level resource budget. Let \(A_b(x)\in\{0,1\}\) denote whether the session for \(L_b\) strictly passes all protocol, transcript, and resource audits. We define the bidirectional score by
\[
    I_{\leftrightarrow}(x)
      = \mathbf{1}\!\left[A_0(x)\oplus A_1(x)\right].
\]
Thus, the agent receives credit exactly when one of the two sessions is accepted. When this occurs, the index of the accepted branch also serves as the agent's verifier-certified prediction of the decision bit. Acceptance of both branches and rejection of both branches receive zero credit. Importantly, the runtime evaluator need not store the true decision bit or compare the trajectory with a reference proof.

\paragraph{Analysis.} Suppose that both instantiated protocols have completeness and soundness errors at most \(2^{-23}\). Let $b=\mathbf{1}[x\in L]$. Then an sufficiently capable agent that realizes an honest prover strategy for the true branch \(b^\star\) within the resource budget will receive a score $1$ with probability at least
$$
\Pr[I_{\leftrightarrow}(x)]\geq \Pr\!\left[A_{b^\star}(x)=1\wedge A_{1-b^\star}(x)=0\right] \geq 1-2^{-23}-2^{-23}=1-2^{-22}.
$$
Simularly, a less capable agent that cannot solve any direction receive a score $1$ with probability at most $2^{-22}$. 

\paragraph{Saturation-resistant test-case construction.} The bidirectional evaluation mode is accompanied by a saturation-resistant test-case construction procedure, obtained from our original pipeline by removing the per-instance feasibility check. Because the same instance is evaluated under protocols for two complementary claims, the constructor need not know its truth value or provide a reference proof: when exactly one branch is accepted, that branch determines the verifier-certified answer. 

This procedure allows the construction agent to include finitely encoded open problems that can be faithfully formalized as supported interactive-proof tasks. For example, the existence of the Conway \(99\)-graph remains open and can be formalized by a QBF decision problem; bidirectional evaluation can test both its existence and nonexistence without knowing which claim is true. Up to the negligible protocol error, the resulting suite cannot be reliably ceiling-saturated as long as it contains at least one instance that the evaluated AI systems cannot resolve. Therefore, as suitable unresolved instances continue to be incorporated, this mode retains unsolved evaluation headroom. It does not, however, guarantee discrimination at the current capability level, since an instance on which every agent obtains \((0,0)\) prevents ceiling saturation but does not distinguish among those agents. We therefore retain the feasibility check in the present benchmark and reserve this construction for a future evaluation mode.
\clearpage
\section{Tier-1: Deterministic Certificates}
\label{app:tier1}

Tier-1 consists of one-message NP search tasks.  A public instance $x$ is
given to both parties; Merlin sends one candidate witness $w$, and Arthur
deterministically evaluates the task-specific polynomial-time relation
$R(x,w)$.  Acceptance means $R(x,w)=1$.  There is no verifier challenge,
interaction round, or opportunity to revise a rejected witness.  Each task
below has a 600-second wall-clock limit, and a timeout or malformed witness is
rejected.

\subsection{Max-Flow / Min-Cut}
\label{sec:tier1-maxflow}

The Max-Flow / Min-Cut task asks Merlin to certify, in one message, the exact
maximum $s$--$t$ flow value of a capacitated directed network.  The certificate
contains both a feasible flow, which proves a
lower bound, and a cut of the same value, which proves the matching upper
bound.  This is the standard primal--dual characterization of maximum flow
\cite{ford-fulkerson-1956}.

Formally, the public instance is a directed multigraph
$G=(V,E)$ with distinct directed edge occurrences $e=(u_e,v_e)$, capacities
$c_e\in\mathbb Z_{\geq0}$, and distinct vertices $s,t\in V$.  Parallel edges
are retained as separate occurrences.  Merlin submits an integer $F$, an
integer flow $f_e$ for every $e\in E$, and a set $S\subseteq V$.  Arthur
accepts precisely when
\[
  0\leq f_e\leq c_e\quad(e\in E),
  \qquad
  \sum_{e:v_e=v}f_e=\sum_{e:u_e=v}f_e
       \quad(v\in V\setminus\{s,t\}),
\]
and
\[
  \sum_{e:u_e=s}f_e-\sum_{e:v_e=s}f_e=F,
  \qquad
  \sum_{e:v_e=t}f_e-\sum_{e:u_e=t}f_e=F.
\]
Finally, $s\in S$, $t\notin S$ and, writing $\bar S:=V\setminus S$,
\[
  \operatorname{cap}(S,\bar S)
    :=\sum_{e:u_e\in S,\ v_e\notin S}c_e=F.
\]
Only edges directed from $S$ to its complement contribute to this cut.

\begin{lemma}[Exactness of the certificate]
If Arthur accepts $(f,F,S)$, then $f$ is a maximum $s$--$t$ flow and $S$ is a
minimum $s$--$t$ cut, both of value $F$.
\end{lemma}

\begin{proof}
Flow conservation implies that the value of any feasible flow equals its net
flow across every $s$--$t$ cut.  Capacity constraints therefore give
$F\leq\operatorname{cap}(T,V\setminus T)$ for every such cut $T$.  The submitted
flow and cut attain equality at $S$, so neither can be improved.
\end{proof}

After constructing a graph with the vertex names used below, the generator
independently samples a uniformly random bijection on its vertex identifiers,
carries the distinguished vertices $s,t$ through that bijection, and uniformly
permutes the list of directed edge occurrences.  These operations preserve
capacities, feasible flows, and cut values.

\paragraph{C1.1.1 Independent-channel calibration.}
Choose $k$ uniformly from $\{8,\ldots,16\}$ and independently and uniformly
sample $a_i\in\{3,\ldots,29\}$ for $1\leq i\leq k$.  The graph has vertices
$s,t,v_1,\ldots,v_k$ and exactly the edges $s\to v_i$ and $v_i\to t$, both
of capacity $a_i$.  Setting both edges of channel $i$ to flow $a_i$ gives
$F=\sum_i a_i$, and the singleton source side $S=\{s\}$ has that capacity.

\paragraph{C1.1.2 Regular bipartite matching network.}
Create two indexed sides $L=\{\ell_i:i\in\mathbb Z_{200}\}$ and
$R=\{r_i:i\in\mathbb Z_{200}\}$.  Add unit-capacity edges
$s\to\ell_i$ and $r_i\to t$.  Uniformly sample a 21-element subset
$D'\subseteq\mathbb Z_{200}\setminus\{0\}$ and set $D=\{0\}\cup D'$.
Add the unit-capacity edges $\ell_i\to r_{i+d}$ for all $i$ and $d\in D$,
with indices modulo 200.  Finally, uniformly sample without replacement 200
pairs $(i,j)$ for which $\ell_i\to r_j$ is not already present, and add those
unit-capacity edges.  The
result has 402 vertices and 5,000 edges.  The offset-zero edges form a perfect
matching, while the cut $\{s\}$ has capacity 200; hence the maximum-flow and
minimum-cut values are both 200.

\paragraph{C1.1.3 Large-integer sharded serial network.}
Let $v_0=s,v_1,\ldots,v_{256}=t$.  Between every consecutive pair
$v_{j-1},v_j$, split
$B=3{,}000{,}000{,}007$ into 72 positive parts as follows.  Independently for
each segment, uniformly sample 71 distinct cut points from
$\{1,\ldots,B-1\}$; if $0=q_0<q_1<\cdots<q_{71}<q_{72}=B$, use
$q_h-q_{h-1}$ as shard $h$'s capacity.  Uniformly sample without replacement
512 pairs $(a,b)$ from
$\{(a,b):0\leq a<b\leq256,\ b\geq a+2\}$ and add the return edges
$v_b\to v_a$, with capacities sampled independently and uniformly from
$\{1,\ldots,10^6\}$.  Thus the graph has 257 vertices and 18,944 edge
occurrences.  Sending every forward shard at capacity and zero flow on every
return edge gives value $B$; for every $0\leq j\leq255$, the prefix
$\{v_0,\ldots,v_j\}$ is an $s$--$t$ cut of capacity $B$.

\paragraph{C1.1.4 Wide layered permutation network.}
Arrange 64 columns of 64 vertices, indexed $x_{c,r}$ with
$c,r\in\mathbb Z_{64}$.  Add unit edges from $s$ to column 0 and from column
63 to $t$.  Independently at each transition $c\to c+1$, uniformly sample a
two-element subset $\{d_1,d_2\}\subseteq\mathbb Z_{64}\setminus\{0\}$ and add
the three unit-capacity cyclic matchings
\[
  x_{c,r}\longrightarrow x_{c+1,r+d},
  \qquad d\in\{0,d_1,d_2\},\quad r\in\mathbb Z_{64}.
\]
The graph has 4,098 vertices and 12,224 edges.  The offset-zero matchings give
64 edge-disjoint source--sink paths, and the source cut has capacity 64.

\paragraph{C1.1.5 Sparse cluster-interface network.}
For $0\leq a\leq47$, cluster $a$ has entry vertices $u_{a,r}$ and exit
vertices $v_{a,r}$ for $r\in\mathbb Z_{128}$.  Independently for each cluster,
uniformly sample a two-element subset
$\{d_{a,1},d_{a,2}\}\subseteq\mathbb Z_{128}\setminus\{0\}$, and add the
three unit-capacity cyclic matchings
$u_{a,r}\to v_{a,r+d}$ for $d\in\{0,d_{a,1},d_{a,2}\}$.  For every
$0\leq a<47$, independently sample $q_a$ uniformly from $\mathbb Z_{128}$ and
add the unit-capacity interface edges
$v_{a,r}\to u_{a+1,r+q_a}$.  Add unit-capacity edges $s\to u_{0,r}$ and
$v_{47,r}\to t$.  The graph has 12,290 vertices and 24,704 edges.  Following
the offset-zero internal matching in every cluster and every interface
matching yields 128 internally vertex-disjoint $s$--$t$ paths, matching the
capacity of the source cut.

\paragraph{C1.1.6 Deep width-two diagnostic network.}
Create layers $\{x_{i,0},x_{i,1}\}$ for $1\leq i\leq3000$.  For
$1\leq i<3000$ and $p,q\in\{0,1\}$, add the unit-capacity directed edge
$x_{i,p}\to x_{i+1,q}$.  Also add unit-capacity edges $s\to x_{1,p}$ and
$x_{3000,p}\to t$ for $p\in\{0,1\}$.  The graph has 6,002 vertices and
12,000 edges.  The two paths
$s,x_{1,p},\ldots,x_{3000,p},t$, $p\in\{0,1\}$, are internally
vertex-disjoint and contain 3,001 edges each; the source cut has capacity 2.

\subsection{Weighted Partial MaxSAT}
\label{sec:tier1-maxsat}

Weighted Partial MaxSAT asks for a Boolean assignment that satisfies every
mandatory clause while keeping the total weight of violated optional clauses
below a public threshold.  Thus the public object is a CNF formula whose
clauses are divided into a hard family $\mathcal H$ and a weighted soft family
$\mathcal S$.  A literal is $x_i$ or $\neg x_i$, a clause is a disjunction of
literals, and a CNF formula is a conjunction of clauses.  For an assignment
$a\in\{0,1\}^n$, define
\[
  \operatorname{cost}(a)
  =\sum_{(C,w)\in\mathcal S:\,C(a)=0}w.
\]
Merlin sends all $n$ Boolean values once.  Arthur accepts if every
$C\in\mathcal H$ is true and
$\operatorname{cost}(a)\leq K$, where $K$ is the published threshold.  Every
case sets
\[
 K=\min\{\operatorname{cost}(a):a\text{ satisfies every clause in }\mathcal H\}
\]
by the stated exact procedure.

After generation, the variables are uniformly permuted, every variable is
independently complemented with probability $1/2$, and the hard clauses, soft
clauses, and literals within each clause are independently uniformly
permuted.  This Boolean gauge transformation is a
bijection on assignments and preserves clause satisfaction and soft cost.

\paragraph{C1.2.1 Propagation calibration.}
Choose the number of planted-true variables uniformly from $\{79,80,81\}$,
then choose that many of the 160 variables uniformly without replacement.
Independently choose a uniformly random ordering
$v_1,\ldots,v_{160}$.  Add the unit clause $(v_1)$ if its planted bit is one
and $(\neg v_1)$ otherwise.  For every consecutive pair $u,v$, encode equality
as
$(\neg u\lor v)\land(u\lor\neg v)$ and disequality as
$(u\lor v)\land(\neg u\lor\neg v)$, choosing between them according to the
two planted bits.  These 319 hard clauses propagate a unique assignment.

Index the 240 soft clauses by $q=0,\ldots,239$.  Independently choose each
arity uniformly from $\{2,3\}$, choose that many distinct variables uniformly,
and choose one of their positions uniformly.  If $q\equiv0\pmod7$, give every
literal its planted-false polarity.  Otherwise give the distinguished position
its planted-true polarity and every other position its planted-false polarity.
Choose each weight independently and uniformly from $\{1,\ldots,31\}$.  The
instance has 559 clauses, and $K$ is the sum of the 35 violated soft-clause
weights.

\paragraph{C1.2.2 Weighted implication closure.}
The 12,000 variables are $x_{i,j}$ with $0\leq i<80$ and
$j\in\mathbb Z_{150}$.  Independently for each adjacent layer pair, sample six
offsets uniformly without replacement from $\{2,\ldots,149\}$, adjoin offsets
0 and 1, and add implications
$x_{i,j}\Rightarrow x_{i+1,j+d}$.  For each distance-two layer pair, add the
offset-0 and offset-1 implications.  There are 118,200 binary hard clauses
of the form $\neg x\lor y$.  Each variable also receives one weighted soft
unit: independently sample a fair preferred value and a weight uniformly from
$\{1,\ldots,997\}$, using $(x_{i,j})$ for a true preference and
$(\neg x_{i,j})$ for a false preference.  This gives 130,200 clauses in total.
True variables must form a successor-closed set $T$ in the acyclic implication
graph.  Give a vertex profit $p_v=w$ for a true preference and $p_v=-w$ for a
false preference.  If $W_+$ is the sum of all true-preference weights, then
\[
 K=W_+-\max_{T\text{ successor-closed}}\sum_{v\in T}p_v.
\]
The maximum-weight closure is computed exactly by its standard minimum-cut
reduction.

\paragraph{C1.2.3 Bounded-window weighted CNF.}
Order the variables as $x_1,\ldots,x_{2400}$ and sample their planted bits
independently and uniformly.  For every $p\geq2$, generate three hard clauses.
Each has arity $\min(3,p)$, contains $x_p$, and chooses its remaining distinct
variables uniformly from
$x_{\max(1,p-11)},\ldots,x_{p-1}$.  Choose one clause position uniformly and
give it its planted-true polarity; independently for every other position,
use its planted-true or planted-false polarity with equal probability.

At every position $p$, generate four soft clauses.  Choose the arity uniformly
from $\{1,\ldots,\min(3,p)\}$, include $x_p$, choose any remaining variables
uniformly without replacement from the same 11-position predecessor window,
and choose every polarity independently and uniformly.  Choose each weight
independently and uniformly from $\{1,\ldots,101\}$.  The instance has 7,197
hard and 9,600 soft clauses.  An exact left-to-right dynamic program stores the
least cost for each assignment to the last 12 variables, rejecting hard
clauses and charging soft clauses when their rightmost variable is processed;
the minimum over all final states is $K$.

\paragraph{C1.2.4 Locally compiled weighted closure.}
Start from core variables $x_{i,j}$ with $0\leq i<64$ and
$j\in\mathbb Z_{96}$.  Independently for each adjacent layer pair, sample four
cyclic offsets uniformly without replacement from $\mathbb Z_{96}$; for each
distance-two layer pair, sample one offset uniformly.  Add the corresponding
implication from every position to its shifted successor, giving 30,144
implications.  For every implication $x\Rightarrow y$, introduce
a private variable $z$ and use a single global variable $t$.  Add the hard unit
clause $(t)$ and, for each implication, the four hard clauses
\[
 (\neg x\lor\neg t\lor z),\quad(x\lor\neg z),\quad
 (t\lor\neg z),\quad(\neg z\lor y).
\]
When $t=1$, the first three clauses force $z=x$, so the fourth is exactly the
original implication.  For every core variable, independently sample a fair
preferred value and a weight uniformly from $\{1,\ldots,4093\}$, attaching
$(x)$ for a true preference and $(\neg x)$ for a false preference.  The public
formula has 36,289 variables, 120,577 hard clauses, and 6,144 soft clauses.
Its threshold is obtained from the exact closure value by the $W_+$ baseline
formula in C1.2.2.

\paragraph{C1.2.5 Sparse affine parity factors.}
Use 320 disjoint components of 16 free variables.  Each component has 44
three-variable parity factors.  The first 16 use the cyclic triples
$(i,i+1,i+3)$ modulo 16.  For each of the remaining 28 factors, independently
take the first three entries of a uniformly random permutation of the 16
variables; triples may repeat across factors.  Independently sample fair bits
$b,c$ and a weight uniformly from $\{1,\ldots,4093\}$.

To compile one factor on $x,y,z$, independently sample a fair split bit $\rho$
and introduce variables $q,o$ satisfying
\[
 q=x\oplus y\oplus\rho,
 \qquad o=q\oplus z\oplus(b\oplus\rho)
        =x\oplus y\oplus z\oplus b.
\]
For $\delta\in\{0,1\}$, write
$\operatorname{XOR}_{\delta}(u,v;q)$ for the relation
$q=u\oplus v\oplus\delta$.  When $\delta=0$, encode it by the four hard clauses
\[
 (u\lor v\lor\neg q),\quad(u\lor\neg v\lor q),\quad
 (\neg u\lor v\lor q),\quad(\neg u\lor\neg v\lor\neg q);
\]
for $\delta=1$, replace every occurrence of $q$ in this template by $\neg q$.
The two four-clause gadgets are therefore
$\operatorname{XOR}_{\rho}(x,y;q)$ and
$\operatorname{XOR}_{b\oplus\rho}(q,z;o)$.  Attach the weighted soft unit $(o)$
if $c=1$ and $(\neg o)$ otherwise.  Hence there are 5,120 free
variables, 28,160 gate-output
variables, 112,640 hard clauses, and 14,080 soft clauses.  Since components do
not share variables, enumerating the $2^{16}$ assignments inside each
component gives its exact minimum, and $K$ is the sum of the 320 minima.

\paragraph{C1.2.6 Hidden layered binary energy.}
Arrange 3,840 variables in layers $0,\ldots,239$, each of width 16.  Add a
binary factor for every unordered pair within a layer.  Independently for each
two consecutive layers, sample a uniformly random bijection of their 16
positions and add its 16 matched factors.  For every factor, independently
sample a fair preferred parity $b$ and a weight uniformly from
$\{1,\ldots,4093\}$.  If $b=0$, encode the factor by the weight-$w$ soft
clauses $(\neg x\lor y)$ and $(x\lor\neg y)$; if $b=1$, use
$(x\lor y)$ and $(\neg x\lor\neg y)$.  Both clauses are satisfied when
$x\oplus y=b$, and exactly one is violated otherwise.

There are $240\binom{16}{2}+239\cdot16=32{,}624$ factors and 65,248 soft
clauses, with no hard clauses.  For a layer state
$s\in\{0,1\}^{16}$, let $U_i(s)$ be its within-layer cost and let $E_i(r,s)$
be the matching cost between layers $i-1$ and $i$.  The exact transfer is
\[
 D_0(s)=U_0(s),\qquad
 D_i(s)=U_i(s)+\min_r\bigl(D_{i-1}(r)+E_i(r,s)\bigr),
 \qquad K=\min_sD_{239}(s).
\]
Because $E_i$ is a sum of 16 matched one-bit penalties, the minimization is
implemented by 16 successive one-bit min-plus updates on the $2^{16}$ states.

\subsection{Structured Set Cover}
\label{sec:tier1-set-cover}

Set Cover asks for a low-weight collection of public sets whose union contains
every element of a public universe \cite{chvatal-1979}.  Formally, an instance
contains a universe $U$, candidates $(S_i,w_i)$ with
$S_i\subseteq U$ and $w_i\in\mathbb Z_{>0}$ for $i\in[m]=\{0,\ldots,m-1\}$,
and a budget $B$.  Merlin submits an index set $I\subseteq[m]$.  Arthur accepts
exactly when
\[
  \bigcup_{i\in I}S_i=U
  \qquad\text{and}\qquad
  \sum_{i\in I}w_i\leq B.
\]
Several cases are more naturally described as weighted hitting set: choosing
a candidate represents selecting one variable or variable--value pair, and a
constraint is represented by a distinct universe element contained in exactly
the candidates that satisfy it.  This is the transpose of the same incidence
relation.  Every case publishes the exact minimum weight as $B$.  The
finalizer applies a uniformly random permutation to the universe labels and a
uniformly random permutation to the candidates, assigns the candidates fresh
opaque identifiers, and independently shuffles each published incidence list.

\paragraph{C1.3.1 Weighted group calibration.}
Partition $U$ into 24 disjoint groups $U_g$ of six elements.  For each group,
sample $w_g$ independently and uniformly from $\{5,\ldots,17\}$ and include
the full set $U_g$ with weight $w_g$.  Uniformly sample without replacement
eight of the $2^6-2=62$ nonempty proper subsets of $U_g$; each sampled subset
receives an independent uniform weight in
$\{w_g+1,\ldots,2w_g+1\}$.  No candidate meets two groups.  The instance has
144 elements and 216 candidates.  In each group, the full set has weight
$w_g$, whereas every cover omitting it contains at least one strictly heavier
proper subset.  Hence the unique optimum consists of the 24 full sets and has
weight $B=\sum_gw_g$.

\paragraph{C1.3.2 Relabeled cyclic overlap.}
Let the 3,072 universe elements be the edges $e_i=\{i,i+1\}$ of a cycle, with
indices modulo 3,072.  Candidate $S_i$ contains the two elements
$e_{i-1},e_i$ and receives an independent uniform weight
$w_i\in\{7,\ldots,41\}$.  Thus a selection vector
$x\in\{0,1\}^{3072}$ is a cover exactly when
$x_i\lor x_{i+1}=1$ for every $i$ modulo 3,072: it is a weighted vertex cover
of the cycle.  For each fixed value $a=x_0\in\{0,1\}$, a two-state dynamic
program scans $i=1,\ldots,3071$, retaining only transitions
$x_{i-1}\lor x_i=1$ and adding $w_ix_i$; the terminal state must also satisfy
$x_{3071}\lor a=1$.  The smaller of the two resulting costs is published as
$B$.

\paragraph{C1.3.3 Forced-marker Boolean-path calibration.}
Use 42 consecutive blocks, each containing 12 Boolean variables, for 504
variables in total.  Variable $x$ has two equal-weight candidates
$S_{x,0},S_{x,1}$ and a private selector element contained in exactly this
pair.  The 504 distinct variable weights are a permutation of
$10{,}000,\ldots,10{,}503$.  Sample a uniform 12-bit word and repeat it in all
42 blocks.  In each block, first use the 12 cyclic triples of coordinates
$(j,j+1,j+2)$ modulo 12, and then independently obtain 20 further triples by
uniformly shuffling the 12 coordinates and taking the first three.  For each
selected tuple, independently give every literal a fair polarity; if the
resulting clause is false under the planted word, choose one of its positions
uniformly and reverse that literal's polarity.

For each pair of adjacent blocks, first add 24 binary equality clauses: for
every coordinate $j$, add $x_j\lor\neg y_j$ and
$\neg x_j\lor y_j$, where $x_j$ and $y_j$ are the corresponding variables in
the left and right blocks.  For that same adjacent pair, add 24 additional
four-literal clauses.  For each such clause, independently shuffle the
variables of each block, take two from each shuffled order, and apply the same
polarity-and-repair rule.  Every clause occurrence is a distinct universe
element; for a literal satisfied by $x=b$, that element belongs to $S_{x,b}$.

The 42 local clause families and 41 boundary families each receive one marker
candidate.  Its private singleton element forces that marker to be selected,
and the same marker is incident to every clause element in its family.  Thus
all clause elements are covered by the 83 forced markers, while each selector
element forces at least one of its two value candidates.  Every marker has
weight $10{,}505$, so the exact lower bound
\[
 B=\sum_{q=10{,}000}^{10{,}503}q+83\cdot10{,}505
\]
is attained by selecting all markers and exactly one value candidate per
variable.  In particular, the forced markers already cover all clause
elements; the clauses do not impose an additional feasibility condition at
this budget.  The public instance has 3,899 elements, 1,091 candidates, and
3,312 represented clauses.

\paragraph{C1.3.4 Marked width-16 weighted hitting-set path.}
Create ordinary candidates $v_0,\ldots,v_{999}$ with independently and
uniformly sampled weights in $\{17,\ldots,997\}$, and define
$B_t=\{v_{8t},\ldots,v_{8t+15}\}$ for $0\leq t<124$.  Thus consecutive bags
share eight candidates.  In each bag, uniformly shuffle its 16 candidates and
use the four consecutive groups of four as constraints.  Until the bag has 32
constraints, independently choose a size uniformly from $\{3,4\}$, uniformly
shuffle the bag, take the first chosen number of candidates, and reject the
result if that candidate subset already occurs in the bag.  Every occurrence
of a constraint, including equal subsets in different bags, becomes a
distinct universe element incident precisely to its listed candidates.

Exact path-decomposition dynamic programming enumerates the $2^{16}$ choices
within each bag, discards every choice missing one of that bag's constraints,
and joins consecutive choices only when they agree on the eight shared
candidates.  Candidate weights are charged when their candidate is first
introduced.  Let $B$ be the resulting minimum ordinary-candidate cost.  After
computing $B$, add to every bag one marker candidate of weight $B+1$ incident
to all 32 of its constraint elements.  Consequently no marker occurs in a
cover of weight at most $B$.  There are 3,968 elements and 1,124 candidates.

\paragraph{C1.3.5 Marked width-18 weighted hitting-set path.}
Create 1,125 ordinary candidates with independent uniform weights in
$\{19,\ldots,1993\}$ and define
$B_t=\{v_{9t},\ldots,v_{9t+17}\}$ for $0\leq t<124$; consecutive bags share
nine candidates.  For each bag, uniformly shuffle its 18 candidates and split
the ordering consecutively into four groups of four and one group of two.
These five groups are the initial constraints.  Generate the remaining 27
constraints by the same uniform size-$3$-or-$4$ sampling and within-bag
duplicate rejection as in C1.3.4.  Each constraint occurrence is again a
distinct universe element.

The exact path dynamic program enumerates $2^{18}$ bag choices, retains those
hitting all 32 local constraints, and joins adjacent choices on their nine
common candidates.  Let $B$ be its minimum ordinary-candidate cost.  Only
afterwards, add one marker of weight $B+1$ per bag, incident to the 32 elements
of that bag.  The public instance has 3,968 elements and 1,249 candidates.

\paragraph{C1.3.6 Marked width-16 weighted hitting-set tree.}
Number 109 bags in breadth-first order, with root 0 and parent
$\lfloor(t-1)/3\rfloor$ for bag $t>0$.  The root bag contains
$v_0,\ldots,v_{15}$ and owns $v_0,\ldots,v_9$.  All 16 root-bag candidates are
introduced and charged at the root; ``owned'' denotes only the ten candidates
eligible to occur in a child separator.  Every non-root bag introduces ten
fresh candidates and owns them.  For the first, second, and third child of
a bag whose owned candidates in order are $a_0,\ldots,a_9$, respectively use
the six shared candidates
\[
 (a_0,a_1,a_2,a_3,a_4,a_5),\quad
 (a_4,a_5,a_6,a_7,a_8,a_9),\quad
 (a_0,a_1,a_2,a_7,a_8,a_9).
\]
Together with the ten fresh candidates, these form the child's width-16 bag.
There are 1,096 ordinary candidates, each given an independent uniform weight
in $\{23,\ldots,1597\}$.

For every bag, a uniform shuffle partitioned into four groups of four gives
the first four constraints.  Apply the size-$3$-or-$4$ sampling and within-bag
duplicate rejection of C1.3.4 until there are 34 constraints.  Every
constraint occurrence gives a distinct universe element.  A bottom-up exact
tree-decomposition dynamic program enumerates feasible $2^{16}$ choices for
each bag; for every six-bit parent-separator choice, it adds the costs of the
three child subtrees and charges only candidates introduced at the current
bag.  Let $B$ be the minimum root cost.  Finally, add to each bag a marker of
weight $B+1$ incident to all 34 elements of that bag.  The instance has 3,706
elements and 1,205 candidates.

\subsection{Structured Closest Vector}
\label{sec:tier1-cvp}

The Closest Vector Problem asks for a lattice point near a target.  Here the
lattice is given by a full-column-rank integer matrix
$B\in\mathbb Z^{d\times n}$, whose columns generate
$L(B)=\{Bz:z\in\mathbb Z^n\}$.  The public target is the rational vector
$\tau/D$, where $\tau\in\mathbb Z^d$ and $D\in\mathbb Z_{>0}$, and the public
data also contain a squared-radius numerator $K$ and coefficient bound $C$.
Merlin submits $z\in\mathbb Z^n$.  Arthur accepts exactly when
\[
  |z_j|\leq C\quad(1\leq j\leq n),
  \qquad
  \lVert DBz-\tau\rVert_2^2\leq K.
\]
All arithmetic in this relation is integral and exact.  The generated
instances have exactly one lattice point satisfying the acceptance inequality:
the planted point, whose squared residual numerator is $K$.  To hide the
canonical coordinates, independently sample uniformly random signed
permutation matrices $R$ and $Q$ on the rows and columns, respectively, and
publish
\[
 B'=RBQ,\qquad \tau'=R\tau.
\]
The planted coefficient vector becomes $z'=Q^{-1}z$.  Thus
$DB'z'-\tau'=R(DBz-\tau)$, and both the Euclidean residual and the coefficient
bound are preserved.  All six cases publish $C=32$.

\paragraph{C1.4.1 Orthogonal residual calibration.}
Let $n=24$ and independently sample each diagonal scale $a_i$ uniformly from
$\{11,\ldots,29\}$.  Before hiding, $B\in\mathbb Z^{48\times24}$ has the
diagonal matrix
$\operatorname{diag}(a_1,\ldots,a_{24})$ in its first 24 rows and 24 zero rows
below it.  For each $i$, sample $w_i$ uniformly from $[-9,9]\cap\mathbb Z$,
replacing a sampled zero by an independent fair sign in $\{-1,1\}$; sample
$r_i$ by an independent fair sign times a uniform magnitude in
$\{1,2,3\}$.  Set
$\tau=(a_1w_1,\ldots,a_{24}w_{24},r_1,\ldots,r_{24})$, $D=1$, and
$K=\sum_i r_i^2$.  For any $z$, the squared residual numerator is
$K+\sum_i a_i^2(z_i-w_i)^2$, so $z=w$ is the unique integer vector satisfying
the threshold.

\paragraph{C1.4.2 Hidden signed path equations.}
Let $n=128$.  Construct a lower-bidiagonal matrix $A$ by independently and
uniformly sampling every positive diagonal entry from $\{7,\ldots,31\}$ and
every subdiagonal magnitude from the same set, with an independent fair sign.
The positive diagonal makes $A$ nonsingular.  Set
$B=\begin{bmatrix}A\\0_{128\times128}\end{bmatrix}
\in\mathbb Z^{256\times128}$.  Sample each $w_i$ uniformly from
$[-13,13]\cap\mathbb Z$, replacing zero by an independent fair sign, and
sample each $r_i$ as an independent fair sign times a uniform magnitude in
$\{1,\ldots,5\}$.  Publish $\tau=(Aw,r)$, $D=1$, and
$K=\lVert r\rVert_2^2$.  The squared residual is
$\lVert A(z-w)\rVert_2^2+K$, which is at most $K$ exactly when $z=w$.

\paragraph{C1.4.3 Paired-residual low-width tree.}
Index 128 variables by $i=0,\ldots,127$ in a rooted ternary heap, with parent
$\lfloor(i-1)/3\rfloor$ for $i>0$.  Independently sample the positive entry
$A_{i,i}$ uniformly from $\{17,\ldots,43\}$.  When present, the entries at
the parent and grandparent columns have independent uniform magnitudes in the
same set and independent fair signs; all other entries are zero.  Thus $A$ is
lower triangular with positive diagonal.  Sample $w_i$ uniformly from
$[-15,15]\cap\mathbb Z$, replacing zero by a fair sign, and sample $r_i$
uniformly from $\{1,\ldots,7\}$.  Rows $2i$ and $2i+1$ of $B$ are both
$A_i$, with respective target values $A_iw+r_i$ and $A_iw-r_i$; set $D=1$.
Thus $B$ has 256 rows and
\[
 (A_i(z-w)-r_i)^2+(A_i(z-w)+r_i)^2
   =2(A_i(z-w))^2+2r_i^2.
\]
The threshold is $K=2\sum_i r_i^2$.  The positive diagonal makes $A$
nonsingular, so equality is unique.

\paragraph{C1.4.4 Hidden modular evaluation lattice.}
Construction A converts a linear code $C\subseteq\F_q^n$ into the integer
lattice $\{y\in\mathbb Z^n:y\bmod q\in C\}$.  In this case, over $\F_{257}$,
let $C$ be the length-128, dimension-20 Reed--Solomon code
obtained by evaluating polynomials of degree below 20 at
$1,\ldots,128$ \cite{reed-solomon-1960}.  Its minimum Hamming distance is 109.
Write $G\in\{0,\ldots,256\}^{128\times20}$ for the integer lift of the
Vandermonde evaluation matrix,
$G_{i,j}=i^j\bmod257$ for $1\leq i\leq128$ and $0\leq j<20$.  The generated
matrix is
\[
 B=\bigl[G\mid257e_{21}\mid\cdots\mid257e_{128}\bigr]
   \in\mathbb Z^{128\times128}.
\]
The first 20 rows of $G$ form a nonsingular Vandermonde matrix, so $B$ has full
rank.  Its column lattice is a sublattice of the Construction-A lattice
$L_A=\{y\in\mathbb Z^{128}:y\bmod257\in C\}$; no claim of equality with
$L_A$ is needed below.

Independently sample each coefficient uniformly from $[-7,7]\cap\mathbb Z$,
replacing zero by a fair sign, and let $y=Bw$.  Let the error set contain
coordinates $1,\ldots,20$, and uniformly choose its remaining 12 coordinates
without replacement from $\{21,\ldots,128\}$.  Set $D=5$ and define
\[
 \tau_i=\begin{cases}5y_i-3,&i\text{ erroneous},\\5y_i+2,&i\text{ clean}.
 \end{cases}
 \qquad K=32\cdot3^2+96\cdot2^2=672.
\]
Every nonzero vector of $L_A$, and hence of the column lattice of $B$, has
squared norm at least 109: a nonzero residue codeword has Hamming weight at
least 109, whereas a nonzero vector congruent to zero has a coordinate of
magnitude at least 257.  Since
$4K<5^2\cdot109$, the planted point lies strictly within half the
minimum lattice distance and is the unique point in the threshold ball.

\paragraph{C1.4.5 Column-mixed tensor-code lattice.}
Use the tensor product of two $[16,12]$ Reed--Solomon codes over $\F_{257}$.
Its 256 coordinates are ordered with the $12\times12$ grid
$I=\{1,\ldots,12\}^2$ first, followed lexicographically by the remaining
points of $\{1,\ldots,16\}^2$.  Let $G$ be the integer lift in
$\{0,\ldots,256\}$ of the 144 columns evaluating $X^iY^j$ for
$0\leq i,j<12$, and initially set
\[
 B_0=\bigl[G\mid257e_{145}\mid\cdots\mid257e_{256}\bigr].
\]
The evaluation submatrix on $I$ is nonsingular, so $B_0$ has full rank.  Its
column lattice is a sublattice of the tensor-code Construction-A lattice,
whose minimum Hamming distance is $5^2=25$.

Mix the columns by elementary determinant-one shears.  For 18 layers,
uniformly shuffle columns $0,\ldots,143$ except columns $0,1,12$, pair
successive columns, and for each ordered pair $(s,t)$ replace
$B_t\leftarrow B_t+\epsilon B_s$ for an independent fair sign $\epsilon$.
Apply another 18 such layers to columns $144,\ldots,255$.  Finally, for each
$t=0,\ldots,143$, perform three independent updates
$B_t\leftarrow B_t+\epsilon B_s$, where $s$ is uniformly sampled from
$\{144,\ldots,255\}$ and $\epsilon$ is a fair sign.  These operations preserve
the column lattice.

Sample every coefficient as in C1.4.4 and let $y=Bw$.  Independently and
uniformly permute $1,\ldots,16$ twice, obtaining $\alpha$ and $\beta$, and use
the perfect-matching error set
$E=\{(\alpha_j,\beta_j):1\leq j\leq16\}$.  With $D=13$, define the signed
target offset $\eta=\tau-13y$ by $\eta_{a,b}=-7$ on $E$ and $+1$ elsewhere.
Thus $K=16\cdot49+240=1024$.  Every nonzero vector in the containing
Construction-A lattice has squared norm at least 25, and
$4K<13^2\cdot25$ makes the planted point unique.

\paragraph{C1.4.6 Column-mixed Boolean-cube code lattice.}
Let $C$ be the degree-at-most-three multilinear evaluation code on
$\{0,1\}^8$ over $\F_{257}$, the Boolean-cube analogue of the order-three
Reed--Muller code.  Its 93 basis functions are the square-free monomials of
degree at most three, and every nonzero codeword has Hamming weight at least
$2^{8-3}=32$ \cite{reed-muller-1954}.

Order the monomials first by degree and then lexicographically.  Enumerate the
256 Boolean points lexicographically and greedily retain a point whenever its
93-entry evaluation row is linearly independent modulo 257 of the retained
rows; stop after 93 points, then append all remaining points in lexicographic
order.  The selected information set is precisely
$I=\{x\in\{0,1\}^8:\operatorname{wt}(x)\leq3\}$.  Its monomial-evaluation
minor is a unimodular zeta matrix after compatible subset ordering.  Hence the
integer evaluation columns, followed by $257e_j$ for the 163 coordinates
outside $I$, form a basis $B_0$ of the Construction-A lattice.

For 28 layers, uniformly shuffle the 163 modulus columns, pair successive
columns, and update the second column of each pair by adding the first with an
independent fair sign.  The final unpaired column is left unchanged in that
layer.  Then, for each of the 93 code columns, perform eight
updates that add an independently and uniformly sampled modulus column with
an independent fair sign.  Let $B$ be the resulting basis.  Sample every
coefficient as in C1.4.4 and set $y=Bw$.  With $D=13$, define
$\eta=\tau-13y$ to be $-7$ at the first 15 points of the ordered information
set and $+1$ at the other 241 coordinates.  The threshold is
$K=15\cdot49+241=976$.  Since $4K<13^2\cdot32$, the planted point is the
unique lattice point satisfying the threshold.

\subsection{List Graph Homomorphism}
\label{sec:tier1-list-hom}

List Graph Homomorphism asks whether a source graph can be mapped into a
target graph while respecting both adjacency and a vertex-specific set of
allowed images \cite{hell-nesetril-2004}.  The public instance consists of
finite undirected simple graphs $G=(V_G,E_G)$ and $H=(V_H,E_H)$ and a nonempty
list $L(v)\subseteq V_H$ for every $v\in V_G$.  Merlin submits a total map
$h:V_G\to V_H$; it need not be injective.  Arthur accepts exactly when
\[
 h(v)\in L(v)\quad(v\in V_G),
 \qquad
 \{h(u),h(v)\}\in E_H\quad(\{u,v\}\in E_G).
\]
The generator applies independent uniformly random permutations to the source
and target vertex labels and randomizes the serialization order of edges,
list records, and entries within each list, while retaining each record's
source-vertex identifier.  The two relabelings induce a bijection between the
valid homomorphisms before and after this finalization.

Cases C1.5.3--C1.5.6 use the Cai--F{\"u}rer--Immerman (CFI) construction
\cite{cai-furer-immerman-1992}.  For a cubic base graph $Q$, order the three
incidences at each base vertex $v$.  Replace $v$ by six outer vertices
$a_{v,i}^b$, where $i\in\{0,1,2\}$ and $b\in\F_2$, and four middle vertices
$m_{v,x}$ indexed by
\[
 \mathcal E=\{000,011,101,110\}.
\]
For every $x\in\mathcal E$ and $i\in\{0,1,2\}$, add the internal edge
$\{m_{v,x},a_{v,i}^{x_i}\}$.  If base edge $e=uv$ occupies
incidence $i$ at $u$ and $j$ at $v$, its twist $z_e\in\F_2$ adds the two
external edges
\[
 \{a_{u,i}^b,a_{v,j}^{b\oplus z_e}\},\qquad b\in\F_2.
\]
For any even frame $p_v\in\mathcal E$ at every base vertex, replacing
$z_e$ by $z_e\oplus p_{u,i}\oplus p_{v,j}$ produces an isomorphic graph: the
isomorphism sends $a_{v,i}^b$ to $a_{v,i}^{b\oplus p_{v,i}}$ and
$m_{v,x}$ to $m_{v,x\oplus p_v}$.  This explicit map is the planted witness
used below.  In each CFI case, sample $z$ uniformly from the even-parity
vectors in $\F_2^{E(Q)}$ by sampling its first $|E(Q)|-1$ coordinates fairly
and choosing the last to make the XOR zero.  Independently sample every
$p_v$ uniformly from $\mathcal E$ and use the displayed rule to obtain the
target twist vector.  For each published role $r$, a source vertex of role
$r$ receives as its list all target vertices of the same role.

\paragraph{C1.5.1 Anchored binary path calibration.}
The source is the disjoint union of 64 paths of 16 vertices, hence has 1,024
vertices and 960 edges; the target is the triangle $K_3$.  Order each path as
$v_0,\ldots,v_{15}$.  Independently for each path, uniformly sample an ordered
pair $(a,b)$ of distinct target vertices.  Set $L(v_0)=\{a\}$ and
$L(v_i)=\{a,b\}$ for $1\leq i<16$.  Since $K_3$ is loopless, each source edge
forces its endpoints to take different values.  The unique component map is
therefore $h(v_i)=a$ for even $i$ and $h(v_i)=b$ for odd $i$.

\paragraph{C1.5.2 Hidden tree permutation constraints.}
The source is the complete rooted 4-ary tree with the root at depth zero and
leaves at depth five; it has 1,365 vertices and 1,364 edges.  Assign every
source vertex $v$ a disjoint six-vertex target domain
$D_v=\{(v,0),\ldots,(v,5)\}$, so the target has 8,190 vertices.  For each
source edge from parent $u$ to child $v$, independently and uniformly sample a
permutation $\pi_{uv}$ of the six states.  Let the target edge set be exactly
the union of the matching edges
\[
 \{(u,i),(v,\pi_{uv}(i))\},\qquad 0\leq i<6,
\]
over all source edges.  Independently sample a uniform root state $s_0$ and set
$L(r)=\{(r,s_0)\}$; every other source vertex $v$ has list $D_v$.
Propagating the matching constraints down the tree determines a unique map.
The target has $1{,}364\cdot6=8{,}184$ edges.

\paragraph{C1.5.3 Coarsely listed switched CFI isomorphism.}
Independently sample three uniformly random perfect matchings between two
sides of 120 vertices each, rejecting the attempt if two matchings share an
edge.  Also reject unless their union is connected and the following 240
root-marked color-refinement signatures are pairwise distinct.  For a root
$r$, initially color vertices by their bipartition side and append one bit
indicating whether the vertex is $r$.  Repeatedly replace each color by the
pair consisting of its current color and the sorted multiset of neighbor
colors.  Encode these pairs canonically in lexicographic order, using the same
encoding rule in every rooted run, and perform 240 refinement rounds.  The
signature is the sequence of the 240 sorted histograms obtained after these
rounds.  The resulting cubic base has 240 vertices and 360 edges.

Build two 2,400-vertex, 3,600-edge CFI graphs whose twist vectors are related
by the twists and frames defined above.  The four published roles are
$(\mathrm{outer},\mathrm{left})$,
$(\mathrm{outer},\mathrm{right})$,
$(\mathrm{middle},\mathrm{left})$, and
$(\mathrm{middle},\mathrm{right})$.  Thus the two outer-role lists have size
720 and the two middle-role lists size 480; neither a base vertex nor an
incidence is specified.  The frame map stays within each role and is a valid
list-respecting isomorphism.

\paragraph{C1.5.4 Double-line role-listed switched CFI isomorphism.}
Repeat the three-matching sampling, connectivity test, and 120-round
root-marked refinement test of C1.5.3 with 60 vertices per side.  In this
refinement test only, all unmarked base vertices initially have one color;
the later list roles still retain the two bipartition sides.  Each of the two
CFI graphs has 1,200 vertices and 1,800 edges.  For a simple graph $X$, its
line graph $L(X)$ has
one vertex for every edge of $X$, with two vertices adjacent exactly when the
corresponding edges share an endpoint.  Apply this operation twice to each of
the two frame-related CFI graphs.  The first line graph has 1,800 vertices and
3,600 edges; the published graph $L(L(X))$ has 3,600 vertices and 10,800
edges.  An isomorphism of $X$ acts on edges and therefore induces an
isomorphism through both line-graph operations.

A published vertex represents a pair of incident edges of $X$, or a
``wedge'' centered at their common endpoint.  Lists group target wedges by six
roles.  A middle-centered wedge is classified by its base side.  An
outer-centered wedge is classified by its base side and by whether the third
edge incident to its center, omitted from the wedge, is external to the CFI
gadget.  For each side, the resulting target-list sizes are 720 for middle
centers, 360 for outer centers with an omitted external edge, and 720 for
outer centers with an omitted internal edge.  A source wedge receives the
entire target class of its role.

\paragraph{C1.5.5 Symmetric-cover switched CFI isomorphism.}
Begin with the incidence graph of the generalized quadrangle $GQ(2,2)$
\cite{payne-thas-2009}.  Concretely, its two parts are the 15 unordered pairs
of a six-element set and the 15 partitions of that set into three unordered
pairs; incidence means membership of a pair in a partition.  This is a cubic
30-vertex, 45-edge bipartite graph.  Number the 15 pairs and the 15 partitions
lexicographically, with the pair vertices first, and take the breadth-first
spanning tree rooted at vertex zero, visiting neighbors in increasing order.
Orient every base edge from its smaller-numbered endpoint to its
larger-numbered endpoint, assign a voltage
$\nu_{uv}\in\mathbb Z_6$, and set $\nu_{vu}=-\nu_{uv}$.  Give every tree edge
voltage zero, uniformly choose one non-tree edge and give it voltage one, and
give every other non-tree edge an independent uniform voltage in
$\mathbb Z_6$.  Replace a base vertex $v$ by $(v,k)$ for $k\in\mathbb Z_6$
and an oriented base edge $uv$ by the six edges
\[
 (u,k)(v,k+\nu_{uv}),\qquad k\in\mathbb Z_6.
\]
The voltage-one fundamental cycle generates $\mathbb Z_6$, so the cyclic lift
is connected.  It has 180 vertices, 270 edges, and girth at least eight.
Applying the CFI construction gives source and target graphs, each with 1,800
vertices and 2,700 edges.  The four roles are outer/middle crossed with the
pair/partition side of the lifted incidence graph; their target-list sizes are
540 for each outer role and 360 for each middle role.

\paragraph{C1.5.6 Independent double-line switched-CFI instance.}
Repeat all matching sampling, rejection tests, CFI switching, two line-graph
operations, wedge-role lists, and final source/target relabelings of C1.5.4
using an independent domain-separated random stream.  The resulting source
and target again have 3,600 vertices and 10,800 edges each, with the same six
wedge-role list classes.

\subsection{Rooted Group Steiner Tree}
\label{sec:tier1-group-steiner}

Rooted Group Steiner Tree asks for a low-weight connected subgraph that starts
at a distinguished root and reaches at least one representative of every
prescribed vertex group \cite{garg-konjevod-ravi-2000}.  The public instance is
a connected simple undirected graph $G=(V,E)$ with positive integer weights
$w_e$, a root $r$, nonempty groups
$\mathcal G=(G_1,\ldots,G_q)$, and budget $B$.  Groups may overlap.  Merlin
submits one edge set $T\subseteq E$.  Let
$R(T)=\{v\in V:v\text{ is reachable from }r\text{ in }(V,T)\}$.  Arthur
accepts exactly when both endpoints of every edge in $T$ lie in $R(T)$,
$R(T)\cap G_j\ne\varnothing$ for every $j$, and
$\sum_{e\in T}w_e\leq B$.  Thus $T$ is a directly checkable one-message
certificate.

Cases C1.6.3--C1.6.5 use a common binary-choice gadget.  Variable $x_i$ has a
center $z_i$, two leaves $x_i^0,x_i^1$, a root edge $rz_i$, and choice edges
$z_ix_i^0,z_ix_i^1$.  The singleton group $\{z_i\}$ forces the root edge, and
the group $\{x_i^0,x_i^1\}$ forces at least one choice.  Choice-edge weights
use a common base $M>n(31-3)$ for $n$ gadgets.  Every constraint group in
these cases is satisfied by a planted assignment, so a feasible solution with
one choice edge per gadget exists; the base inequality then makes every
minimum solution select exactly one such edge per gadget.  A group
$\{x_{i_1}^{b_1},\ldots,x_{i_k}^{b_k}\}$ is covered exactly when the Boolean
clause $(x_{i_1}=b_1)\lor\cdots\lor(x_{i_k}=b_k)$ is true.

\begin{figure}[H]
\centering
\begin{tikzpicture}[x=1.45cm,y=1cm]
  \node[graph vertex,label=below:$r$] (r) at (0,0) {};
  \node[graph vertex,label=below:$z_i$] (z) at (2,0) {};
  \node[graph vertex,label=right:$x_i^0$] (x0) at (4.2,0.9) {};
  \node[graph vertex,label=right:$x_i^1$] (x1) at (4.2,-0.9) {};
  \draw (r)--node[above,font=\scriptsize]{spoke} (z);
  \draw (z)--node[above,sloped,font=\scriptsize]{choice 0} (x0);
  \draw (z)--node[below,sloped,font=\scriptsize]{choice 1} (x1);
  \node[draw,dashed,rounded corners,fit=(z),inner sep=5pt] {};
  \node[font=\scriptsize] at (2,0.7) {singleton group};
  \node[draw,dashed,rounded corners,fit=(x0)(x1),inner sep=6pt] {};
  \node[font=\scriptsize] at (4.2,1.55) {choice group};
\end{tikzpicture}
\caption{The binary-choice gadget.  Root connectivity turns group coverage
into the selection of a truth-value edge.}
\label{fig:tier1-choice-gadget}
\end{figure}
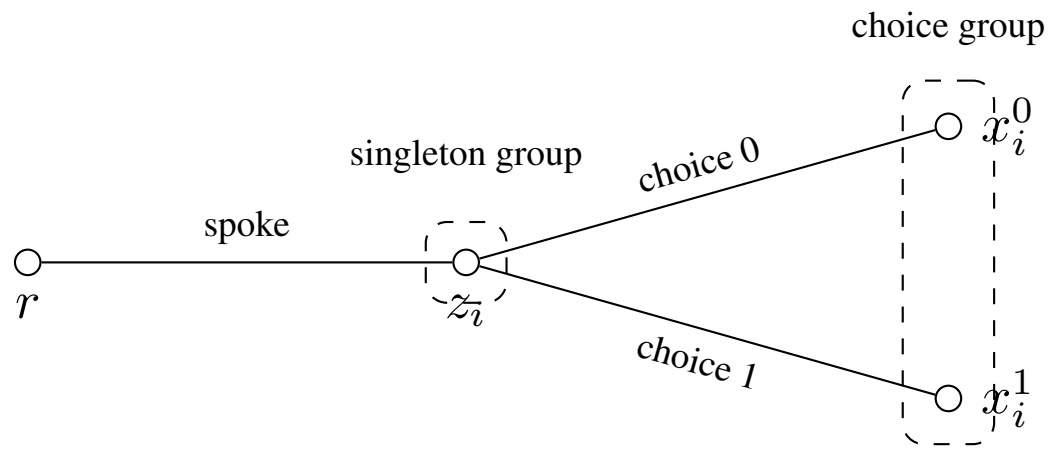

Every generated instance is finally subjected to a uniformly random vertex
permutation, which also transports the root.  The serialization orders of
edges, group records, and members within each group are independently
randomized.

\paragraph{C1.6.1 Mandatory-cut calibration.}
Take the complete 4-ary tree with its root at level zero and leaves at level
five.  It has 1,365 vertices, 1,364 tree edges, and 1,024 leaves; every leaf is
a singleton group.  Independently and uniformly sample every tree-edge weight
from $\{7,\ldots,29\}$ and let their sum be $B$.  Repeatedly sample an ordered
pair of distinct vertices uniformly, sort it into an unordered pair, and
reject it if it is a tree edge or was sampled earlier, until 1,600 non-tree
edges have been obtained.  Give each such edge weight
$B+1+U$, where $U$ is independently uniform in $\{0,\ldots,9999\}$.  Any
accepted solution therefore excludes all added edges, while reaching every
singleton leaf forces every tree edge.  The public graph has 2,964 edges and
the unique accepted edge set has 1,364 edges.

\paragraph{C1.6.2 Serial cactus boundary constraints.}
Chain 768 four-cycles by identifying the right articulation of one cycle with
the left articulation of the next, and take the first cell's left articulation
as $r$.  In cell $i$, denote the internal vertices of the two internally
disjoint two-edge routes by $t_i$ and $d_i$.  Independently and uniformly
sample all 3,072 edge weights from $\{3,\ldots,31\}$; the graph has 2,305
vertices.  Independently sample a fair planted mode
$s_i\in\{\mathrm{top},\mathrm{bottom}\}$ for each cell.  At boundary $i$, the
four candidate groups are
$\{u_i,v_{i+1}\}$ for $u,v\in\{t,d\}$.  Exclude the unique pair containing
neither planted branch vertex and uniformly choose two of the remaining three
groups without replacement.  Add the singleton group containing the final
articulation, giving 1,535 groups.

A connected solution traverses each cell in one of three modes.  The top and
bottom costs are the sums of the corresponding two route-edge weights.  The
both cost is the minimum of the two three-edge subgraphs obtained by taking
one entire route and the cheaper edge of the other route.  A three-state path
dynamic program discards adjacent mode pairs that miss either boundary group,
adds these local costs, and publishes its minimum as $B$.

\paragraph{C1.6.3 Relabelled bounded-frontier choices.}
Use 1,536 binary-choice gadgets.  Spoke weights lie in
$\{3,\ldots,31\}$; each choice-edge weight is
$43{,}009$ plus an offset in the same interval, with all these samples
independent and uniform.  Sample a planted assignment $\widehat x$ of 1,536
independent fair bits.  Add the 1,536 singleton center
groups and 1,536 choice groups, as well as pair groups
$\{z_i,z_{i+1}\}$ for $0\leq i<1535$; the latter are already hit by the forced
centers and encode the path $z_0,z_1,\ldots,z_{1535}$ up to reversal.  For
every right index $i>0$, uniformly choose without replacement exactly
$\min\{9,i\}$ indices from
$\{\max(0,i-12),\ldots,i-1\}$.  For each selected $j$, sample $(a,b)$ as two
independent fair bits; if it equals
$(\widehat x_j,\widehat x_i)$, replace $a$ by $1-a$.  Add
$\{x_j^{1-a},x_i^{1-b}\}$.  This group is missed exactly by
$(x_j,x_i)=(a,b)$.  There are 13,779 such constraint groups and 18,386 groups
overall.  The graph has 4,609 vertices and 4,608 edges.  A width-12 frontier
dynamic program scans $i$ from left to right, stores the preceding 12 values,
adds the spoke and selected choice-edge cost for $x_i$, and rejects a state
realizing any forbidden pair whose right endpoint is $i$.  Its minimum is
published as $B$.

\paragraph{C1.6.4 Higher-order bounded-frontier coverage.}
Use 1,280 choice gadgets, with choice base
$35{,}841=1280(31-3)+1$.  Spoke weights and choice-edge offsets are sampled
independently and uniformly from $\{3,\ldots,31\}$, and the planted assignment
consists of independent fair bits.  Groups $\{z_i,z_{i+1},z_{i+2}\}$ for
$0\leq i<1278$ impose no additional feasibility condition because their
centers are forced; they encode consecutive triples of variables.

For every $2\leq i<1280$, set
$P_i=\{\max(0,i-15),\ldots,i-1\}$ and generate
$\min\{7,2^{|P_i|}\}$ distinct clauses having rightmost variable $x_i$.
Uniformly choose the clause size from $\{3,4\}$, capped at $|P_i|+1$;
uniformly shuffle $P_i$ and use the first one fewer variables together with
$i$.  Give the selected variables independent fair literal values.  If none
matches the planted assignment, uniformly choose one literal and change its
value to the planted value; reject and resample if the same signed clause was
already generated for $i$.  The corresponding group contains $x_j^b$ for
every literal $(x_j=b)$.  The instance has 3,841 vertices, 3,840 edges, and
12,781 groups, including 8,943 clause groups.  A left-to-right exact transfer
program retains the previous 15 bits, checks all clauses ending at $i$, adds
the local spoke and choice cost, and publishes its minimum as $B$.

\paragraph{C1.6.5 Running-intersection bag-tree coverage.}
Number 121 bags in breadth-first order in the complete rooted ternary tree of
depth four, with parent $p(t)=\lfloor(t-1)/3\rfloor$ and child slot
$c(t)=(t-1)\bmod3$ for $t>0$.  Bag 0 contains variables $0,\ldots,15$.
Each non-root bag shares a six-variable separator with its parent and
introduces ten fresh variables.  If the parent is the root, child slot $c$
uses parent positions $6c,\ldots,6c+5$ modulo 16.  Otherwise write the parent
bag in its stored order as six inherited variables followed by ten introduced
variables.  Child slot $c$ takes inherited positions
$3c,3c+1,3c+2$ modulo six and introduced positions
$3c,3c+1,3c+2$ modulo ten.  Appending ten globally fresh variables gives the
child's stored order.  This rule ensures that the bags containing any fixed
variable form a connected subtree.

There are 1,216 binary-choice gadgets, with choice base
$34{,}049=1216(31-3)+1$.  Spoke weights and choice offsets are independent and
uniform in $\{3,\ldots,31\}$, and the planted assignment consists of fair
bits.  In every bag, independently choose a clause size uniformly from
$\{3,4\}$, uniformly shuffle the 16 variables and take the selected prefix,
give its variables fair literal values, and repair one uniformly selected
literal to its planted value if necessary.  Reject within-bag duplicate signed
clauses until the bag has ten.  For every bag $\beta$ add the center-only group
$\{z_i:x_i\in\beta\}$, and for every parent separator $S$ add
$\{z_i:x_i\in S\}$; forced center singletons make these groups redundant for
feasibility.

The graph has 3,649 vertices, 3,648 edges, and 3,883 groups: 1,216 singleton
centers, 1,216 choice groups, 121 bag groups, 120 separator groups, and 1,210
clause groups.  A bottom-up dynamic program enumerates the $2^{16}$ assignments
to each bag, conditions each child table on the six separator bits, and charges
a variable's choice cost only in the bag where it is introduced.  Its root
minimum is published as $B$.

\paragraph{C1.6.6 Deep shared-routing ancestor coverage.}
Start with the complete binary tree whose root $r$ is at level zero and whose
leaves are at level 11; it has 4,095 routing vertices and 4,094 routing edges.
Number its vertices in breadth-first order, so the parent of $v>0$ is
$\lfloor(v-1)/2\rfloor$.  Attach two terminal leaves
$x_v^0,x_v^1$ to every routing vertex $v$.  Routing-edge weights lie in
$\{3,\ldots,31\}$; terminal-edge weights are
$114{,}661=4095(31-3)+1$ plus an offset in that interval.  The terminal pair at
every owner is a group, forcing at least one terminal edge.  Reaching such a
leaf forces the unique root-to-owner route, so all routing edges occur in
every feasible solution; the common base makes exactly one terminal per owner
optimal.  All routing weights and terminal offsets are sampled independently
and uniformly, and every owner receives an independent fair planted bit.

For an owner $v$, list $v$ followed by its nearest ancestors, stopping after
ten vertices or at the root.  For each of four coverage groups, uniformly
shuffle this list, take its first $\min\{3,\operatorname{depth}(v)+1\}$ owners,
and, if $v$ is absent, replace the first selected owner by $v$.  Independently
sample one fair terminal value for each selected owner; if none agrees with
the planted bits, choose one selected position uniformly and replace its value
by the corresponding planted bit.  The resulting terminals form the group;
different generated groups are retained even when their member sets coincide.

The resulting graph has 12,285 vertices, 12,284 edges, and 20,475 groups.
Every edge set accepted at the optimum budget contains all 4,094 routing edges
and exactly one of the two terminal edges at each owner, for 4,095 terminal
edges in total.  A bottom-up dynamic program conditions each subtree on the
values of its nine closest ancestors, checks its owner's four groups, and
publishes the minimum routing-plus-terminal cost as $B$.

\clearpage
\section{Tier-2: Interactive Proofs}
\label{app:tier2}

Tier-2 contains interactive tasks in which Merlin must commit to a message
before Arthur samples the fresh randomness used to check it.  The tasks
documented here use hidden graph relabelings, the Sum-Check protocol, the GKR
reduction for layered arithmetic circuits, affine hashing of circuit images,
Shamir-style arithmetization of quantified Boolean formulas, and a
bind--challenge--open proof of Hamiltonicity.  Each run is subject to a
uniform 600-second wall-clock budget.

\subsection{Graph Non-Isomorphism}
\label{sec:gni}

Graph Non-Isomorphism (GNI) asks whether two graphs have different structures
even after their vertices are renamed.  Interaction turns this decision
problem into a source-identification game: Arthur secretly chooses one of two
public non-isomorphic graphs, randomly relabels it, and asks Merlin which source
was used.  The task is a parallel repetition of the classical GNI protocol
\cite{goldreich-micali-wigderson-1991}.

A vertex-colored undirected simple graph is a triple
\[
  G=([N],E,\chi),
  \qquad [N]=\{0,1,\ldots,N-1\},
\]
where $E\subseteq\binom{[N]}2$ and
$\chi:[N]\to\mathbb Z_{\geq0}$.  A color-preserving isomorphism from $G$ to
$G'$ is a bijection $\varphi:[N]\to[N]$ such that
\[
  \{u,v\}\in E(G)\iff\{\varphi(u),\varphi(v)\}\in E(G'),
  \qquad \chi(u)=\chi'(\varphi(u)).
\]
We write $G\iso G'$ when such a bijection exists.  The protocol is defined for
two same-order public graphs $G_0,G_1$.  Perfect completeness is evaluated when
$G_0\not\iso G_1$, whereas soundness is evaluated when $G_0\iso G_1$.  All
seven generated pairs are single-colored and non-isomorphic, and
their non-isomorphism follows from the construction-specific invariants stated
below.

Merlin first sends a fixed, content-independent start signal.  Arthur then
creates 24 independent challenge lanes.  In lane $j$ he samples
\[
  \beta_j\xleftarrow{\$}\{0,1\},\qquad
  \pi_j\xleftarrow{\$}S_N,\qquad
  H_j=\pi_j(G_{\beta_j}),
\]
where $\pi_j$ transports adjacency and colors.  The bit $\beta_j$ has no graph-
theoretic meaning beyond identifying the chosen source.  Arthur returns the 24
graphs $H_j$, after which Merlin submits one ordered vector
$(\widehat\beta_0,\ldots,\widehat\beta_{23})$, and Arthur accepts exactly when
$\widehat\beta_j=\beta_j$ for every lane.  Thus the protocol has two Merlin
messages and one intervening private-coin challenge.

\begin{lemma}[Perfect completeness and soundness]
\label{lem:tier2-gni}
If $G_0\not\iso G_1$, an unbounded honest Merlin is accepted with probability
$1$.  If the two sources in the soundness experiment are isomorphic, every
Merlin is accepted with probability at most $2^{-24}$.
\end{lemma}

\begin{proof}
For non-isomorphic sources, the color-preserving isomorphism orbits of $G_0$
and $G_1$ are disjoint, so every challenge graph determines a unique source.
If the sources are isomorphic, a uniformly relabeled copy has the same
distribution for $\beta_j=0$ and $\beta_j=1$.  Conditional on the entire
challenge vector $(H_0,\ldots,H_{23})$, the source vector
$(\beta_0,\ldots,\beta_{23})$ is therefore uniform on $\{0,1\}^{24}$, giving
the bound $2^{-24}$ even when Merlin processes all lanes jointly.
\end{proof}

Every case is first constructed in logical coordinates.  Before publication,
the two source graphs are independently relabeled by seed-derived uniform
permutations, and their order is swapped with probability $1/2$.  These
transformations erase the generator coordinates without changing isomorphism
type.

\paragraph{C2.1.1 Connected prism--M\"obius baseline.}
An odd integer $r$ is sampled uniformly from the odd values in $[701,1001]$.
The first source is the circular prism
\[
  P_r=C_r\mathbin\square K_2,
\]
where the Cartesian product $X\square Y$ has vertex set $V(X)\times V(Y)$ and
joins $(x,y)$ to $(x',y')$ exactly when either $x=x'$ and $yy'\in E(Y)$, or
$y=y'$ and $xx'\in E(X)$.  The second source is the M\"obius ladder on
$\mathbb Z_{2r}$,
\[
  E(M_{2r})=E(C_{2r})\cup
  \bigl\{\{i,i+r\}:0\leq i<r\bigr\}.
\]
Both sources are connected cubic graphs with $2r$ vertices and $3r$ edges.
The prism contains an odd rim $C_r$ and is not bipartite.  Parity of the index
gives a bipartition of $M_{2r}$ because both a cycle edge and an antipodal chord
change parity when $r$ is odd.

\begin{figure}[H]
  \centering
  \begin{tikzpicture}[x=1cm,y=1cm]
    \node[diagram label] at (-3.1,1.72) {(a) $P_5=C_5\square K_2$};
    \foreach \i in {0,...,4} {
      \coordinate (po\i) at ({-3.1+1.12*cos(90-72*\i)},
                              {1.12*sin(90-72*\i)});
      \coordinate (pi\i) at ({-3.1+0.58*cos(90-72*\i)},
                              {0.58*sin(90-72*\i)});
      \node[graph vertex] at (po\i) {};
      \node[graph vertex] at (pi\i) {};
      \draw[diagram edge] (po\i)--(pi\i);
    }
    \foreach \i in {0,...,4} {
      \pgfmathtruncatemacro{\j}{mod(\i+1,5)}
      \draw[diagram edge] (po\i)--(po\j);
      \draw[diagram edge] (pi\i)--(pi\j);
    }
    \node[diagram label] at (3.0,1.72) {(b) $M_{10}$};
    \foreach \i in {0,...,9} {
      \coordinate (mo\i) at ({3.0+1.22*cos(90-36*\i)},
                              {1.22*sin(90-36*\i)});
      \node[graph vertex] at (mo\i) {};
    }
    \foreach \i in {0,...,9} {
      \pgfmathtruncatemacro{\j}{mod(\i+1,10)}
      \draw[diagram edge] (mo\i)--(mo\j);
    }
    \foreach \i in {0,...,4} {
      \pgfmathtruncatemacro{\j}{\i+5}
      \draw[diagram edge] (mo\i)--(mo\j);
    }
  \end{tikzpicture}
  \caption{The two C2.1.1 constructions at the illustrative value $r=5$;
  generated instances use an odd $r\in[701,1001]$.}
  \label{fig:prism-moebius}
\end{figure}
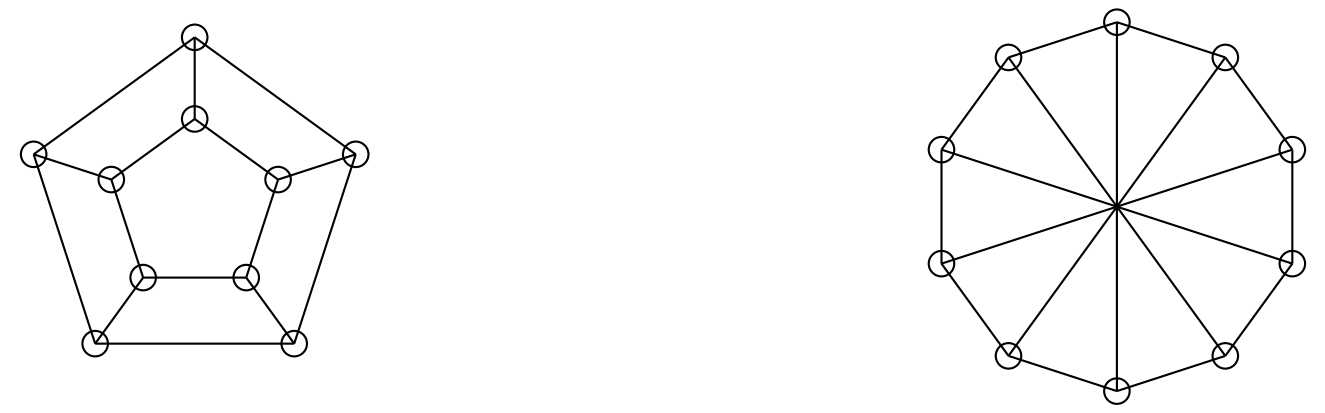

\paragraph{C2.1.2 Connected Shrikhande--rook amplification.}
The Shrikhande graph $S$ has vertex set $\mathbb Z_4^2$, with $x$ adjacent to
$y$ when
\[
  y-x\in\{\pm(1,0),\ \pm(0,1),\ \pm(1,1)\}.
\]
The $4\times4$ rook graph $R_{4,4}$ has the board cells as vertices and joins
two cells in a common row or column.  Both graphs are strongly regular with
parameters $\operatorname{srg}(16,6,2,2)$
\cite{brouwer-haemers-2012}, but $R_{4,4}$ contains row and column copies of
$K_4$, whereas the largest clique in $S$ has size three.

The generator samples $q$ uniformly from the odd integers in $[89,127]$ and
publishes
\[
  S\mathbin\square C_q
  \qquad\text{and}\qquad
  R_{4,4}\mathbin\square C_q.
\]
They are connected 8-regular graphs with $16q$ vertices and $64q$ edges.  A
clique in a Cartesian product lies in a single factor fiber; hence the rook
product contains a $K_4$ and the Shrikhande product does not.  For every vertex
$v$, the two induced neighborhood graphs are
\[
  (S\square C_q)[N(v)]\cong C_6\sqcup2K_1,
  \qquad
  (R_{4,4}\square C_q)[N(v)]\cong2K_3\sqcup2K_1,
\]
where the two isolated vertices are the neighbors in the cycle direction.

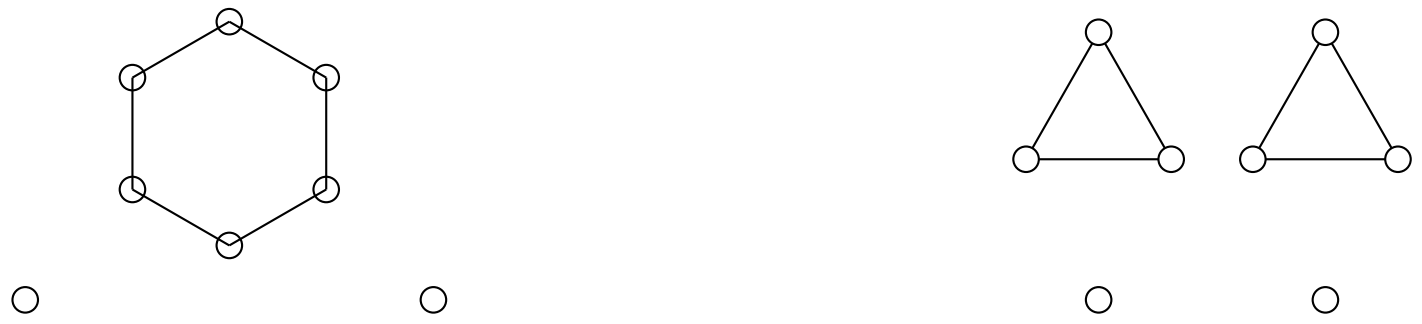
\begin{figure}[H]
  \centering
  \begin{tikzpicture}[x=1cm,y=1cm]
    \node[diagram label] at (-3.4,1.55)
      {$S\square C_q:\ G[N(v)]\cong C_6\sqcup2K_1$};
    \foreach \i in {0,...,5} {
      \coordinate (sn\i) at ({-3.4+0.74*cos(90-60*\i)},
                              {0.35+0.74*sin(90-60*\i)});
      \node[graph vertex] at (sn\i) {};
    }
    \foreach \i in {0,...,5} {
      \pgfmathtruncatemacro{\j}{mod(\i+1,6)}
      \draw[diagram edge] (sn\i)--(sn\j);
    }
    \node[graph vertex] at (-4.75,-0.75) {};
    \node[graph vertex] at (-2.05,-0.75) {};

    \node[diagram label] at (3.1,1.55)
      {$R_{4,4}\square C_q:\ G[N(v)]\cong2K_3\sqcup2K_1$};
    \foreach \cx in {2.35,3.85} {
      \draw[diagram edge]
        (\cx,1.02)--({\cx-0.48},0.18)--({\cx+0.48},0.18)--cycle;
      \node[graph vertex] at (\cx,1.02) {};
      \node[graph vertex] at ({\cx-0.48},0.18) {};
      \node[graph vertex] at ({\cx+0.48},0.18) {};
    }
    \node[graph vertex] at (2.35,-0.75) {};
    \node[graph vertex] at (3.85,-0.75) {};
  \end{tikzpicture}
  \caption{The subgraph induced by the eight neighbors of a selected vertex
  in C2.1.2.  The selected vertex and its incident edges are omitted.}
  \label{fig:shrikhande-rook-neighborhoods}
\end{figure}

\paragraph{C2.1.3 Locally matched group-table geometries.}
For a finite Abelian group $H$, form its addition table and let $L(H)$ be the
Latin-square graph whose vertices are cells $(x,y)\in H^2$.  Two distinct
cells are adjacent when they have the same row, the same column, or the same
symbol $x+y$.  The sources use
\[
  H_0=C_{16}\times C_2,
  \qquad H_1=C_8\times C_4.
\]
Here $C_m=\mathbb Z/m\mathbb Z$ denotes the cyclic group, rather than a cycle
graph.  The construction is deterministic before the final random relabeling.
Each graph has 1,024 vertices, degree $3(32-1)=93$, and 47,616 edges; both have
strongly regular parameters $\operatorname{srg}(1024,93,32,6)$.  The maximal
32-cliques recover the row, column, and symbol classes, up to a permutation of
these three families.  Permuting the three roles is a \emph{parastrophe};
independently renaming the row, column, and symbol labels is an \emph{isotopy}.
Thus a graph isomorphism would recover the two group tables up to parastrophe
and isotopy.  Parastrophes of a group table are isotopic to the group, while
isotopic groups are isomorphic \cite{albert-1943}.

The group element-order distributions are
\[
\begin{array}{c|rrrrr}
 &1&2&4&8&16\\ \hline
 C_{16}\times C_2&1&3&4&8&16\\
 C_8\times C_4&1&3&12&16&0
\end{array}
\]
If the two Latin-square graphs were isomorphic, the preceding recovery would
make their group tables isotopic and hence the groups isomorphic, contradicting
the different element-order distributions.  The two graphs are therefore
non-isomorphic.

\paragraph{Common CFI construction for C2.1.4--C2.1.7.}
The last four cases use the Cai--F\"urer--Immerman (CFI) construction
\cite{cai-furer-immerman-1992}.  Its starting graph $Q$ is the incidence graph
of the generalized quadrangle $GQ(2,2)$ \cite{payne-thas-2009}.  One bipartite
class consists of the 15 unordered pairs from a six-element set and the other
of the 15 partitions of that set into three pairs; a pair is adjacent to a
partition precisely when it is one of its parts.  Thus $Q$ is connected,
cubic, bipartite, and has 30 vertices, 45 edges, and girth eight.

\begin{figure}[H]
  \centering
  \begin{tikzpicture}[x=1cm,y=1cm]
    \node[outer vertex] (p23) at (-3.0,1.40) {};
    \node[outer vertex] (p45) at (-3.0,0.70) {};
    \node[outer vertex] (p01) at (-3.0,0.00) {};
    \node[outer vertex] (p24) at (-3.0,-0.70) {};
    \node[outer vertex] (p35) at (-3.0,-1.40) {};
    \node[font=\scriptsize,left=2pt] at (p01.west) {$\{0,1\}$};
    \node[font=\scriptsize,left=2pt] at (p23.west) {$\{2,3\}$};
    \node[font=\scriptsize,left=2pt] at (p45.west) {$\{4,5\}$};
    \node[font=\scriptsize,left=2pt] at (p24.west) {$\{2,4\}$};
    \node[font=\scriptsize,left=2pt] at (p35.west) {$\{3,5\}$};
    \node[draw,rectangle,inner sep=4pt,font=\small] (partA) at (2.0,0.70)
      {$01\mid23\mid45$};
    \node[draw,rectangle,inner sep=4pt,font=\small] (partB) at (2.0,-0.70)
      {$01\mid24\mid35$};
    \draw[diagram edge] (p01)--(partA);
    \draw[diagram edge] (p23)--(partA);
    \draw[diagram edge] (p45)--(partA);
    \draw[diagram edge] (p01)--(partB);
    \draw[diagram edge] (p24)--(partB);
    \draw[diagram edge] (p35)--(partB);
  \end{tikzpicture}
  \caption{Two partition vertices of the $GQ(2,2)$ incidence graph.  A
  rectangle is adjacent to the three pair vertices appearing in its
  partition; the full construction contains all 15 vertices of each type.}
  \label{fig:gq-incidence}
\end{figure}
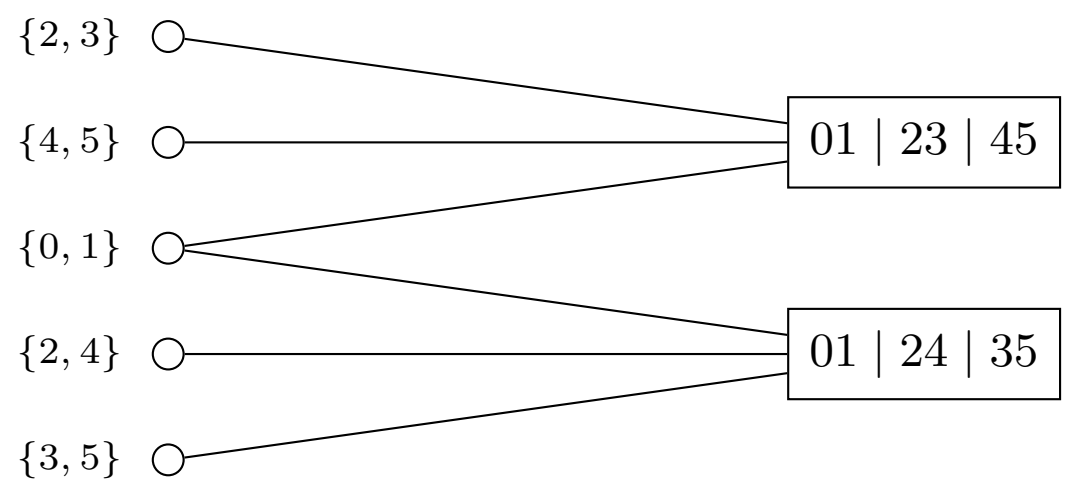

Index the vertices of $Q$ by listing the 15 pairs and the 15 partitions in
lexicographic order, and index its edges lexicographically.  For a lift factor
$f$, orient every edge from its smaller endpoint to its larger endpoint and
assign it a voltage $z_{uv}\in\mathbb Z_f$, with $z_{vu}=-z_{uv}$.  The cyclic
lift $B_f$ has
\[
  V(B_f)=V(Q)\times\mathbb Z_f,
  \qquad (u,a)\sim(v,a+z_{uv}).
\]
The spanning tree is obtained by breadth-first search from vertex zero, visiting
neighbors in increasing order.  Tree edges receive voltage zero; one non-tree
edge is selected uniformly and receives voltage one; every other non-tree edge
receives an independent uniform element of $\mathbb Z_f$.  All choices are
drawn from the instance's seeded random stream.  The voltage-one fundamental
cycle makes the lift connected.  Consequently $B_f$ is a connected cubic graph
with $30f$ vertices, $45f$ edges, and girth at least eight.

Each vertex of $B_f$ is replaced by a ten-vertex gadget.  Its three incident
base edges, ordered by their edge indices, correspond to three \emph{port pairs}
$\{i{:}0,i{:}1\}$, $i\in\{1,2,3\}$; a port is an outer vertex through which
the gadget connects to a neighboring gadget.  Four middle vertices are indexed
by the even-parity words $000,011,101,110$, and word $a_1a_2a_3$ is adjacent
to $i{:}a_i$ for every $i$.  Across a base edge $e=uv$, a twist bit
$\tau_e$ determines a two-edge matching as follows.  If $e$ has local
incidence rank $i_u(e)$ at $u$ and $i_v(e)$ at $v$, add exactly
\[
  \bigl\{\{(u,i_u(e),b),(v,i_v(e),b\oplus\tau_e)\}:b\in\{0,1\}\bigr\}.
\]
Thus ``equal'' and ``opposite'' refer only to the binary labels within the two
endpoint-specific port pairs; the local incidence ranks need not agree.

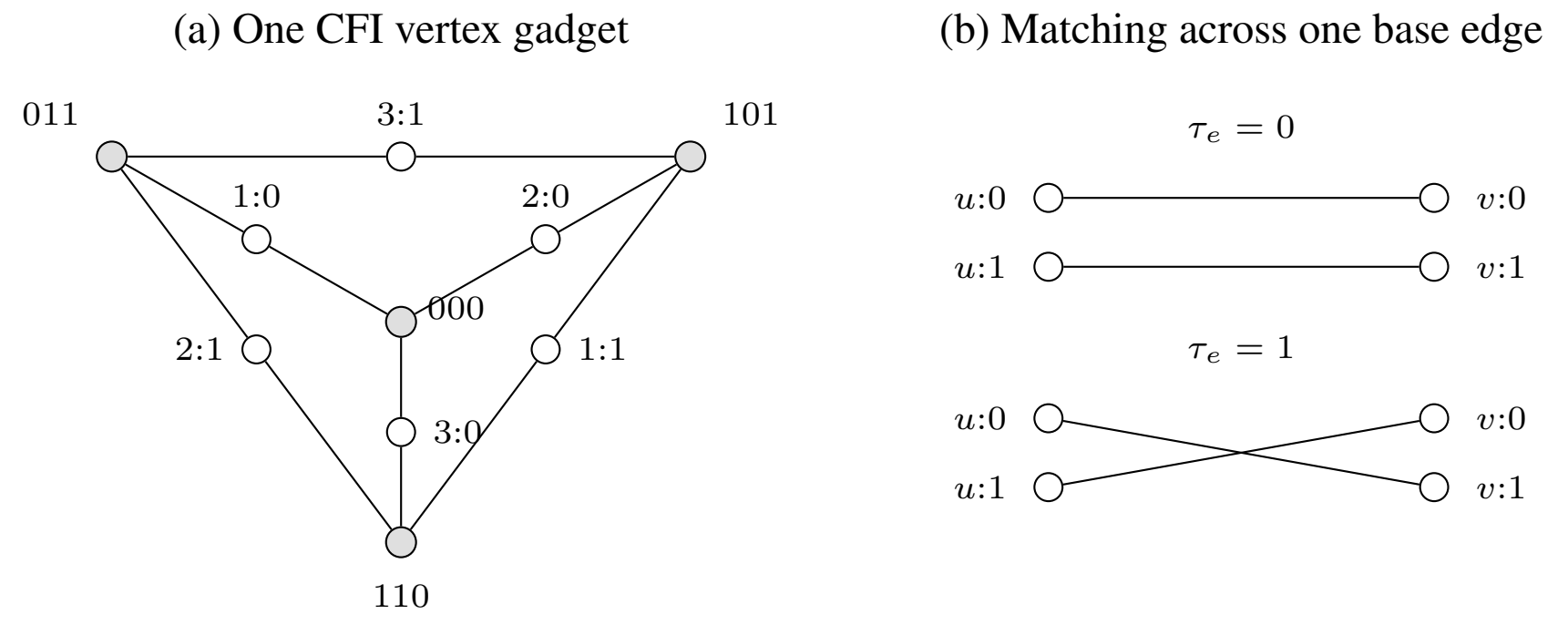
\begin{figure}[H]
  \centering
  \begin{tikzpicture}[x=1cm,y=1cm]
    \node[diagram label] at (-3.5,2.25) {(a) One CFI vertex gadget};
    \node[middle vertex] (m000) at (-3.5,0.15) {};
    \node[middle vertex] (m011) at (-5.6,1.35) {};
    \node[middle vertex] (m101) at (-1.4,1.35) {};
    \node[middle vertex] (m110) at (-3.5,-1.45) {};
    \node[font=\scriptsize] at (-3.10,0.25) {$000$};
    \node[font=\scriptsize,above left=1pt] at (m011.north west) {$011$};
    \node[font=\scriptsize,above right=1pt] at (m101.north east) {$101$};
    \node[font=\scriptsize,below=2pt] at (m110.south) {$110$};
    \node[outer vertex,label={[font=\scriptsize]above:$1{:}0$}]
      (o10) at (-4.55,0.75) {};
    \node[outer vertex,label={[font=\scriptsize]right:$1{:}1$}]
      (o11) at (-2.45,-0.05) {};
    \node[outer vertex,label={[font=\scriptsize]above:$2{:}0$}]
      (o20) at (-2.45,0.75) {};
    \node[outer vertex,label={[font=\scriptsize]left:$2{:}1$}]
      (o21) at (-4.55,-0.05) {};
    \node[outer vertex,label={[font=\scriptsize]right:$3{:}0$}]
      (o30) at (-3.5,-0.65) {};
    \node[outer vertex,label={[font=\scriptsize]above:$3{:}1$}]
      (o31) at (-3.5,1.35) {};
    \foreach \a/\b in {m000/o10,m011/o10,m101/o11,m110/o11,
                         m000/o20,m101/o20,m011/o21,m110/o21,
                         m000/o30,m110/o30,m011/o31,m101/o31}
      \draw[diagram edge] (\a)--(\b);

    \node[diagram label] at (2.6,2.25) {(b) Matching across one base edge};
    \node[font=\scriptsize] at (2.6,1.55) {$\tau_e=0$};
    \node[outer vertex] (u00) at (1.2,1.05) {};
    \node[outer vertex] (u01) at (1.2,0.55) {};
    \node[outer vertex] (v00) at (4.0,1.05) {};
    \node[outer vertex] (v01) at (4.0,0.55) {};
    \node[font=\scriptsize,left=2pt] at (u00.west) {$u{:}0$};
    \node[font=\scriptsize,left=2pt] at (u01.west) {$u{:}1$};
    \node[font=\scriptsize,right=2pt] at (v00.east) {$v{:}0$};
    \node[font=\scriptsize,right=2pt] at (v01.east) {$v{:}1$};
    \draw[diagram edge] (u00)--(v00);
    \draw[diagram edge] (u01)--(v01);
    \node[font=\scriptsize] at (2.6,-0.05) {$\tau_e=1$};
    \node[outer vertex] (u10) at (1.2,-0.55) {};
    \node[outer vertex] (u11) at (1.2,-1.05) {};
    \node[outer vertex] (v10) at (4.0,-0.55) {};
    \node[outer vertex] (v11) at (4.0,-1.05) {};
    \node[font=\scriptsize,left=2pt] at (u10.west) {$u{:}0$};
    \node[font=\scriptsize,left=2pt] at (u11.west) {$u{:}1$};
    \node[font=\scriptsize,right=2pt] at (v10.east) {$v{:}0$};
    \node[font=\scriptsize,right=2pt] at (v11.east) {$v{:}1$};
    \draw[diagram edge] (u10)--(v11);
    \draw[diagram edge] (u11)--(v10);
  \end{tikzpicture}
  \caption{The CFI gadget and its straight and crossed port matchings.  The
  three-bit label of a gray vertex specifies its neighbor in every port pair.}
  \label{fig:cfi-gadget}
\end{figure}

For $b\in\{0,1\}$, sample the first $|E(B_f)|-1$ twist bits independently and
uniformly, and set the final bit so that
\[
  \bigoplus_{e\in E(B_f)}\tau^{(b)}_e=b.
\]
This samples $\tau^{(b)}$ uniformly from the parity-$b$ affine subspace.  Let
$R_f^{(b)}$ denote the resulting CFI graph over $B_f$; the two vectors
$\tau^{(0)}$ and $\tau^{(1)}$ are sampled independently.
No outer/middle colors are retained.  Nevertheless, the gadget decomposition
is intrinsic.  Form the auxiliary graph on the same vertex set by joining two
vertices exactly when they have a common neighbor in the CFI graph.  The four
middle vertices of each gadget are exactly the maximal $K_4$ sets that recover
the gadgets in this auxiliary graph;
their six neighbors in the original graph split into three complementary port
pairs.  Choose an arbitrary order of the three pairs and an
arbitrary $0/1$ orientation within each pair.  The four recovered middle words
then have one common parity $p_v\in\F_2$.  For a reconstructed base edge $e$,
let $h_e\in\F_2$ be the parity of its two-edge matching in these local frames.
The quantity
\[
  I(X)=\bigoplus_v p_v\ \oplus\!\bigoplus_e h_e
\]
is independent of all frame choices: reversing one port pair flips exactly one
$p_v$ and one $h_e$, while permuting the three pairs preserves parity.  In the
generated frame, $p_v=0$ and $h_e=\tau_e$, so $I(X)=\bigoplus_e\tau_e$.
The opposite-parity CFI roots are therefore non-isomorphic.  A root with lift
factor $f$ has $300f$ vertices, $450f$ edges, and degree three.

\paragraph{C2.1.4 Uncolored enlarged CFI parity.}
This case publishes $R_{12}^{(0)}$ and $R_{12}^{(1)}$ without a further graph
transformation.
Each source is connected, single-colored, and cubic, with 3,600 vertices and
5,400 edges.  The unordered pair of intrinsic invariant values is $\{0,1\}$.

\paragraph{C2.1.5 Fisher-transformed uncolored CFI parity.}
For $b\in\{0,1\}$, first subdivide every edge of $R_{10}^{(b)}$ once and then
take the line graph.  Thus the two published graphs are obtained by applying
\[
  F(R)=L(S(R)).
\]
The vertices of a line graph $L(X)$ are the edges of $X$, and two are adjacent
when their corresponding edges share an endpoint.  In $F(R)$, the three
incidences at every cubic vertex of $R$ form a triangle, and the two incidences
of every original edge are joined by one triangle-external edge.  Every output
vertex lies in exactly one such triangle and exactly one external matching
edge.  These are exactly the triangles of $F(R)$, so they are intrinsically
identifiable; contracting each triangle recovers $R$.  Each published graph is
connected, single-colored, cubic, and has 9,000 vertices and 13,500 edges.

\begin{figure}[H]
  \centering
  \begin{tikzpicture}[x=1cm,y=1cm]
    \node[diagram label] at (-3.8,1.65) {a cubic root edge $uv$};
    \node[graph vertex,label=left:$u$] (ru) at (-5.0,0.45) {};
    \node[graph vertex,label=right:$v$] (rv) at (-2.6,0.45) {};
    \draw[diagram edge] (ru)--(rv);
    \draw[diagram edge] (ru)--(-5.8,1.15);
    \draw[diagram edge] (ru)--(-5.8,-0.25);
    \draw[diagram edge] (rv)--(-1.8,1.15);
    \draw[diagram edge] (rv)--(-1.8,-0.25);

    \draw[diagram arrow] (-0.9,0.45)--(0.35,0.45);
    \node[diagram label] at (3.1,1.65) {the corresponding Fisher fragment};
    \draw[diagram edge] (1.2,0.95)--(2.0,0.15)--(1.2,-0.65)--cycle;
    \draw[diagram edge] (4.2,0.95)--(3.4,0.15)--(4.2,-0.65)--cycle;
    \foreach \x/\y in {1.2/0.95,2.0/0.15,1.2/-0.65,
                         4.2/0.95,3.4/0.15,4.2/-0.65}
      \node[graph vertex] at (\x,\y) {};
    \draw[very thick] (2.0,0.15)--(3.4,0.15);
  \end{tikzpicture}
  \caption{Local form of the Fisher transform in C2.1.5.  Each root vertex
  becomes a triangle; each root edge becomes one edge between two triangles.}
  \label{fig:fisher-transform}
\end{figure}
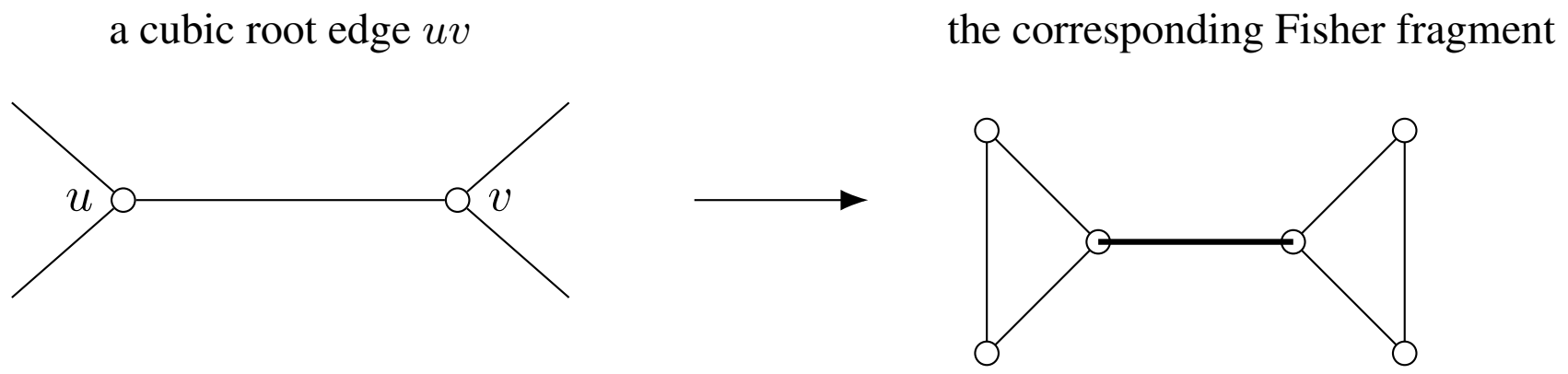

\paragraph{C2.1.6 Line graph of uncolored CFI parity.}
This case publishes $L(R_{16}^{(0)})$ and $L(R_{16}^{(1)})$.  Each CFI root is
triangle-free: the gadget is bipartite internally, and an inter-gadget edge
cannot close a three-cycle.  Hence the maximal $K_3$ subgraphs of $L(R)$ are
exactly the three root edges incident with a common root vertex.  Reconstructing
these stars recovers $R$; equivalently, uniqueness follows from Whitney's
line-graph theorem \cite{whitney-1932}, with neither root belonging to its
$K_3/K_{1,3}$ exceptional pair.  Each source is connected, single-colored, and
4-regular, with 7,200 vertices and 14,400 edges.

\paragraph{C2.1.7 Double line graph of uncolored CFI parity.}
This case publishes $L(L(R_{12}^{(0)}))$ and $L(L(R_{12}^{(1)}))$.  Since
$L(R)$ is 4-regular, the maximal $K_4$ subgraphs of $L(L(R))$ are exactly the
four-edge stars of $L(R)$ and recover that graph up to isomorphism.  Since the
root $R$ is cubic and triangle-free, the maximal $K_3$ subgraphs of the
recovered $L(R)$ are exactly the three-edge stars of $R$ and recover $R$.
Thus two applications of line-graph inversion determine the opposite-parity
roots up to isomorphism.  Each source is connected, single-colored, and
6-regular, with 10,800 vertices and 32,400 edges.

\newpage
\subsection{Boolean-Hypercube Sum-Check}
\label{sec:sumcheck}

This task asks Merlin to compute a weighted sum over all Boolean assignments to
a public factor model and then prove that sum by interaction.  The model first
specifies Boolean variables and local tables, each depending on only a few
variables.  Multiplying the tables gives a weight for every global assignment;
the desired quantity is the sum of those weights.  Directly enumerating the
Boolean cube would require at least $2^{288}$ assignments in the generated
cases.  The Sum-Check protocol instead reduces the claimed sum to one public
polynomial evaluation by fixing the variables one at a time
\cite{lund-fortnow-karloff-nisan-1992}.

All arithmetic is over
\[
  \F_p,\qquad p=2^{61}-1,
\]
where $p$ is prime.
The public instance contains variables $x_0,\ldots,x_{m-1}$ and a nonempty
family of factors.  Factor $a$ has an ordered scope
$S_a=(s_{a,0},\ldots,s_{a,k_a-1})$ of distinct variable indices and a table
$f_a:\{0,1\}^{k_a}\to\F_p$.  Its multilinear extension is
\[
  \widetilde f_a(y)=
  \sum_{b\in\{0,1\}^{k_a}} f_a(b)
  \prod_{j=0}^{k_a-1}
  \bigl(b_jy_j+(1-b_j)(1-y_j)\bigr).
\]
We refer to $\widetilde f_a$ as the multilinear extension (MLE) of $f_a$.
Define
\[
  g(x)=\prod_a\widetilde f_a(x|_{S_a}),
  \qquad
  Z=\sum_{b\in\{0,1\}^m}g(b)\pmod p.
  \tag{1}
\]
On Boolean inputs, $\widetilde f_a$ agrees with its table, so (1) is precisely
the partition function of the factor model.  In all seven cases every variable
occurs in at most five factors; hence $g$ has individual degree at most five.

Merlin first chooses an arbitrary permutation
$\sigma=(\sigma_0,\ldots,\sigma_{m-1})$ of the variables and claims $Z$.  Write
$z_i=x_{\sigma_i}$ and define
\[
  g_\sigma(z_0,\ldots,z_{m-1})=g(x_0,\ldots,x_{m-1}),
  \qquad x_{\sigma_i}=z_i.
\]
After challenges
$r_0,\ldots,r_{i-1}$ have been fixed, round $i$ concerns the univariate
polynomial
\[
  q_i(T)=
  \sum_{z_{i+1},\ldots,z_{m-1}\in\{0,1\}}
  g_\sigma(r_0,\ldots,r_{i-1},T,z_{i+1},\ldots,z_{m-1}).
  \tag{2}
\]
Merlin submits the six coefficients of $q_i$ in increasing-power order,
including trailing zeros.  Arthur checks
\[
  q_i(0)+q_i(1)=C_i,
\]
where $C_0$ is the claimed sum and
$C_i=q_{i-1}(r_{i-1})$ thereafter.  Only after this message has been recorded
does Arthur sample a fresh uniform $r_i\in\F_p$ and set
$C_{i+1}=q_i(r_i)$.  After $m$ rounds, Arthur evaluates every public factor MLE
at the assembled random point and accepts exactly when
\[
  q_{m-1}(r_{m-1})=g_\sigma(r_0,\ldots,r_{m-1}).
\]
The claimed value is not compared with a stored answer; acceptance follows
solely from these algebraic checks.

\begin{lemma}[Perfect completeness and soundness]
\label{lem:tier2-sumcheck}
An honest Merlin claiming the value in (1) is accepted with probability $1$.
For a false initial claim, the acceptance probability is at most
\[
  \frac{5m}{p}\leq\frac{2000}{2^{61}-1}<2^{-50}.
\]
\end{lemma}

\begin{proof}
For an honest transcript, (2) is the exact remaining Boolean sum.  Summing its
two Boolean endpoints restores the preceding claim, and after the final round
no variables remain, so the terminal value is $g(r_0,\ldots,r_{m-1})$.
For soundness, whenever the submitted polynomial differs from the true round
polynomial, their nonzero difference has degree at most five and vanishes at
the fresh challenge with probability at most $5/p$.  If it does not vanish, a
false claim persists to the next round and is exposed by the final public
evaluation.  A union bound over $m\leq400$ rounds gives the stated result.
\end{proof}

The generators first build a logical factor model and then use the seeded random
stream to apply four value-preserving transformations: a uniform permutation of
the variable names, an independent uniform permutation of each factor's axes
with the table reordered accordingly, independent uniform gauge bits
$h_v\in\{0,1\}$ applied consistently as $x_v\mapsto x_v\oplus h_v$, and a
uniform permutation of the factor list.  Each operation is a reparameterization
of the Boolean cube, so it preserves $Z$.
The descriptions below give the logical coordinates before this hiding step.

\paragraph{C2.2.1 Rank-one factors on a hidden 4-regular graph.}
On $m=384$ variables, the generator takes the union of two edge-disjoint
Hamilton cycles.  The first is the cyclic order $(0,1,\ldots,m-1)$; the second
is obtained by repeatedly sampling a uniform vertex permutation until its cycle
is simple and edge-disjoint from the first.  The result is a simple 4-regular
graph with 768 edges.  Every edge $e=\{u,v\}$, with $u<v$, receives the binary
table
\[
  f_e(x_u,x_v)=a_{e,u}(x_u)a_{e,v}(x_v),
  \qquad
  f_e(0,0)f_e(1,1)=f_e(1,0)f_e(0,1).
\]
The four vector coordinates $a_{e,u}(0),a_{e,u}(1),a_{e,v}(0),a_{e,v}(1)$ are
sampled independently and uniformly from $\{1,\ldots,7\}\subset\F_p^*$.
There are 768 binary factors and no other factors, so every variable occurs
four times.  The product consequently separates by variables,
\[
  g(x)=\prod_v A_v(x_v),
  \qquad
  A_v(T)=\prod_{e\ni v}a_{e,v}(T),
\]
and $Z=\prod_v(A_v(0)+A_v(1))$.  Axis permutations and Boolean gauges preserve
the rank-one identity.

\paragraph{C2.2.2 Weighted signed-equivalence model.}
This case uses an independently generated instance of the same two-cycle
construction on $m=360$ variables, with 720 binary factors.  Independent
uniform bits $s_v$ are drawn in logical vertex order.
For every edge $uv$, the binary table has nonzero entries exactly on the two
assignments satisfying
\[
  x_u\oplus x_v=s_u\oplus s_v,
\]
with independent uniform weights from $\{1,\ldots,7\}$ on the two allowed
entries.  A unary table whose two entries are sampled independently and
uniformly from $\{1,\ldots,7\}$ is attached to every variable.  Thus the model
has 360 unary and 720 binary factors, and every variable occurs five times.
Because the two-cycle graph is connected, the common support of the binary
tables consists of exactly two global assignments, $x=s$ and
$x=\mathbf 1\oplus s$.

\paragraph{C2.2.3 Cylindrical two-spin model of width 16.}
The logical interaction graph is the Cartesian product
$C_{16}\square P_{25}$.  Its Boolean spins are $x_{r,c}$ for
$r\in\mathbb Z_{16}$ and $0\leq c<25$; edges join $(r,c)$ to
$(r+1\bmod16,c)$ and, when $c<24$, to $(r,c+1)$.  Each unary table has two
independent uniform entries in $\{1,2,3\}$, and each binary table has four such
entries, with an edge's smaller variable index used as its first axis.  There
are 400
cycle-direction edges and $16(25-1)=384$ longitudinal edges, giving 400 unary
and 784 binary factors.  Boundary spins occur in four factors and all other
spins in five.

\begin{figure}[H]
  \centering
  \begin{tikzpicture}[x=1cm,y=0.78cm]
    \foreach \c in {0,...,5} {
      \foreach \r in {0,...,3} {
        \node[graph vertex] (c\c r\r) at ({1.15*\c},{\r}) {};
      }
      \foreach \r in {0,...,2}
        \draw[diagram edge] (c\c r\r)--(c\c r\the\numexpr\r+1\relax);
      \draw[diagram edge,bend left=35] (c\c r0) to (c\c r3);
    }
    \foreach \c in {0,...,4} {
      \foreach \r in {0,...,3}
        \draw[diagram edge] (c\c r\r)--(c\the\numexpr\c+1\relax r\r);
    }
    \node[font=\scriptsize] at (2.88,-0.65) {$P_c$ direction};
    \node[font=\scriptsize,rotate=90] at (-0.7,1.5) {$C_r$ direction};
  \end{tikzpicture}
  \caption{A small schematic of $C_r\square P_c$.  Each displayed column is
  cyclic; the generated C2.2.3 and C2.2.4 instances use $(r,c)=(16,25)$ and
  $(12,33)$, respectively.}
  \label{fig:sumcheck-cylinder}
\end{figure}
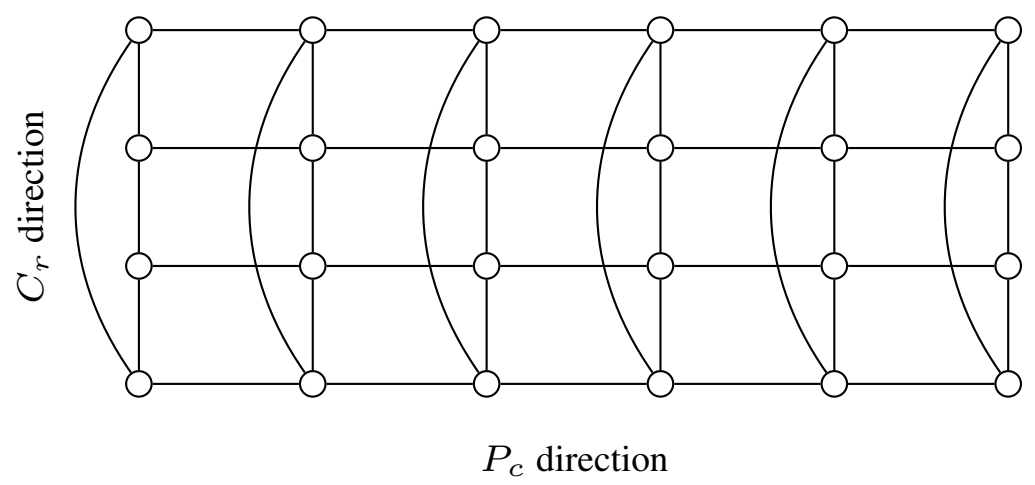

\paragraph{C2.2.4 Cylindrical two-spin model of width 12.}
Here the logical graph is $C_{12}\square P_{33}$.  Its 396 spins again receive
unary factors whose two entries are independent and uniform in $\{1,2,3\}$;
its 780 edges receive binary factors whose four entries are sampled in the same
way.  Coordinates and factor-axis order follow C2.2.3, with
$r\in\mathbb Z_{12}$ and $0\leq c<33$.  The edge set consists of 396
cycle-direction edges and
$12(33-1)=384$ longitudinal edges.  The factor counts are therefore 396 unary
and 780 binary, with maximum variable occurrence five.  Figure
\ref{fig:sumcheck-cylinder} shows the common product geometry.

\paragraph{C2.2.5 Cylindrical monomer--dimer model.}
Let $H=C_6\square P_{30}$, with 180 vertices and 354 edges.  There is one
Boolean variable $x_e$ per edge, where $x_e=1$ means that $e$ is selected.
Each edge has unary weight $(1,w_e)$, where $w_e$ is sampled independently and
uniformly from $\{2,\ldots,8\}$ with edges ordered lexicographically by their
endpoint pairs.  For every
vertex $v$, an occupancy factor on the incident edge variables takes value one
exactly when at most one of them equals one, and zero otherwise:
\[
  \psi_v((x_e)_{e\ni v})=
  \mathbf 1\!\left[\sum_{e\ni v}x_e\leq1\right].
\]
The scope axes of $\psi_v$ list the incident edges in increasing order of their
global lexicographic edge indices.
Hence the nonzero Boolean assignments are precisely the matchings of $H$; an
uncovered vertex is a monomer and a selected edge is a dimer
\cite{heilmann-lieb-1972}.  The model has 354 unary edge factors, 12 arity-three
occupancy factors at the two boundary cycles, and 168 arity-four occupancy
factors in the interior.  Every edge variable occurs in its unary factor and
in the two endpoint occupancy factors.

\begin{figure}[H]
  \centering
  \begin{tikzpicture}[x=1cm,y=1cm]
    \node[sum factor,minimum size=8mm] (v) at (0,0) {$\psi_v$};
    \node[sum variable] (e1) at (-2.2,1.1) {$x_{e_1}$};
    \node[sum variable] (e2) at (2.2,1.1) {$x_{e_2}$};
    \node[sum variable] (e3) at (-2.2,-1.1) {$x_{e_3}$};
    \node[sum variable] (e4) at (2.2,-1.1) {$x_{e_4}$};
    \foreach \x in {e1,e2,e3,e4} \draw[diagram edge] (v)--(\x);
    \node[font=\scriptsize] at (0,-1.55)
      {$\psi_v=1$ iff at most one incident edge is selected};
  \end{tikzpicture}
  \caption{The arity-four occupancy factor at an interior vertex of the
  monomer--dimer graph in C2.2.5.}
  \label{fig:sumcheck-monomer-dimer}
\end{figure}
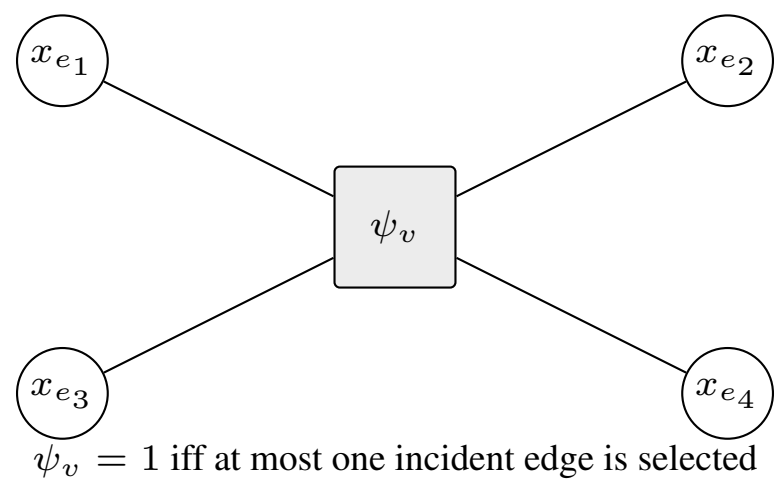

\paragraph{C2.2.6 Six-vertex model on a torus.}
The site graph is the toroidal grid $C_6\square C_{24}$, with 144 sites and
288 edges.  Index sites in row-major order; from each site, index its eastward
edge and then its southward edge, with coordinates taken modulo the two side
lengths.  The four factor axes are its incident edge indices in increasing
order.  Each edge carries one
Boolean variable.  At each site, an arity-four factor has nonzero values exactly
on the six configurations of Hamming weight two, with an independent uniform
weight in $\{1,2,3\}$ for each such configuration.  This local two-of-four
constraint is the six-vertex, or ice, rule.  Every edge also receives a unary
factor $(1,w_e)$, where $w_e$ is sampled independently and uniformly from
$\{1,2,3\}$ and independently of all site weights.  Thus there are 144
arity-four factors and 288 unary
factors, and every variable occurs three times.  The Boolean gauge described
above can
complement edge variables, so the published six allowed entries need not
literally be the Hamming-weight-two entries even though they are a consistent
relabeling of this rule.

\begin{figure}[H]
  \centering
  \begin{tikzpicture}[x=0.9cm,y=0.9cm]
    \foreach \c in {0,...,7} {
      \foreach \r in {0,...,3} {
        \node[graph vertex] (t\c r\r) at (\c,\r) {};
      }
    }
    \foreach \c in {0,...,6} {
      \foreach \r in {0,...,3}
        \draw[diagram edge] (t\c r\r)--(t\the\numexpr\c+1\relax r\r);
    }
    \foreach \c in {0,...,7} {
      \foreach \r in {0,...,2}
        \draw[diagram edge] (t\c r\r)--(t\c r\the\numexpr\r+1\relax);
    }
    \draw[diagram arrow,<->] (-0.2,-0.55)--(7.2,-0.55)
      node[midway,below,font=\scriptsize] {left--right boundaries identified};
    \draw[diagram arrow,<->] (7.65,-0.1)--(7.65,3.1);
    \node[font=\scriptsize,align=center,anchor=west] at (7.92,1.5)
      {top--bottom\\boundaries identified};
  \end{tikzpicture}
  \caption{A fundamental rectangle of a toroidal grid.  C2.2.6 uses six
  rows and 24 columns, with opposite boundaries identified coordinatewise.}
  \label{fig:sumcheck-torus}
\end{figure}
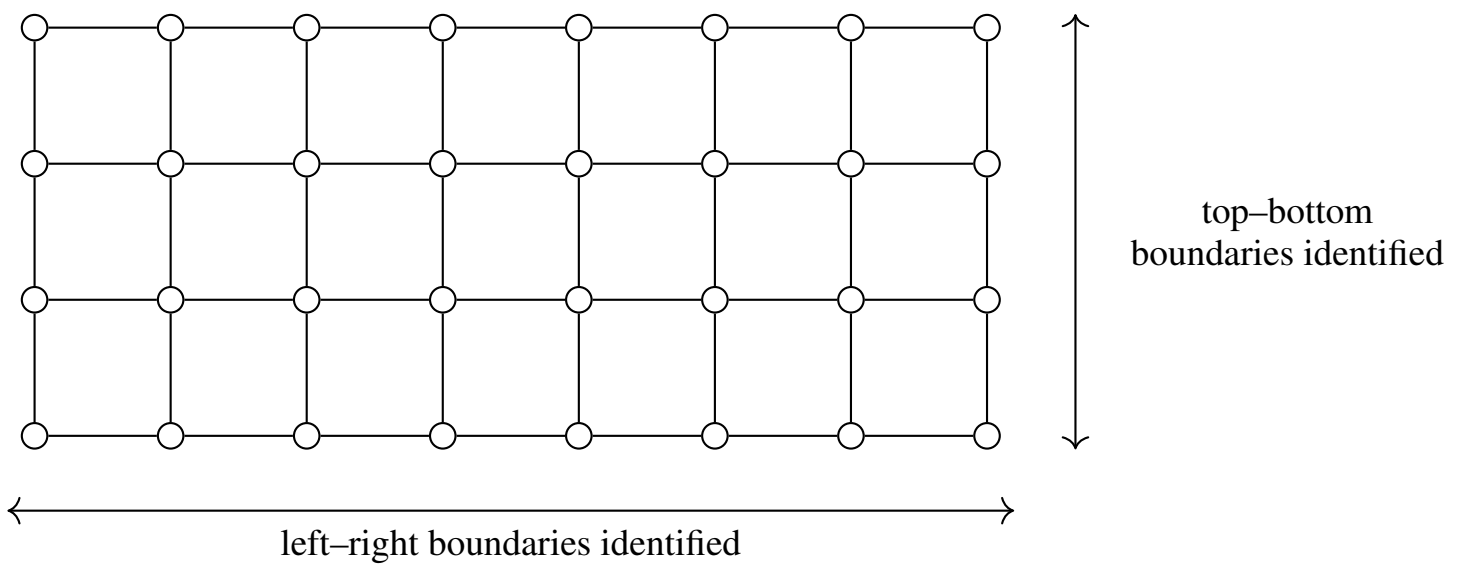

\paragraph{C2.2.7 Memory-ten tail-biting enumerator.}
Let $n=160$ and interpret all subscripts modulo $n$.  The 320 variables are
input bits $u_0,\ldots,u_{n-1}$ and output bits $v_0,\ldots,v_{n-1}$.  For
each time $t$, define the arity-four factor, in the displayed argument order,
by
\[
  Q_t(v_t,u_t,u_{t-3},u_{t-10})
  =\mathbf 1[v_t=u_t\oplus u_{t-3}\oplus u_{t-10}];
  \tag{3}
\]
exactly eight of its 16 table entries are one.  The largest delay is ten, so this
is a memory-ten encoder.  Using all indices modulo $n$ is the tail-biting
convention: the ten delayed input positions wrap around cyclically.  Each $v_t$
has a unary factor $(1,\mu_t)$, where the $\mu_t$ are independent and uniform
in $\{2,\ldots,6\}$.  Consecutive input bits have
the rank-one factor
\[
  J_t(u_t,u_{t+1})=\lambda_t^{u_t+u_{t+1}},
  \qquad \lambda_t\xleftarrow{\$}\{2,\ldots,6\},
\]
where all $\lambda_t$ are independent of one another and of the $\mu_t$.  The
scopes $(u_t,u_{t+1})$, including $(u_{n-1},u_0)$, form the cyclic time-order
graph; algebraically, $J_t$ contributes one factor $\lambda_t$ for each of its
two incident inputs that equals one.  The model
has 160 factors of each arity one, two, and four.  Every input bit occurs in
$Q_t,Q_{t+3},Q_{t+10},J_{t-1},J_t$, and every output bit in $Q_t$ and its
unary factor.

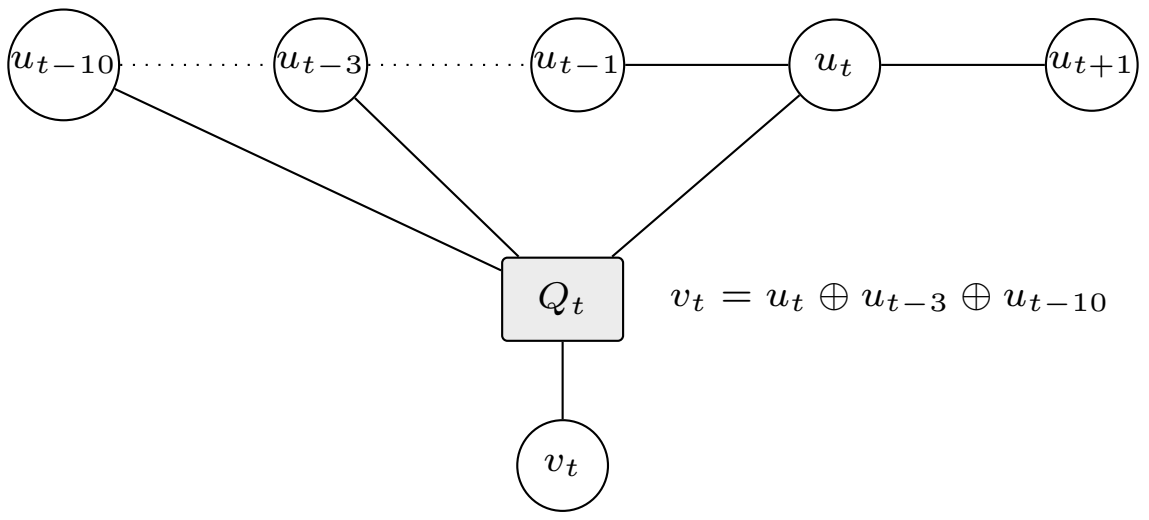
\begin{figure}[H]
  \centering
  \begin{tikzpicture}[x=1cm,y=1cm]
    \foreach \x/\name/\lab in {
      -5.0/utm10/{u_{t-10}},-3.3/utm3/{u_{t-3}},-1.6/utm1/{u_{t-1}},
      0.1/ut/{u_t},1.8/utp1/{u_{t+1}}}
      {
        \node[sum variable] (\name) at (\x,0) {$\lab$};
      }
    \draw[diagram edge,dotted] (utm10)--(utm3);
    \draw[diagram edge,dotted] (utm3)--(utm1);
    \draw[diagram edge] (utm1)--(ut);
    \draw[diagram edge] (ut)--(utp1);
    \node[sum factor] (qt) at (-1.7,-1.55) {$Q_t$};
    \node[sum variable] (vt) at (-1.7,-2.65) {$v_t$};
    \draw[diagram edge] (utm10)--(qt);
    \draw[diagram edge] (utm3)--(qt);
    \draw[diagram edge] (ut)--(qt);
    \draw[diagram edge] (vt)--(qt);
    \node[font=\scriptsize,right=5pt] at (qt.east)
      {$v_t=u_t\oplus u_{t-3}\oplus u_{t-10}$};
  \end{tikzpicture}
  \caption{Local factor-incidence structure in C2.2.7.  Horizontal solid
  edges denote adjacent-input factors; dotted segments abbreviate omitted
  positions of the 160-cycle.}
  \label{fig:sumcheck-tail-biting}
\end{figure}

\newpage
\subsection{GKR Layered Arithmetic-Circuit Evaluation}
\label{sec:gkr}

This task asks Merlin to evaluate a public arithmetic circuit and prove its
scalar output.  The circuit is organized into layers: each gate reads two
values from the next layer, while the last layer contains public inputs.  The
Goldwasser--Kalai--Rothblum (GKR) protocol follows the computation backward,
reducing one claimed layer evaluation to a claimed evaluation of the next
layer until Arthur can check the public input directly
\cite{goldwasser-kalai-rothblum-2008}.  All circuit data are public; the
interaction certifies consistent evaluation rather than knowledge of a hidden
witness.

All arithmetic is over $\F_p$, where the prime $p=2^{61}-1$.  Write
$[N]=\{0,\ldots,N-1\}$.  Layers are numbered from
the scalar output toward the inputs:
\[
  V_0,V_1,\ldots,V_L,
  \qquad N_i=|V_i|=2^{s_i},
  \qquad N_0=1.
\]
The table $V_L$ contains the public inputs.  For $i<L$ and gate position
$g\in[N_i]$, the public circuit specifies field coefficients
$a_i(g),b_i(g),c_i(g)$ and child positions
$\lambda_i(g),\rho_i(g)\in[N_{i+1}]$, with semantics
\[
  V_i(g)=a_i(g)V_{i+1}(\lambda_i(g))
        +b_i(g)V_{i+1}(\rho_i(g))
        +c_i(g)V_{i+1}(\lambda_i(g))V_{i+1}(\rho_i(g)).
  \tag{4}
\]
Equation (4) represents copy, addition, subtraction, scalar multiplication,
and multiplication gates by appropriate coefficient choices.  A semantic gate
list is padded by appending positions until its length is the next power of
two.  Every ordinary padding position is the canonical zero gate
$(a,b,c,\lambda,\rho)=(0,0,0,0,0)$; the explicitly specified expansion gates
in C2.3.5--C2.3.7 are exceptions.  Within every semantic micro-layer, gates
and copy rails are appended in the order stated by the case construction, and
only then is padding appended.  The generator and serialized public
gate table are the canonical specification of this low-level ordering and of
all child indices.  The descriptions below specify the semantic circuit and
all nontrivial instance parameters; they are not intended as a second,
byte-for-byte reimplementation of that serializer.  Merlin initially claims
$y=V_0(0)$.

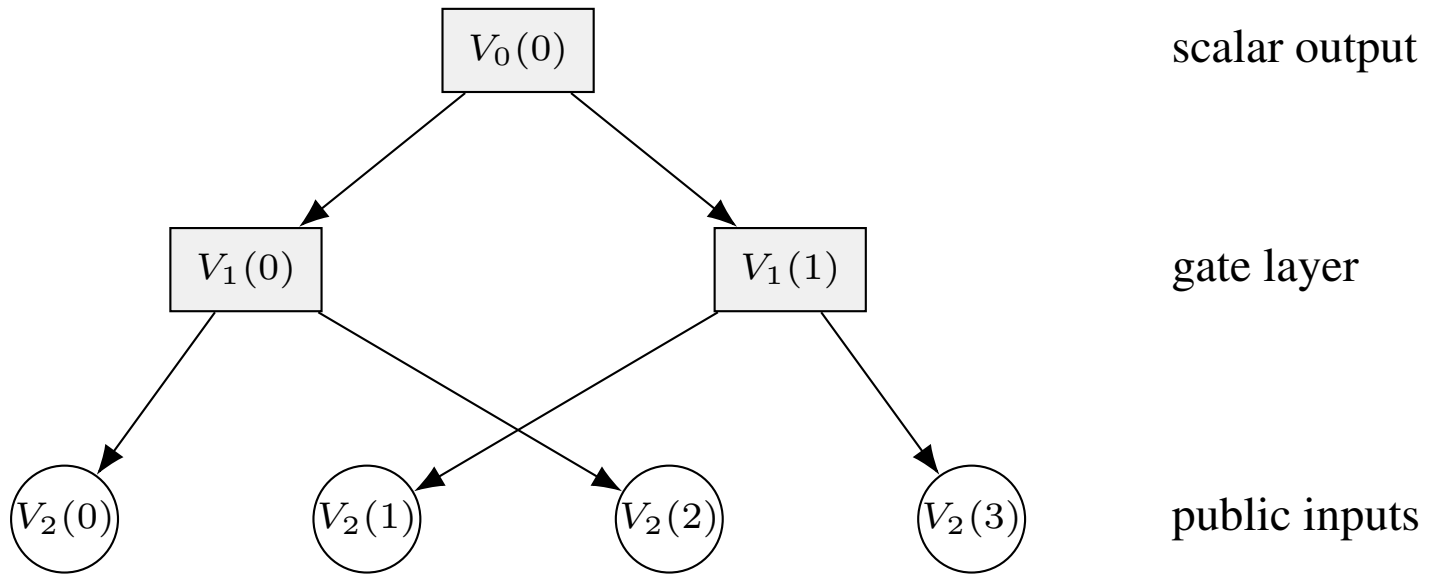
\begin{figure}[H]
  \centering
  \begin{tikzpicture}[x=1cm,y=1cm]
    \node[circuit gate] (out) at (0,2.1) {$V_0(0)$};
    \node[circuit gate] (g10) at (-1.8,0.65) {$V_1(0)$};
    \node[circuit gate] (g11) at (1.8,0.65) {$V_1(1)$};
    \node[sum variable] (x0) at (-3.0,-1.0) {$V_2(0)$};
    \node[sum variable] (x1) at (-1.0,-1.0) {$V_2(1)$};
    \node[sum variable] (x2) at (1.0,-1.0) {$V_2(2)$};
    \node[sum variable] (x3) at (3.0,-1.0) {$V_2(3)$};
    \draw[diagram arrow] (out)--(g10);
    \draw[diagram arrow] (out)--(g11);
    \draw[diagram arrow] (g10)--(x0);
    \draw[diagram arrow] (g10)--(x2);
    \draw[diagram arrow] (g11)--(x1);
    \draw[diagram arrow] (g11)--(x3);
    \node[font=\small,anchor=west] at (4.2,2.1) {scalar output};
    \node[font=\small,anchor=west] at (4.2,0.65) {gate layer};
    \node[font=\small,anchor=west] at (4.2,-1.0) {public inputs};
  \end{tikzpicture}
  \caption{A strictly layered fan-in-two arithmetic circuit.  Each rectangle
  applies (4) to two positions in the next layer.}
  \label{fig:gkr-layered-circuit}
\end{figure}

Identify position $j\in[N_i]$ with
$\operatorname{bits}_{s_i}(j)$, its $s_i$-bit binary vector in
least-significant-bit-first order.  For $z\in\F_p^s$ and
$b\in\{0,1\}^s$, define
\[
  \operatorname{eq}_s(z,b)=
  \prod_{k=0}^{s-1}\bigl((1-z_k)(1-b_k)+z_kb_k\bigr).
\]
The multilinear extension of a layer is
\[
  \widetilde V_i(z)=
  \sum_{g=0}^{N_i-1}
  \operatorname{eq}_{s_i}(z,\operatorname{bits}_{s_i}(g))V_i(g).
  \tag{5}
\]
For $D\in\{A,B,M\}$, let $d_i(g)$ be, respectively,
$a_i(g),b_i(g),c_i(g)$, and set
\[
\begin{split}
  \widetilde D_i(r,u,v)=\sum_{g=0}^{N_i-1}&d_i(g)
  \operatorname{eq}_{s_i}(r,\operatorname{bits}_{s_i}(g))\\
  &\cdot\operatorname{eq}_{s}(u,
      \operatorname{bits}_{s}(\lambda_i(g)))
  \operatorname{eq}_{s}(v,
      \operatorname{bits}_{s}(\rho_i(g))),
\end{split}
\tag{6}
\]
where $s=s_{i+1}$.  Expanding (4) on the Boolean child positions gives
\[
  \widetilde V_i(r)=\sum_{u,v\in\{0,1\}^{s}}F_i(r,u,v),
  \tag{7}
\]
with
\[
  F_i(r,u,v)=\widetilde A_i(r,u,v)\widetilde V_{i+1}(u)
     +\widetilde B_i(r,u,v)\widetilde V_{i+1}(v)
     +\widetilde M_i(r,u,v)\widetilde V_{i+1}(u)\widetilde V_{i+1}(v).
\]
Every coordinate of $F_i$ has degree at most two.

The protocol starts from the unique point $r_0=()\in\F_p^0$ and $C_0=y$.
At the beginning of layer $i$, the current claim is
$C_i=\widetilde V_i(r_i)$.  Merlin and Arthur apply Sum-Check to (7), with
claimed sum $C_i$, over the $2s$ coordinates
$u_0,\ldots,u_{s-1},v_0,\ldots,v_{s-1}$ in that order.  In each round Merlin
submits exactly three coefficients of the degree-at-most-two remaining-sum
polynomial, in increasing-power order and including a zero quadratic
coefficient when necessary.  Arthur checks that the first polynomial's values
at zero and one sum to $C_i$, and thereafter that they sum to the preceding
polynomial evaluated at the previous challenge.  Only then does he sample the
next challenge independently and uniformly from $\F_p$.  The challenges define
$u^*,v^*\in\F_p^s$ and leave one claimed value $H$ for $F_i$.

Merlin next supplies
\[
  U=\widetilde V_{i+1}(u^*),
  \qquad V=\widetilde V_{i+1}(v^*),
\]
and all $s+1$ coefficients, in increasing-power order, of the line restriction
\[
  d(T)=\widetilde V_{i+1}\bigl(u^*+T(v^*-u^*)\bigr),
  \qquad \deg d\leq s.
  \tag{8}
\]
Arthur evaluates the wiring polynomials (6) at $(r_i,u^*,v^*)$ from the public
gate table and checks
\[
  H=\widetilde A_i(r_i,u^*,v^*)U
    +\widetilde B_i(r_i,u^*,v^*)V
    +\widetilde M_i(r_i,u^*,v^*)UV,
  \qquad d(0)=U,\qquad d(1)=V.
\]
After these values are committed, he samples an independent uniform
$t_i\in\F_p$ and sets
\[
  r_{i+1}=u^*+t_i(v^*-u^*),
  \qquad C_{i+1}=d(t_i).
\]
After layer $L-1$, Arthur computes $\widetilde V_L(r_L)$ from the public input
table and accepts exactly when it equals $C_L$.  The transcript contains
\[
  1+\sum_{i=0}^{L-1}(2s_{i+1}+1)
\]
Merlin messages: one output claim, $2s_{i+1}$ Sum-Check polynomials per layer,
and one line-reduction message per layer.

\begin{lemma}[Perfect completeness and soundness]
\label{lem:tier2-gkr}
An honest Merlin claiming $V_0(0)$ is accepted with probability $1$.  For a
false output claim, the generated instances satisfy
\[
  \Pr[\mathrm{accept}]
  \leq\frac{5\sum_{i=0}^{L-1}s_{i+1}}p<2^{-49}.
\]
\end{lemma}

\begin{proof}
For an honest Merlin, (7) is the exact expansion of the gate semantics, every
Sum-Check polynomial is the true remaining sum, and (8) is the actual
restriction of the next-layer MLE.  All checks therefore hold exactly.
At a layer carrying a false claim, degree-two Sum-Check over $2s_{i+1}$
coordinates contributes at most $4s_{i+1}/p$ soundness error.  If a false
next-layer value passes the endpoint checks, the submitted line polynomial and
the true restriction differ by a nonzero polynomial of degree at most
$s_{i+1}$, which agrees at the fresh $t_i$ with probability at most
$s_{i+1}/p$.  Otherwise the false claim reaches the final public input check.
A union bound proves the formula.  Every generated circuit uses at most 512
verifier challenges, so $\sum_i s_{i+1}<256$, which gives the numerical bound.
\end{proof}

Each case is built in semantic coordinates and padded to power-of-two layer
widths.  The input layer and every non-output gate layer are then permuted by
independent uniform permutations from the seeded random stream, and all child
indices are rewritten consistently.  These permutations preserve the scalar
output but remove the original semantic ordering.
Before these permutations, a \emph{logical position} is the zero-based index
in the ordered semantic gate list after ordinary power-of-two padding.  In an
expanded layer, the active application gates occupy the prefix in their
original order and the explicitly defined extra gates occupy the remaining
indices.  References to an ``original logical prefix'' in C2.3.5--C2.3.7 use
this convention; the following layer refers to its images after permutation.
For the numerical summaries below, $L$ is the number of non-input layers,
$W=\max_{0\leq i\leq L}N_i$, $S=\sum_{i=0}^{L}N_i$, and
$R=\sum_i(2s_{i+1}+1)$ is the number of post-claim randomness barriers.

\paragraph{C2.3.1 Balanced inner product.}
The generator samples all coordinates of $a,b\in\F_p^{64}$ independently and
uniformly.  In semantic input order, positions $0,\ldots,63$ store $a$ and
positions $64,\ldots,127$ store $b$.  One layer computes $a_jb_j$ at position
$j$, and each of six balanced addition layers adds consecutive pairs of its
child layer, reducing them to
\[
  y=\sum_{j=0}^{63}a_jb_j.
\]
From input to output the widths are $128,64,32,16,8,4,2,1$.  Thus
$(L,W,S,R)=(7,128,255,63)$.

\paragraph{C2.3.2 Batched second-order affine recurrences.}
There are $B=8$ independent batches and $T=6$ states per batch.  For each
$b\in\{0,\ldots,7\}$, the generator samples
$\alpha_b,\beta_b,\gamma_b,u_{b,0},u_{b,1},\eta_b$ independently and uniformly
from $\F_p$, stores these six rails consecutively for each batch, and computes
\[
  u_{b,t+2}=\alpha_bu_{b,t+1}+\beta_bu_{b,t}+\gamma_b,
  \qquad 0\leq t\leq3.
\]
For each update, the first fan-in-two layer copies the six rails and appends
$\alpha_bu_{b,t+1}$ and $\beta_bu_{b,t}$; the second copies
$\alpha_b,\beta_b,\gamma_b,u_{b,t+1},\eta_b$ and forms the sum of the products;
the third adds $\gamma_b$ and restores the six-rail order with state pair
$(u_{b,t+1},u_{b,t+2})$.  After four updates, one layer computes the eight
products $\eta_bu_{b,5}$, and three consecutive-pair addition layers produce
\[
  y=\sum_{b=0}^{7}\eta_bu_{b,5}.
\]
The resulting profile is $(L,W,S,R)=(16,64,847,184)$.

\paragraph{C2.3.3 Matrix triple product.}
The public inputs are three matrices
$A,B,C\in\F_p^{8\times8}$, whose entries are sampled independently and
uniformly and stored in matrix order $A,B,C$, each in row-major order.  A
512-gate product table, ordered lexicographically by $(i,j,k)$, forms
$A_{ik}B_{kj}$.  The 64 entries of $C$ are copied after these products.  Three
balanced layers add consecutive $k$-pairs while continuing to copy $C$; this
computes $(AB)_{ij}$.  The 64 results are multiplied by $C_{ji}$, and six
consecutive-pair addition layers compute
\[
  y=\operatorname{tr}(ABC)
   =\sum_{i,j,k=0}^{7}A_{ik}B_{kj}C_{ji}.
\]
The 512 products together with 64 copied entries pad to width 1,024.  The
profile is $(L,W,S,R)=(11,1024,2303,137)$.

\paragraph{C2.3.4 Walsh--Hadamard XOR convolution.}
The generator samples every coordinate of $a,b,d\in\F_p^{128}$ independently
and uniformly and stores the three vectors consecutively.  It defines
\[
  c_z=\sum_{x=0}^{127}a_xb_{x\oplus z},
  \qquad y=\sum_{z=0}^{127}c_zd_z,
\]
where $\oplus$ is bitwise exclusive-or on seven-bit indices.  The circuit
computes $c$ using the fast Walsh--Hadamard transform.  At butterfly stage
$q\in\{0,\ldots,6\}$, indices $j$ with bit $q$ equal to zero are paired with
$j+2^q$; the two output positions receive
\[
  (x,y)\longmapsto(x+y,x-y)
\]
The seven stages are applied in parallel to $a$ and $b$, while all entries of
$d$ are copied.  The transformed vectors are multiplied coordinatewise; the
same seven stages are then applied to that product, followed by multiplication
by $128^{-1}\bmod p$.  Finally, the result is multiplied coordinatewise by $d$
and reduced through seven consecutive-pair addition layers.
The profile is $(L,W,S,R)=(24,512,6655,368)$.

\paragraph{Proof-relevant inactive expansion layers.}
C2.3.5--C2.3.7 each contain one layer of width $2^{18}$ or $2^{19}$.  A prefix of that
layer carries the application computation, while its remaining gates are not
referenced by the next semantic layer and therefore do not affect $V_0(0)$.
They are nevertheless not removable from the GKR proof.  A random MLE
evaluation of the entire wide layer is a linear combination of all its
positions; after line reduction, the next GKR claim therefore depends on the
values of both the active and inactive positions.  Per-layer permutation
removes the contiguous-prefix layout from the published numbering.

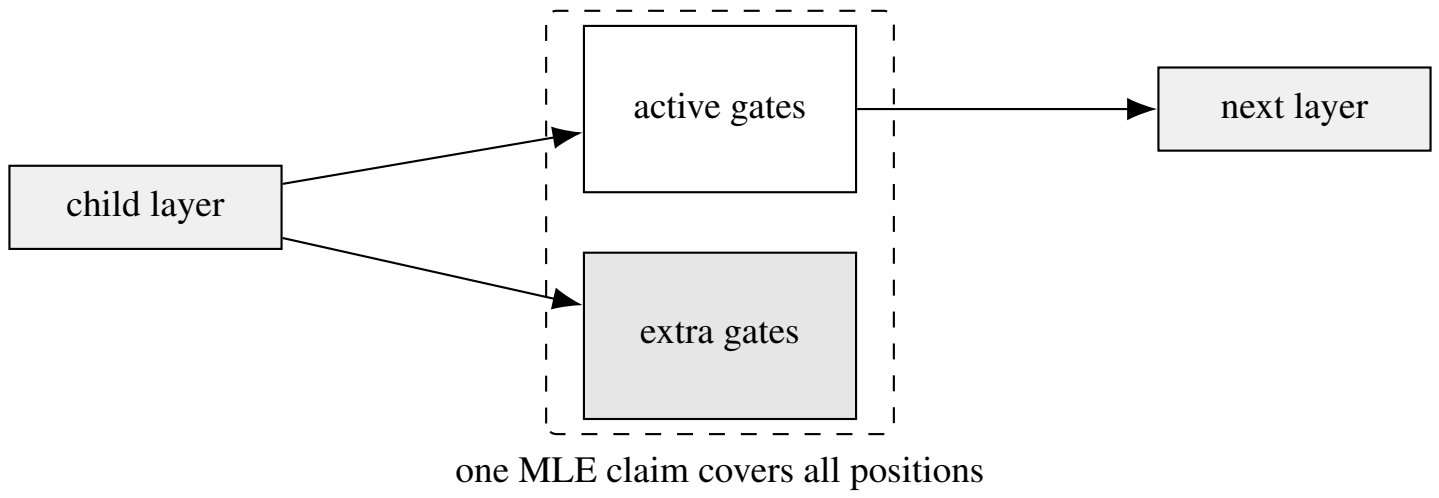
\begin{figure}[H]
  \centering
  \begin{tikzpicture}[x=1cm,y=1cm]
    \node[circuit gate,minimum width=18mm] (prev) at (-3.8,0) {child layer};
    \node[draw,rectangle,minimum width=18mm,minimum height=11mm,
      fill=white,font=\scriptsize] (active) at (0,0.65) {active gates};
    \node[draw,rectangle,minimum width=18mm,minimum height=11mm,
      fill=gray!20,font=\scriptsize] (extra) at (0,-0.85) {extra gates};
    \node[circuit gate,minimum width=18mm] (next) at (3.8,0.65) {next layer};
    \draw[dashed,rounded corners=2pt] (-1.15,-1.50) rectangle (1.15,1.30);
    \draw[diagram arrow] (prev)--(active);
    \draw[diagram arrow] (prev)--(extra);
    \draw[diagram arrow] (active)--(next);
    \node[font=\scriptsize] at (0,-1.76)
      {one MLE claim covers all positions};
  \end{tikzpicture}
  \caption{Logical role of an expanded layer before its positions are
  permuted.  Extra gates do not feed the application output, but their values
  enter a random multilinear evaluation of the complete layer.}
  \label{fig:gkr-expansion}
\end{figure}

For the exact definitions below, $j\mathbin{\mathrm{shr}}k$ denotes
$\lfloor j/2^k\rfloor$, $\mathbin{\oplus_b}$ denotes bitwise XOR on
nonnegative integers, and all gate coefficients are reduced modulo $p$.

\paragraph{C2.3.5 Bitonic Boolean-key payload routing.}
The input consists of eight records, each with a Boolean key $k_i$ and four
independent uniform nonzero field-valued payloads.  The keys are independent
uniform bits, and records are stored in index order with the key before its
four payloads.  A six-stage Batcher bitonic sorting network
\cite{batcher-1968} routes the records into nondecreasing key order.  For an
ascending comparator on keys $a,b$, let
\[
  s=a(1-b)=a-ab,
  \qquad a'=a-s,\quad b'=b+s.
\]
For a payload pair $P,Q$, the same swap bit gives
\[
  P'=P+s(Q-P),\qquad Q'=Q+s(P-Q).
  \tag{9}
\]
A descending comparator uses $s=(1-a)b$.  Each comparator stage is expanded
into five fan-in-two layers: copy all record fields and compute $ab$; copy them
and compute the direction-specific swap bit; copy them and $s$ and compute both
payload differences; multiply each difference by $s$; and write the updated
keys and payloads.  Formula (9) is applied independently to all four payload
coordinates.  The six comparator stages, listing $(i,j)$ followed by $+$ for
ascending and $-$ for descending, are
\[
\begin{array}{c|l}
1&(01)+,(23)-,(45)+,(67)-\\
2&(02)+,(13)+,(46)-,(57)-\\
3&(01)+,(23)+,(45)-,(67)-\\
4&(04)+,(15)+,(26)+,(37)+\\
5&(02)+,(13)+,(46)+,(57)+\\
6&(01)+,(23)+,(45)+,(67)+
\end{array}
\]
After sorting, 40 independent uniform nonzero field weights multiply the
flattened records, and six consecutive-pair addition layers reduce the padded
64-position table to the scalar digest.

The third micro-layer of the first comparator stage (the first layer that
contains payload differences) has 128 active gates and is extended to
$D=2^{19}$ positions.  The preceding layer has width 64.  For
$128\leq j<D$, the extra gate reads
\[
  \ell_j=j\bmod64,
  \qquad
  r_j=((j\mathbin{\mathrm{shr}}6)\mathbin{\oplus_b}(17j)
       \mathbin{\oplus_b}45)\bmod64,
\]
and has coefficients
\[
\begin{split}
 a_j&=\alpha+\mathtt{0x9e3779b1}\,j,\\
 b_j&=\beta +\mathtt{0x85ebca77}\,j,\\
 c_j&=\gamma+\mathtt{0xc2b2ae3d}\,j,
\end{split}
\]
where $\alpha,\beta,\gamma$ are sampled independently and uniformly from
$\F_p^*$.
Only the original 128 logical positions feed the next comparator layer.  The
profile is $(L,W,S,R)=(37,524288,527039,499)$.

\paragraph{C2.3.6 Rule 110 periodic spacetime.}
Rule 110 is a radius-one binary cellular automaton.  The generator samples a
Boolean state $x^{(0)}\in\{0,1\}^{256}$ as independent uniform bits on a ring
and performs four synchronous
updates, with indices modulo 256:
\[
  x_j^{(t+1)}=R(x_{j-1}^{(t)},x_j^{(t)},x_{j+1}^{(t)}),
  \qquad R(\ell,c,r)=c+r-cr-\ell cr.
  \tag{10}
\]
On neighborhoods $111,110,101,100,011,010,001,000$, (10) gives
$0,1,1,0,1,1,1,0$, respectively.  Each update is expanded into four binary
layers.  The first copies the 256 cells and appends all $cr$ products.  The
second computes all $c+r$, copies $cr$, and computes $\ell(cr)$.  The third
forms $c+r-cr$ and copies $\ell cr$; the fourth subtracts these two values.
After four updates, 256 independent uniform nonzero field weights multiply
$x^{(4)}$, and eight consecutive-pair addition layers form the scalar digest.

The second micro-layer of the first update has 1,024 logical gate and rail
positions and is expanded to $D=2^{19}$ gates;
its child layer has width 512.  For $j\geq1024$, define
\[
  \ell_j=(j\mathbin{\oplus_b}(j\mathbin{\mathrm{shr}}9)
             \mathbin{\oplus_b}\mathtt{0x151})\bmod512,
  \quad
  r_j=(29j+(j\mathbin{\mathrm{shr}}4)+\mathtt{0x6d})\bmod512,
\]
with coefficients
\[
  (a_j,b_j,c_j)=
  (\alpha+\mathtt{0x27d4eb2d}j,
   \beta +\mathtt{0x165667b1}j,
   \gamma+\mathtt{0xd3a2646c}j).
\]
$\alpha,\beta,\gamma$ are sampled independently and uniformly from $\F_p^*$.
Only the original 1,024 logical positions feed the next micro-layer.  The profile is
$(L,W,S,R)=(25,524288,533247,419)$.

\paragraph{C2.3.7 Boolean repeated-squaring transitive closure.}
The input is a directed graph on $n=16$ vertices.  The diagonal of its
adjacency matrix is set to one; every off-diagonal arc is otherwise included
independently with probability $1/4$.  The resulting row-major Boolean table is
$R^{(0)}$.  Four
Boolean matrix squarings are then performed:
\[
  R^{(t+1)}_{ij}=
  \bigvee_{k=0}^{15}
  \bigl(R^{(t)}_{ik}\wedge R^{(t)}_{kj}\bigr),
  \qquad 0\leq t<4.
  \tag{11}
\]
The arithmetic circuit represents AND by $xy$ and binary OR by $x+y-xy$.
Each squaring uses a 4,096-gate product layer in lexicographic $(i,j,k)$ order,
followed by four balanced OR layers that combine consecutive $k$-pairs for each
fixed $(i,j)$.  Induction gives
\[
  R^{(t)}_{ij}=1
  \iff
  \text{there is an $i$-to-$j$ walk of length at most $2^t$}.
\]
The diagonal ones pad shorter walks, and every reachable pair has a simple path
of length at most $15<2^4$.  Thus $R^{(4)}$ is the transitive closure.  Eight
consecutive-pair addition layers compute
$y=\sum_{i,j}R^{(4)}_{ij}$, which counts reachable ordered pairs as an ordinary
integer because $0\leq y\leq256<p$.

The first product layer is expanded from 4,096 to $D=2^{18}$ positions; its
child layer is the 256-entry relation table.  For $j\geq4096$, define
\[
\begin{split}
  \ell_j={}&(j\mathbin{\oplus_b}(j\mathbin{\mathrm{shr}}8)
        \mathbin{\oplus_b}(j\mathbin{\mathrm{shr}}16)
        \mathbin{\oplus_b}\mathtt{0xa7})\bmod256,\\
  r_j={}&(37j+(j\mathbin{\mathrm{shr}}5)+\mathtt{0x53})\bmod256,
\end{split}
\]
with coefficients
\[
  (a_j,b_j,c_j)=
  (\alpha+\mathtt{0x7feb352d}j,
   \beta +\mathtt{0x846ca68b}j,
   \gamma+\mathtt{0x9e3779b9}j).
\]
$\alpha,\beta,\gamma$ are sampled independently and uniformly from $\F_p^*$.
Only the original 4,096 logical product positions, or their images under the
subsequent layer permutation, feed the first OR-reduction layer.  The
profile is $(L,W,S,R)=(28,262144,290303,512)$.

\newpage
\subsection{Boolean-Circuit Image-Size Lower Bound}
\label{sec:image-size-lower-bound}

This task gives a public Boolean circuit and asks Merlin to demonstrate that
the circuit has many distinct outputs.  The set of possible outputs is the
\emph{image} of the function computed by the circuit; if many inputs collide,
the image may be much smaller than the domain.  Merlin does not report the
image size directly.  Instead, Arthur chooses random affine constraints on the
output, and Merlin supplies circuit inputs whose outputs satisfy sufficiently
many of them.  A large image is likely to intersect most of the random affine
buckets, whereas a small image is unlikely to do so.  This is a finite-field
instantiation of the hashing technique used in public-coin set-size lower-bound
protocols \cite{goldwasser-sipser-1986}.

The public circuit uses Boolean constants and the gates $\mathsf{NOT}$,
$\mathsf{XOR}$, and $\mathsf{AND}$ and computes
\[
  F:\{0,1\}^m\longrightarrow\{0,1\}^n.
\]
Its gates are given in topological order, so Arthur can evaluate $F(x)$ for any
submitted input.  Define
\[
  S=\operatorname{Im}F
   =\{F(x):x\in\{0,1\}^m\},
  \qquad T=2^t,
\]
where $t\geq0$ is part of the instance.  The promise is
\[
  |S|\geq16T \quad(\mathsf{YES}),
  \qquad\text{or}\qquad
  |S|\leq T \quad(\mathsf{NO}).
  \tag{12}
\]
All generated instances are $\mathsf{YES}$ instances.  A \emph{preimage} of an
output $z\in S$ is any input $x$ satisfying $F(x)=z$; no canonical preimage is
required.

All hashing arithmetic is over $\F_2$.  Put $k=t+2$.  For each lane
$j\in\{0,\ldots,127\}$, Arthur independently samples
\[
  A_j\xleftarrow{\$}\F_2^{k\times n},
  \qquad b_j\xleftarrow{\$}\F_2^k.
\]
The corresponding affine bucket is
\[
  B_j=\{z\in\F_2^n:A_jz=b_j\}.
  \tag{13}
\]
Uniform affine hashing is pairwise independent
\cite{carter-wegman-1979}: for distinct $z,z'$, membership of $z$ and $z'$ in
a fixed random bucket is independent.

The three-step interaction contains two Merlin messages.  Merlin first sends a
fixed, content-independent start signal.  Only after it has been recorded does Arthur sample and
return all 128 pairs $(A_j,b_j)$.  Merlin then returns one entry per lane: either
an input $x_j\in\{0,1\}^m$ or a null entry.  For every non-null entry, Arthur
evaluates the public circuit and checks
\[
  A_jF(x_j)=b_j.
\]
A null entry fails; a non-null lane succeeds exactly when this equality holds,
and Arthur accepts exactly when at least 64 lanes succeed.  Thus a purported output is
insufficient: every counted lane must contain an actual circuit preimage. The
statistical statement below abstracts away this operational limit.  In the
generated instances $m\leq12$, so exhaustive evaluation of at most $2^{12}$
inputs suffices to find a preimage for every answerable bucket.

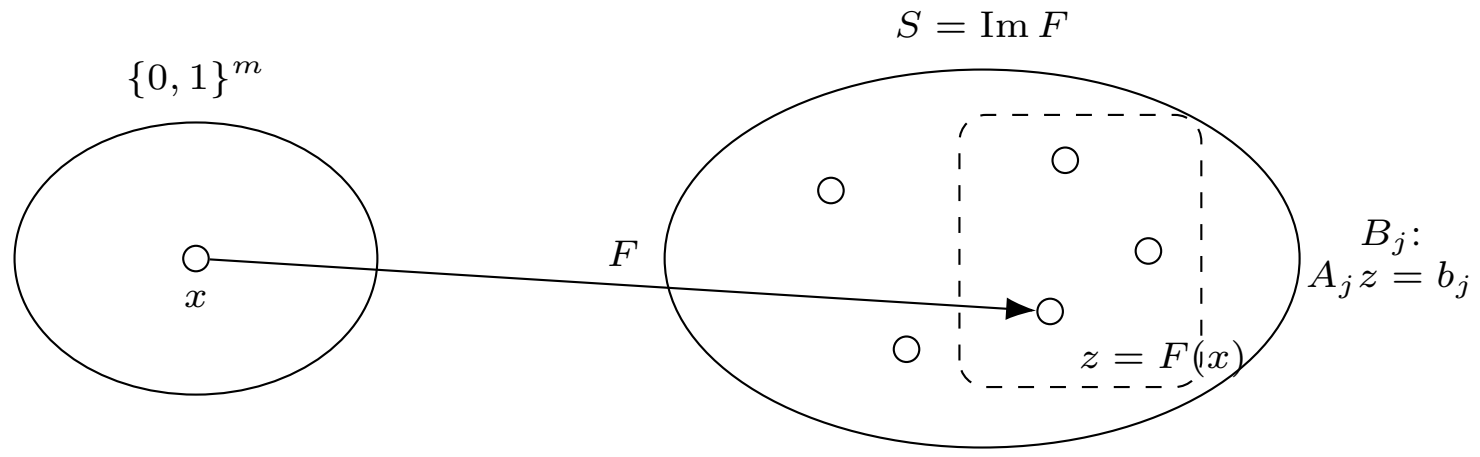
\begin{figure}[H]
  \centering
  \begin{tikzpicture}[x=1cm,y=1cm]
    \node[draw,ellipse,minimum width=24mm,minimum height=18mm]
      (dom) at (-4.2,0) {};
    \node[font=\scriptsize] at (-4.2,1.18) {$\{0,1\}^m$};
    \node[graph vertex,label={[font=\scriptsize]below:$x$}]
      (x) at (-4.2,0) {};
    \node[draw,ellipse,minimum width=42mm,minimum height=25mm]
      (img) at (1.0,0) {};
    \node[font=\scriptsize] at (1.0,1.55) {$S=\operatorname{Im}F$};
    \node[graph vertex] at (0.0,0.45) {};
    \node[graph vertex] at (0.5,-0.60) {};
    \node[graph vertex] at (1.55,0.65) {};
    \node[graph vertex,label={[font=\scriptsize]below right:$z=F(x)$}]
      (z) at (1.45,-0.35) {};
    \node[graph vertex] at (2.1,0.05) {};
    \draw[dashed,rounded corners=5pt] (0.85,-0.85) rectangle (2.45,0.95);
    \node[font=\scriptsize,align=center] at (3.7,0)
      {$B_j$:\\$A_jz=b_j$};
    \draw[diagram arrow] (x)--node[above,font=\scriptsize] {$F$} (z);
  \end{tikzpicture}
  \caption{One lane of the image-size protocol.  A valid response gives an
  input $x$ whose circuit output lies in Arthur's random affine bucket $B_j$.}
  \label{fig:image-size-bucket}
\end{figure}

\begin{lemma}[Statistical completeness and soundness]
\label{lem:tier2-image}
On a $\mathsf{YES}$ instance, an unbounded honest Merlin is rejected with
probability less than $10^{-10}$.  On a $\mathsf{NO}$ instance, every Merlin is
accepted with probability less than $2^{-23}$.
\end{lemma}

\begin{proof}
For one lane, let
\[
  X=|S\cap B_j|,
  \qquad \lambda=\frac{|S|}{2^k}.
\]
Pairwise independence gives
$\mathbb E[X]=\lambda$ and
$\mathbb E[X^2]\leq\lambda^2+\lambda$.  The Paley--Zygmund inequality at
threshold zero therefore yields
\[
  \Pr[X>0]\geq\frac{\lambda}{1+\lambda}.
\]
For a $\mathsf{YES}$ instance, $2^k=4T$ and $\lambda\geq4$, so a lane is
answerable with probability at least $4/5$.  For a $\mathsf{NO}$ instance, the
union bound gives $\Pr[X>0]\leq|S|/2^k\leq1/4$.  The 128 lane indicators are
independent.  Hoeffding's inequality consequently gives
\[
\begin{split}
 \Pr_{\mathsf{YES}}[\text{fewer than 64 successes}]
 &\leq \exp\!\bigl(-2\cdot128(0.8-0.5)^2\bigr)<10^{-10},\\
 \Pr_{\mathsf{NO}}[\text{at least 64 successes}]
 &\leq \exp\!\bigl(-2\cdot128(0.5-0.25)^2\bigr)<2^{-23}.
\end{split}
\]
A Merlin cannot make an empty bucket successful, because Arthur independently
evaluates $F(x_j)$ and (13) for every counted response.
\end{proof}

For every case, the construction has an exact image size $s\geq16$, as
established below, and the generator sets
\[
  t=\lfloor\log_2s\rfloor-4.
  \tag{14}
\]
Thus $16\cdot2^t\leq s<32\cdot2^t$, so the instance satisfies the
$\mathsf{YES}$ side of (12).  The logical construction assigns input roles by a
uniform permutation from the seeded random stream, then applies an independent
uniform output-coordinate permutation and independent uniform output-complement
bits.  Writing $\pi$ for the linear map induced by the coordinate permutation
and $c\in\F_2^n$ for the complement mask, the latter operation has the form
$z\mapsto\pi(z)\oplus c$ and is a bijection of $\F_2^n$, so it preserves the
image size.  The following descriptions use the logical coordinates before
this hiding transformation.  The canonical compiler numbers the input wires
$0,\ldots,m-1$ and then appends each requested gate immediately, so every gate
reads only earlier wires.  Constants zero and one are each emitted on first use
and reused thereafter.  Iterated XORs and conjunctions are left-associated;
the empty XOR is zero and the empty conjunction is one.  A requested OR
$a\vee b$ is emitted as three successive gates
$u=a\oplus b$, $v=a\wedge b$, and $u\oplus v$.  The logical output list is
permuted only after all of its wires have been constructed, and the NOT gates
that implement output complementation are then appended.  Together with each
case's stated loop order, these conventions determine the published
topological gate table.

\paragraph{C2.4.1 Systematic subcube image.}
The circuit has $m=10$ inputs and $n=12$ outputs.  Eight randomly selected
inputs are copied to eight logical output coordinates: specifically, a uniform
permutation of $[10]$ is sampled and its first eight distinct entries are
$(i_1,\ldots,i_8)$ in output order.  The other four output coordinates are
zero, and the remaining two inputs are unused.  Before the
final affine hiding, the map is
\[
  (x_1,\ldots,x_{10})\longmapsto
  (x_{i_1},\ldots,x_{i_8},0,0,0,0).
\]
The eight active bits vary independently, so
$|\operatorname{Im}F|=2^8=256$.  Equation (14) gives $t=4$, and each challenge
has $k=6$ hash equations.

\paragraph{C2.4.2 Hidden affine image.}
This case has $m=12$ and $n=14$.  Eight shuffled input coordinates are active.
The first eight logical outputs copy them individually, while each of the six
remaining outputs is determined as follows: independently include every active
input with probability $1/2$; if fewer than two are included, replace the list
by two independently sampled active coordinates, allowing repetition.  The
output is the left-associated XOR of that list.  After
the output permutation and shift, the circuit computes
\[
  F(x)=Lx\oplus c,
  \qquad L\in\F_2^{14\times12},
  \qquad \operatorname{rank}L=8.
\]
Before the output permutation, the eight systematic rows form an identity
submatrix; row permutation preserves rank irrespective of the additional rows.
Hence the image is an affine subspace of dimension eight and has size
256; again $t=4$ and $k=6$.

\paragraph{C2.4.3 Hamming ball via truncation.}
Here $m=n=12$.  The generator chooses a hidden ordering of the input
coordinates by a uniform permutation $\sigma$ and scans them in that order.  If
$z_{\sigma(q)}$ denotes the retained bit at scan position $q$, then
\[
 z_{\sigma(q)}=x_{\sigma(q)}\wedge
 \mathbf 1\!\left[\sum_{r<q}z_{\sigma(r)}<3\right].
\]
The threshold predicate is maintained by three Boolean rails recording whether
at least one, two, or three set bits have appeared; they are updated after each
input.  Before a scan, let $A_h$ mean that at least $h$ earlier input bits were
one, for $h=1,2,3$, with all three rails initially zero.  For the next scanned
bit $x$, first emit $z=x\wedge\neg A_3$; retain the old rails while emitting,
in the order $h=3,2,1$,
\[
  A'_1=A_1\vee x,\qquad
  A'_h=A_h\vee(A_{h-1}\wedge x)\quad(h=2,3).
\]
The canonical OR expansion above determines the individual gates.  Thus only
the first three set bits survive.
After an output permutation and shift, the image is an affine copy of the
radius-three Hamming ball
\[
  B_{12}(0,3)=\{z\in\F_2^{12}:\operatorname{wt}(z)\leq3\}.
\]
Every circuit output has weight at most three before hiding, while every word
of weight at most three is unchanged by truncation when used as the input.
Consequently
\[
  |\operatorname{Im}F|
  =\sum_{i=0}^{3}\binom{12}{i}
  =1+12+66+220=299.
\]
The threshold parameters are $t=4$ and $k=6$.

\begin{figure}[H]
  \centering
  \begin{tikzpicture}[x=0.72cm,y=0.72cm]
    \node[font=\scriptsize,anchor=east] at (-0.35,0.5) {input};
    \node[font=\scriptsize,anchor=east] at (-0.35,-0.5) {retained};
    \foreach \i/\a/\b in {0/1/1,1/0/0,2/1/1,3/1/1,4/0/0,5/1/0,6/1/0,7/0/0} {
      \draw (\i,0.15) rectangle ++(0.62,0.62);
      \node[font=\scriptsize] at ({\i+0.31},0.46) {$\a$};
      \draw (\i,-0.85) rectangle ++(0.62,0.62);
      \node[font=\scriptsize] at ({\i+0.31},-0.54) {$\b$};
    }
    \draw[diagram arrow] (0.31,0.12)--(0.31,-0.20);
    \draw[diagram arrow] (2.31,0.12)--(2.31,-0.20);
    \draw[diagram arrow] (3.31,0.12)--(3.31,-0.20);
    \draw[diagram arrow,dashed] (5.31,0.12)--(5.31,-0.20);
    \draw[diagram arrow,dashed] (6.31,0.12)--(6.31,-0.20);
    \node[font=\scriptsize,align=left,anchor=west] at (8.25,-0.54)
      {later 1s are deleted};
  \end{tikzpicture}
  \caption{An illustrative scan for C2.4.3.  The first three 1s in the hidden
  coordinate order are retained; the generated circuit has 12 coordinates.}
  \label{fig:image-size-truncation}
\end{figure}
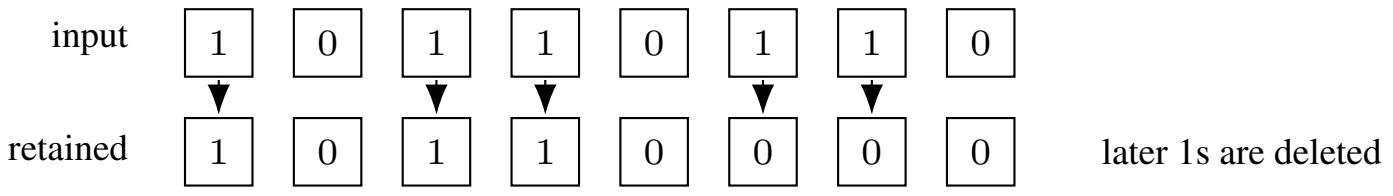

\paragraph{C2.4.4 Separable nonlinear block encoding.}
A uniform permutation of the 12 inputs is sampled, and consecutive pairs in
that order form six blocks $(u_i,v_i)$, $1\leq i\leq6$.
Each block is encoded by the injective nonlinear map
\[
  \gamma(u_i,v_i)=(u_i,v_i,u_iv_i).
\]
Here $u_iv_i$ is multiplication in $\F_2$, implemented by an AND gate.
The six three-bit codewords are concatenated with six zero coordinates to give
24 outputs, which are then permuted and shifted.  Each block has four distinct
images and the blocks vary independently, so
\[
  |\operatorname{Im}F|=4^6=2^{12}=4096.
\]
Here $t=8$ and $k=10$.  The product coordinate $u_iv_i$ makes the map
non-affine, while the two systematic block coordinates preserve injectivity.

\paragraph{C2.4.5 Rank-at-most-one matrices over $\F_2$.}
A uniform permutation of the nine inputs is sampled; its first four entries,
in order, form $u\in\F_2^4$, and its last five form
$v\in\F_2^5$.  The 20 logical outputs are the row-major entries of the
outer-product matrix
\[
  M=uv^{\mathsf T},
  \qquad M_{ij}=u_iv_j.
\]
All pairs with $u=0$ or $v=0$ give the zero matrix.  Every nonzero binary
rank-one matrix determines a unique nonzero pair $(u,v)$, since the only
nonzero scalar in $\F_2$ is one.  Therefore
\[
  |\operatorname{Im}F|
  =1+(2^4-1)(2^5-1)=466.
\]
The output entries are finally permuted and complemented; $t=4$ and $k=6$.

\paragraph{C2.4.6 Grid independent-set image.}
Let $Q=P_3\square P_4$ be the $3\times4$ rectangular grid.  One shuffled input
bit $y_v$ is assigned to every grid vertex by a uniform input permutation, and
the vertices are processed in a second independent uniform permutation.  Thus
$m=n=12$.  When vertex $v$ is reached, the circuit sets
\[
  z_v=y_v\wedge
  \bigwedge_{\substack{u\in N_Q(v)\\u\text{ processed earlier}}}\neg z_u.
  \tag{15}
\]
where $N_Q(v)$ is the set of grid neighbors and the conjunction is one when
$v$ has no earlier neighbor.  The selected vertices $\{v:z_v=1\}$ are always
independent.  Conversely, for any independent set $I$, choosing
$y=\mathbf 1_I$ and inducting over the processing order gives
$z=\mathbf 1_I$.  The image before its final coordinate
permutation and shift is therefore exactly the family of independent sets of
$Q$.
For the canonical netlist, grid vertices have row-major indices.  At each
scanned vertex, earlier neighbors are considered in the fixed order up, down,
left, right; their already computed outputs are individually negated in that
order and then conjoined with $y_v$ from left to right.

Its cardinality is obtained by a column transfer.  Let
\[
  \mathcal M=\{a\in\{0,1\}^3:
    \text{no two consecutive coordinates of $a$ are one}\}
\]
be the vertically independent column masks.  With $D_1(a)=1$ for every
$a\in\mathcal M$, define, for $c=1,2,3$,
\[
  D_{c+1}(b)=\sum_{\substack{a\in\mathcal M\\
    \operatorname{supp}(a)\cap\operatorname{supp}(b)=\varnothing}}D_c(a).
\]
Then $\sum_{b\in\mathcal M}D_4(b)=227$.  Hence
$|\operatorname{Im}F|=227$, $t=3$, and $k=5$.

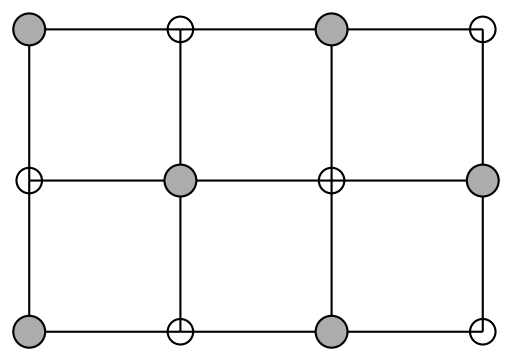
\begin{figure}[H]
  \centering
  \begin{tikzpicture}[x=1.0cm,y=1.0cm]
    \foreach \c in {0,...,3} {
      \foreach \r in {0,...,2} {
        \coordinate (g\c r\r) at (\c,{2-\r});
        \node[graph vertex] at (g\c r\r) {};
      }
    }
    \foreach \c in {0,...,2} {
      \foreach \r in {0,...,2}
        \draw[diagram edge] (g\c r\r)--(g\the\numexpr\c+1\relax r\r);
    }
    \foreach \c in {0,...,3} {
      \foreach \r in {0,...,1}
        \draw[diagram edge] (g\c r\r)--(g\c r\the\numexpr\r+1\relax);
    }
    \foreach \c/\r in {0/0,0/2,1/1,2/0,2/2,3/1}
      \node[circle,draw,fill=gray!65,inner sep=0pt,minimum size=6pt]
        at (g\c r\r) {};
  \end{tikzpicture}
  \caption{One independent set of the $3\times4$ grid in C2.4.6.  No two
  shaded vertices share an edge; every such set occurs as a circuit output.}
  \label{fig:image-size-grid}
\end{figure}

\newpage
\subsection{Shamir-Style TQBF Interactive Proof}
\label{sec:shamir-tqbf}

A true quantified Boolean formula (TQBF) is a Boolean circuit preceded by
universal and existential quantifiers.  A universal variable represents a
choice for which both Boolean values must succeed, whereas an existential
variable represents a response for which at least one value must succeed.
This task asks Merlin to certify the truth of such a formula.  The prover must
supply the polynomial messages below; merely supplying an existential Boolean
strategy is not accepted.  Following Shamir's arithmetization of
TQBF \cite{Shamir1992IP}, Merlin must maintain a sequence of low-degree
polynomial claims, each fixed before Arthur samples the random point used to
check it.

The public formula is
\[
  \Phi=Q_0x_0\;Q_1x_1\cdots Q_{n-1}x_{n-1}:
  C(x_0,\ldots,x_{n-1}),
  \qquad Q_i\in\{\forall,\exists\}.
  \tag{16}
\]
The circuit is acyclic and uses Boolean constants and the gates NOT, AND, OR,
XOR, and ITE.  Here $\operatorname{ITE}(c,a,b)$ returns $a$ when $c=0$ and $b$
when $c=1$.  All protocol arithmetic is over the finite field
\[
  \F_p,\qquad p=2^{61}-1,
\]
using the gate polynomials
\[
\begin{aligned}
 \neg a&=1-a, & a\wedge b&=ab,
 &a\vee b&=a+b-ab,\\
 a\oplus b&=a+b-2ab,
 &\operatorname{ITE}(c,a,b)&=(1-c)a+cb.
\end{aligned}
\tag{17}
\]
These expressions agree with the Boolean gates on $\{0,1\}$, but are also
defined at Arthur's non-Boolean field challenges.

Let $\widehat C=P_n$ be the arithmetized circuit polynomial.  For a polynomial
$f$ and a variable $X_j$, write $f_0=f|_{X_j=0}$ and $f_1=f|_{X_j=1}$, and
define
\[
  \mathcal A_j(f)=f_0f_1,
  \qquad
  \mathcal E_j(f)=f_0+f_1-f_0f_1,
  \qquad
  \mathcal R_j(f)=(1-X_j)f_0+X_jf_1.
  \tag{18}
\]
The first two operators arithmetize universal conjunction and existential
disjunction.  The third is the unique degree-one interpolation in $X_j$ that
preserves the two Boolean restrictions; it is called
\emph{multilinearization in $X_j$}.  Eliminating variables from the inside
out, for $i=n,n-1,\ldots,1$ set
\[
  G_{i-1}=
  \begin{cases}
    \mathcal A_{i-1}(P_i),&Q_{i-1}=\forall,\\
    \mathcal E_{i-1}(P_i),&Q_{i-1}=\exists,
  \end{cases}
  \qquad
  P_{i-1}=\mathcal R_0\mathcal R_1\cdots
           \mathcal R_{i-2}(G_{i-1}),
  \tag{19}
\]
where the product of reduction operators is empty when $i=1$.  The operators
are applied from right to left, so $\mathcal R_0$ is the outermost remaining
operator and is checked first.  On Boolean
outer variables this recurrence is exactly quantified evaluation, so the
formula is true precisely when the constant $P_0$ equals one.  Every generated
instance is promised to be true.

Merlin begins by claiming $P_0=1$.  Arthur then traverses (19) from the outside
toward the circuit.  At stage $i$, the public round order is
\[
\begin{array}{c|l}
 i=1 & Q_0X_0\\
 i=2 & \mathcal R_0,\;Q_1X_1\\
 i=3 & \mathcal R_0,\;\mathcal R_1,\;Q_2X_2\\[-1mm]
 \vdots&\vdots\\
 i=n & \mathcal R_0,\ldots,\mathcal R_{n-2},\;Q_{n-1}X_{n-1}.
\end{array}
\tag{20}
\]
Thus there are $R=n(n+1)/2$ polynomial rounds.  At each stage, the true current
multivariate polynomial is obtained by starting from $G_{i-1}$, retaining the
reduction operators not yet checked in (20), and substituting every coordinate
sampled in earlier rounds.  The current claim $v$ is its value after the
remaining displayed operators are applied.  If the next round concerns $X_j$,
the honest continuation $h^*(T)$ is the restriction of that polynomial with
$X_j=T$ and all earlier challenges fixed.  Merlin submits exactly $d_q+1$
coefficients of a univariate $h(T)$ in increasing-power order, including
trailing zeros, where $d_q$ is the announced degree bound.  Arthur computes
$h_0=h(0)$ and $h_1=h(1)$ and checks exactly one anchor equation:
\[
\begin{array}{rcll}
 v&=&(1-a_j)h_0+a_jh_1,&\text{reduction round},\\
 v&=&h_0h_1,&Q_j=\forall,\\
 v&=&h_0+h_1-h_0h_1,&Q_j=\exists.
\end{array}
\tag{21}
\]
The verifier stores a field assignment vector
$a=(a_0,\ldots,a_{n-1})$, initially all zero.  At any round, $a_j$ denotes the
most recently sampled challenge assigned to $X_j$; before the first round that
concerns $X_j$, its initial value is immaterial because the quantifier anchor
does not use it.  Each successful reduction or quantifier round for $X_j$
refreshes this coordinate with the new challenge.
Only after this equation holds does Arthur sample an independent uniform
$r\in\F_p$, replace $a_j$ by $r$, and replace the claim by $h(r)$.  For all
stages below the final circuit stage, reduction and quantifier rounds have
degree bounds two and one, respectively.  At the final stage, let $\Delta_j$
be the syntactic individual degree of the circuit output in $X_j$.  Input
$X_j$ has degree vector $e_j$; constants have degree zero; NOT copies its input
vector; AND, OR, and XOR add their two vectors.  If the individual-degree
vectors of the condition, zero branch, and one branch of an ITE are
$d_c,d_0,d_1$, respectively, its syntactic vector is
\[
  d_{\operatorname{ITE}}
   =\max\{d_0,\ d_c+d_0,\ d_c+d_1\},
\]
where addition and the maximum are coordinatewise.  This is the envelope of
the unreduced expression $(1-c)a+cb$.  The final-stage reduction bound for
$X_j$ is
$2\Delta_j$, while the final quantifier bound is $\Delta_{n-1}$.  After the
last challenge, Arthur directly evaluates the arithmetized public circuit at
the accumulated field point and compares it with the final claim.  A malformed
message or failed equation terminates the one-shot interaction, whose total
wall-clock budget is 600 seconds.

\begin{lemma}[Perfect completeness and statistical soundness]
\label{lem:tier2-tqbf}
An honest Merlin is accepted with probability one on every generated
instance.  If $\Phi$ is false, the probability that any Merlin makes Arthur
accept the initial claim one is at most
\[
  \frac{D}{p},
  \qquad D=\sum_{q=0}^{R-1}d_q,
  \tag{22}
\]
where $d_q$ is the public degree bound in round $q$.  For the six generated
profiles, the values of $(n,R,\max_qd_q,D)$ are
\[
\begin{array}{c|rrrrrr}
 &\mathrm{C2.5.1}&\mathrm{C2.5.2}&\mathrm{C2.5.3}&
  \mathrm{C2.5.4}&\mathrm{C2.5.5}&\mathrm{C2.5.6}\\ \hline
n&18&24&24&44&44&44\\
R&171&300&300&990&990&990\\
\max_qd_q&2&2&2&6&10&10\\
D&324&576&576&2008&2060&2060
\end{array}
\]
and hence the acceptance probability is less than $2^{-49}$.
\end{lemma}

\begin{proof}
For perfect completeness, Merlin submits the actual univariate restriction of
the appropriate polynomial in (19).  Equations (18)--(21) then hold
identically, and the last restriction evaluates to the public circuit
polynomial.

For soundness, maintain the claim defined by the corresponding true polynomial
at Arthur's accumulated point.  Whenever Merlin passes an anchor while the
submitted continuation differs from the true continuation, their difference
is a nonzero univariate polynomial of degree at most $d_q$.  It vanishes at the
fresh challenge with probability at most $d_q/p$.  Unless this happens, the
false claim persists to a later round and is eventually contradicted by the
direct circuit evaluation.  A union bound over all rounds proves (22).
\end{proof}

The first three cases are compiled from finite-state branching programs.
For a state set $S$, transition functions $\delta_i^0,\delta_i^1$, and terminal
accepting set $A$, the generator constructs the ITE circuit backward from
\[
  F_{n,s}=\mathbf 1[s\in A],
  \qquad
  F_{i,s}=\operatorname{ITE}
  \bigl(x_i,F_{i+1,\delta_i^0(s)},F_{i+1,\delta_i^1(s)}\bigr),
  \tag{23}
\]
and publishes $F_{0,s_0}$.  Seed-derived double negations or conjunctions with
one are appended as semantic padding.  Specifically, 16 independent bits from
the case-separated seeded stream choose, at each step, either
$z\mapsto\neg\neg z$ or $z\mapsto z\wedge1$.  Constants and all state-layer ITE
gates are emitted as distinct DAG nodes.  This padding changes the circuit
representation but not its Boolean or field polynomial.

\paragraph{C2.5.1 Width-four alternating response game.}
There are nine consecutive pairs $\forall u_i\exists e_i$, hence $n=18$.
The four states are \emph{ready}, \emph{remember 0}, \emph{remember 1}, and an
absorbing \emph{reject} state; the initial state is ready.  At a universal
layer, ready moves to the selected remember state and every other state moves
to reject.  At an existential layer, remember 0 moves to ready on input zero
and to reject on input one, while remember 1 does the reverse; ready and reject
move to reject.  Thus the universal choice $u_i$ selects
the corresponding remember state.  The existential choice returns to ready
exactly when $e_i=u_i$; the other choice enters reject.  Only ready is
terminally accepting, so $e_i=u_i$ for every pair is a winning strategy.  The
published circuit has 106 gates, 171 polynomial rounds, and maximum round
degree two.

\begin{figure}[H]
  \centering
  \begin{tikzpicture}[x=1cm,y=1cm,
    tqstate/.style={circle,draw,minimum size=12mm,align=center,font=\scriptsize}]
    \node[tqstate] (r) at (0,0) {ready};
    \node[tqstate] (m0) at (3.0,1.25) {remember\\$0$};
    \node[tqstate] (m1) at (3.0,-1.25) {remember\\$1$};
    \node[tqstate] (bad) at (6.2,0) {reject};
    \draw[diagram arrow] (r) to[bend left=10]
      node[above left,font=\scriptsize] {$u_i=0$} (m0);
    \draw[diagram arrow] (r) to[bend right=10]
      node[below left,font=\scriptsize] {$u_i=1$} (m1);
    \draw[diagram arrow] (m0) to[bend left=10]
      node[below right,font=\scriptsize] {$e_i=0$} (r);
    \draw[diagram arrow] (m1) to[bend right=10]
      node[above right,font=\scriptsize] {$e_i=1$} (r);
    \draw[diagram arrow] (m0) -- node[above right,font=\scriptsize] {$e_i=1$} (bad);
    \draw[diagram arrow] (m1) -- node[below right,font=\scriptsize] {$e_i=0$} (bad);
    \draw[diagram arrow] (bad) to[loop right]
      node[right,font=\scriptsize] {later choices} (bad);
  \end{tikzpicture}
  \caption{One universal--existential pair in C2.5.1.  A matching response
  restores the ready state; a mismatch is irreversible.}
  \label{fig:tqbf-width-four}
\end{figure}
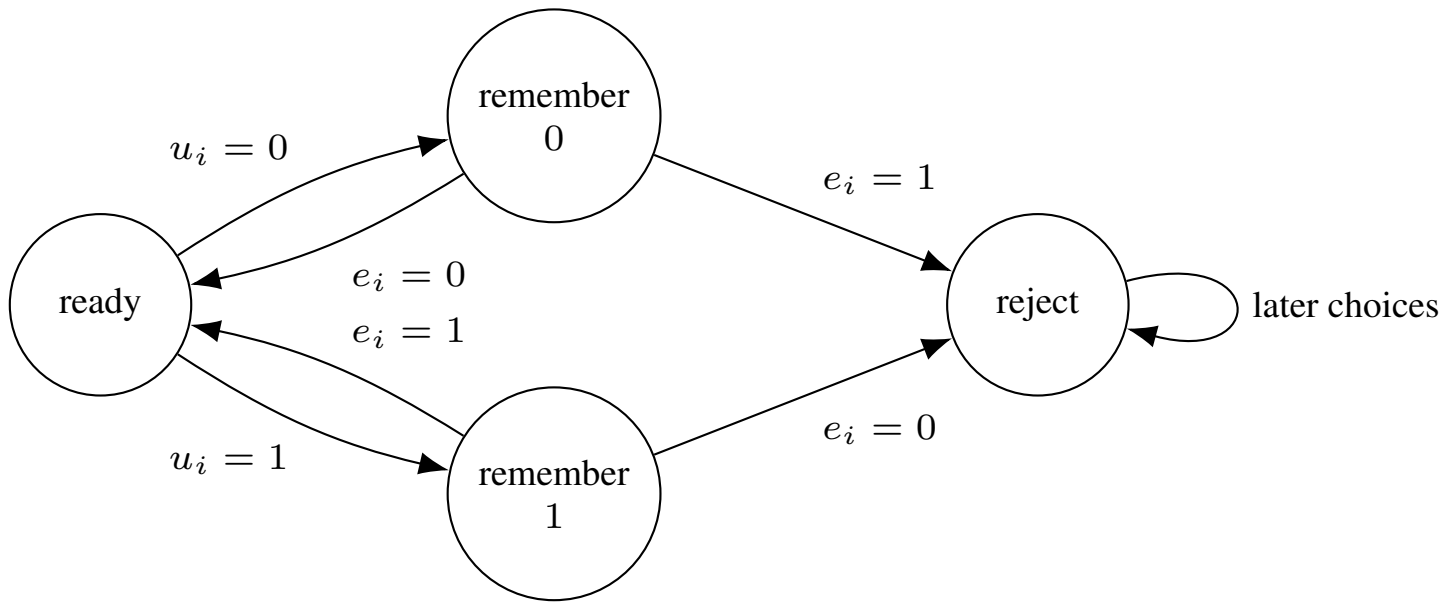

\paragraph{C2.5.2 Memory-five recurrence response game.}
This case has twelve pairs $\forall u_i\exists e_i$ and $n=24$.  A state
label $s\in\{0,\ldots,63\}$ encodes a five-bit register
$m=s\bmod32$ and a bit $b=\lfloor s/32\rfloor$; there is also an absorbing sink
$\bot$, giving 65 states.  At the beginning of a universal layer the reachable
labels have $b=0$, initially $s=0$.  Reading $u_i$ changes the label to
$m+32u_i$, thereby retaining both the old register and the pending universal
bit.  At the following existential layer, the unique non-sink response is
\[
  e_i=u_i\oplus\operatorname{parity}(m),
  \qquad
  m\leftarrow ((m\ll1)\mathbin{\vert}u_i)\bmod 32,
  \tag{24}
\]
where $\operatorname{parity}(m)$ is the XOR of the five register bits; the
register update occurs only after $e_i$ has been checked.  Any other response
enters the sink, while all labels $0,\ldots,31$ are
terminally accepting.  The generated ITE circuit has 1,594 gates and 300
polynomial rounds; every round has degree at most two.

\paragraph{C2.5.3 Rank-five affine-syndrome game.}
The prefix consists of twelve universal bits followed by twelve existential
bits.  The running state, called the \emph{syndrome}, is a vector
$\sigma\in\F_2^5$ initialized to zero.  The generator includes the five
standard basis columns, uniformly permutes all 31 nonzero vectors of $\F_2^5$,
and appends the first seven of that order to obtain
$a_0,\ldots,a_{11}$.  When the current bit $z_i$ equals one, the state updates
as $\sigma\leftarrow\sigma\oplus a_i$; a zero leaves it unchanged.  The
existential block uses the same multiset in an independent uniform permutation
$a_{\pi(0)},\ldots,a_{\pi(11)}$.  Therefore, for every universal assignment
$x$, choosing $y_j=x_{\pi(j)}$ adds the same syndrome a second time and returns it to zero,
the sole accepting state.  The public circuit has 802 gates, 300 polynomial
rounds, and maximum round degree two.

The last three cases share a nonlinear relation on 21 universal bits $x$ and
21 existential bits $y$.  Independent uniform coordinate permutations divide
both vectors into two matching blocks.  Within a block of size $k$, write
$a_i=x_{\pi_{\rm in}(i)}$ and $b_i=y_{\pi_{\rm out}(i)}$ for the permuted input
and output coordinates.  A single-tap block is defined by
\[
\begin{aligned}
 T_0(a)&=\neg a_0,\\
 T_1(a)&=\neg a_1\oplus a_0,\\
 T_i(a)&=\neg a_i\oplus(a_{i-1}a_{i-2}),
 &&2\leq i<k.
\end{aligned}
\tag{25}
\]
A double-tap block additionally XORs $a_{i-3}a_{i-4}$ into $T_i$ for every
$i\geq4$.  In either form,
$T_i(a)=\neg a_i\oplus g_i(a_0,\ldots,a_{i-1})$; the inputs can therefore be
recovered uniquely in order from $T(a)$.  Each block map and their direct
product are bijections on the Boolean cube.

Let $F(x,y)$ be the conjunction of the 21 coordinate equalities
$b_i=T_i(a)$.  Each equality is implemented as
$\neg(b_i\oplus T_i(a))$.  The indicators are uniformly permuted and combined
by a balanced tree: consecutive pairs are conjoined, and an unpaired final
entry is carried unchanged to the next level.
Two final variables $u,e$ give the public quantified wrapper
\[
  W(x,y,u,e)=\operatorname{ITE}
  \bigl(e,\operatorname{ITE}(u,F(x,y),1),0\bigr),
  \tag{26}
\]
with prefix $\forall x\,\exists y\,\forall u\,\exists e$.  Eliminating $e$
returns $\operatorname{ITE}(u,F,1)$ exactly, and eliminating $u$ returns
$F\cdot1=F$ exactly.  Since every $x$ has the unique response $y=T(x)$, the
formula is true.  Moreover, after the 21 universal bits are fixed, the
$2^{21}$ assignments leave pairwise distinct residual predicates
$\mathbf1[y=T(x)]$.

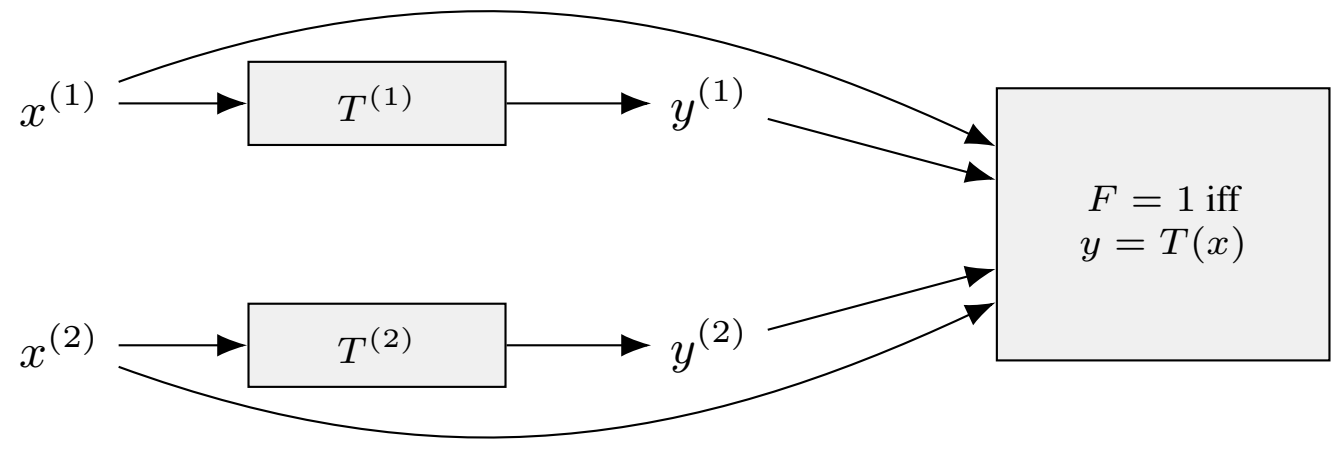
\begin{figure}[H]
  \centering
  \begin{tikzpicture}[x=1cm,y=1cm]
    \node[font=\small] (x1) at (-4.3,0.8) {$x^{(1)}$};
    \node[font=\small] (x2) at (-4.3,-0.8) {$x^{(2)}$};
    \node[circuit gate,minimum width=17mm] (t1) at (-2.2,0.8) {$T^{(1)}$};
    \node[circuit gate,minimum width=17mm] (t2) at (-2.2,-0.8) {$T^{(2)}$};
    \node[font=\small] (y1) at (0,0.8) {$y^{(1)}$};
    \node[font=\small] (y2) at (0,-0.8) {$y^{(2)}$};
    \node[circuit gate,minimum width=22mm,minimum height=18mm,align=center]
      (f) at (3.0,0) {$F=1$ iff\\$y=T(x)$};
    \draw[diagram arrow] (x1)--(t1);
    \draw[diagram arrow] (x2)--(t2);
    \draw[diagram arrow] (t1)--(y1);
    \draw[diagram arrow] (t2)--(y2);
    \draw[diagram arrow] (y1)--(f);
    \draw[diagram arrow] (y2)--(f);
    \draw[diagram arrow] (x1) to[out=20,in=155] (f);
    \draw[diagram arrow] (x2) to[out=-20,in=205] (f);
  \end{tikzpicture}
  \caption{The two disjoint reversible components underlying C2.5.4--C2.5.6.
  The public coordinate permutations hide the displayed block order but do not
  change the relation.}
  \label{fig:tqbf-blocked-map}
\end{figure}

\paragraph{C2.5.4 Hidden blocked reversible map.}
The two block sizes are 11 and 10, and both use the single-tap rule (25).
The common coordinate-equality, aggregation, and wrapper conventions above are
then applied.  The resulting
44-variable circuit has 195 gates and 990 polynomial rounds.  Its largest
announced round degree is six.

\paragraph{C2.5.5 Double-tap blocked reversible map.}
The block sizes are 12 and 9.  Both use the double-tap rule, so coordinates
$i\geq4$ contain the two quadratic terms
$a_{i-1}a_{i-2}$ and $a_{i-3}a_{i-4}$.  All other coordinate-equality,
aggregation, wrapper, and padding conventions are those of C2.5.4.  The public circuit has 234 gates and
990 polynomial rounds, with maximum round degree ten.

\paragraph{C2.5.6 Identity-cloaked double-tap reversible map.}
This case uses the same 12-and-9 block sizes and double-tap definition as
C2.5.5, with independently generated coordinate orders.  Before aggregation,
each equality indicator $z$ is wrapped once, using an independent uniform
choice among the polynomial identities
\[
 z\mapsto\neg\neg z,\qquad z\mapsto z\wedge1,\qquad
 z\mapsto z\vee0,\qquad z\mapsto\operatorname{ITE}(c,z,z),
\]
where the last template uses an independent uniform constant $c\in\{0,1\}$.
Every AND in the conjunction tree is replaced by
$\operatorname{ITE}(a,0,b)=ab$.  These identities hold for all field inputs,
not merely Boolean assignments.  The public syntactic degree schedule uses the
unsimplified gate-level recurrence stated above; for these constant-controlled
or equal-branch templates it has the same maximum bounds reported below.  The
published circuit has 277
gates, 990 polynomial rounds, and maximum round degree ten.

\newpage
\subsection{Hamiltonicity Bind--Challenge--Open}
\label{sec:hamiltonicity-bind-open}

The Hamiltonicity problem asks whether an undirected graph contains a cycle
that visits every vertex exactly once.  In this task the public graph is
promised to be Hamiltonian, and Merlin must prove that it contains such a
cycle without sending the cycle before Arthur chooses his challenge.  Each lane of
the protocol first binds Merlin to a relabeled graph.  Arthur then asks Merlin
to reveal either the relabeling or a Hamilton cycle in the committed graph.
Running 24 parallel lanes with independent challenge bits gives a
constant-round proof with two messages from Merlin and
soundness error at most $2^{-24}$ on a non-Hamiltonian input.  This is the
standard bind--challenge--open pattern underlying classical interactive proofs
for graph properties~\cite{goldreich-micali-wigderson-1991}; the present task
uses the committed graphs as explicit public objects and makes no
cryptographic hiding or zero-knowledge claim.

The generated instances satisfy the Hamiltonicity promise; the soundness
statement below applies more generally when the same protocol is run on an
arbitrary non-Hamiltonian input.  Formally, the input is a simple undirected
graph $G=([n],E)$ with $n\geq 3$.  A Hamilton cycle
is a permutation
\[
  C=(v_0,v_1,\ldots,v_{n-1})
\]
of $[n]$ such that $\{v_i,v_{i+1}\}\in E$ for every $i$, with indices taken
modulo $n$.  For a permutation $\pi\in S_n$, write $\pi(G)$ for the graph with
edge set $\{\{\pi(u),\pi(v)\}:\{u,v\}\in E\}$.

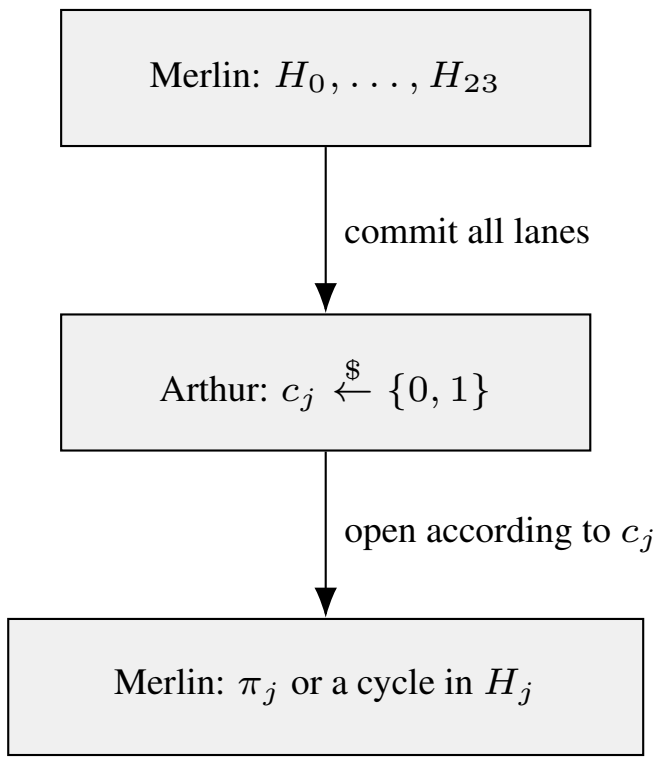
\begin{figure}[H]
  \centering
  \begin{tikzpicture}[node distance=11mm]
    \node[circuit gate,minimum width=35mm,minimum height=9mm] (commit)
      {Merlin: $H_0,\ldots,H_{23}$};
    \node[circuit gate,minimum width=35mm,minimum height=9mm,below=of commit]
      (challenge) {Arthur: $c_j\overset{\$}{\leftarrow}\{0,1\}$};
    \node[circuit gate,minimum width=42mm,minimum height=9mm,below=of challenge]
      (open) {Merlin: $\pi_j$ or a cycle in $H_j$};
    \draw[diagram arrow] (commit) -- node[right,font=\scriptsize]
      {commit all lanes} (challenge);
    \draw[diagram arrow] (challenge) -- node[right,font=\scriptsize]
      {open according to $c_j$} (open);
  \end{tikzpicture}
  \caption{The order of messages in the Hamiltonicity protocol.  The challenge
  bits are sampled only after the complete vector of committed graphs is
  fixed.}
  \label{fig:ham-bind-open-protocol}
\end{figure}

The interaction is a three-message public-coin protocol with two prover
messages, summarized in Figure~\ref{fig:ham-bind-open-protocol}.  In the first
message, ``commit'' means that Merlin sends complete encodings of all 24 graphs
and Arthur records this immutable vector before sampling any challenge bit; it
does not denote a cryptographic commitment.

\begin{enumerate}
  \item Merlin atomically commits to 24 simple undirected graphs
  $H_0,\ldots,H_{23}$, each with vertex set $[n]$.
  \item After all commitments have been recorded, Arthur samples independent
  uniform bits $c_0,\ldots,c_{23}\in\{0,1\}$.
  \item For every lane $j$, Merlin supplies one opening.  If $c_j=0$, the
  opening is a permutation $\pi_j$ and Arthur verifies
  $H_j=\pi_j(G)$.  If $c_j=1$, the opening is a Hamilton cycle of $H_j$ and
  Arthur checks that its vertices form a permutation of $[n]$ and that every
  consecutive pair is an edge.
\end{enumerate}

Arthur accepts only if all 24 lanes pass.  Every committed graph must have a
canonical simple-graph encoding on $[n]$; a self-loop, repeated edge,
out-of-range endpoint, malformed message, or failed check terminates the run
with rejection.  A 600-second wall-clock timer starts when Arthur loads the
protocol runtime and includes both prover messages, verification, and the
terminal response; expiration causes rejection.

\begin{lemma}[Completeness and soundness]
\label{lem:tier2-hamilton}
If $G$ is Hamiltonian and Merlin knows a Hamilton cycle of $G$, the protocol
has perfect completeness.  If $G$ is non-Hamiltonian, every prover is accepted
with probability at most $2^{-24}$ over Arthur's challenge bits.
\end{lemma}

\begin{proof}
For completeness, Merlin independently chooses permutations $\pi_j$, commits
to $H_j=\pi_j(G)$, and answers a zero challenge with $\pi_j$ and a one
challenge with the relabeled known cycle $\pi_j(C)$.  Both verification
conditions then hold.

For soundness, condition on the prover's internal randomness and its complete
first message before Arthur samples the challenges.
If a particular $H_j$ admits a valid zero-opening, then $H_j\cong G$ and is
therefore non-Hamiltonian.  It cannot also admit a valid one-opening.  Thus at
most one of the two values of $c_j$ can be answered successfully in each lane.
The 24 bits are independent and uniform, so the probability that all lanes
receive answerable challenges is at most $(1/2)^{24}$.
\end{proof}

Every test-point generator first constructs a semantically labeled graph
$G_0$ together with an explicit Hamilton cycle $C_0$ and checks the witness.
All random choices are drawn from a domain-separated deterministic stream
derived from the instance seed: bounded integers use rejection sampling, and
permutations use Fisher--Yates shuffling.  The generator finally samples a
uniform permutation $\rho$ of $[n]$ and publishes the graph whose edges are
$\{\{\rho(u),\rho(v)\}:\{u,v\}\in E(G_0)\}$.  This public relabeled graph is
the sole instance supplied to Merlin and Arthur.  Thus ``semantic labels''
below refer to the internal labels before this final relabeling; neither those
labels nor the planted cycle is supplied to the prover or verifier.

The last five cases use the same reduction from a satisfiable CNF formula to an
undirected Hamiltonian graph.  Variables, clauses, and literals within a
clause retain their generation order.  If variable $x_i$ has $k_i$
occurrences, create its \emph{occurrence row}
$u_{i,0},\ldots,u_{i,k_i}$.  If its $t$th occurrence is in clause $q_m$, add
both arcs between $u_{i,t}$ and $u_{i,t+1}$.  A positive occurrence adds the
directed detour
\[
   u_{i,t}\longrightarrow q_m\longrightarrow u_{i,t+1},
\]
whereas a negative occurrence adds
$u_{i,t+1}\to q_m\to u_{i,t}$.  From each endpoint of row $i$ add an arc to
each endpoint of row $i+1$.  A source $s$ has arcs to both endpoints of row
zero, both endpoints of the last row have arcs to a sink $t$, and the arc
$t\to s$ closes the traversal.  A satisfying assignment traverses row $i$
from left to right when $x_i$ is true and from right to left otherwise.  For
each clause, its first satisfied literal in generation order is selected, and
the traversal uses the corresponding detour to visit that clause vertex
exactly once.

Finally, each vertex $v$ of this directed graph is replaced by the undirected
three-vertex chain
\[
  v_{\mathrm{in}}-v_{\mathrm{mid}}-v_{\mathrm{out}},
\]
and every directed arc $u\to v$ is replaced by the undirected edge
$\{u_{\mathrm{out}},v_{\mathrm{in}}\}$.  Expanding every vertex of the
directed witness in the order
$v_{\mathrm{in}},v_{\mathrm{mid}},v_{\mathrm{out}}$ gives the planted
undirected Hamilton cycle.  Conversely, the middle vertex has degree two, so
any Hamilton cycle passes through every triple consecutively.  If the CNF has
$V$ variables, $M$ clauses, and $L$ literal occurrences, the intermediate
directed graph has
\[
 D=L+V+M+2\quad\text{vertices and}\quad
 A=4L+4(V-1)+5\quad\text{arcs};
\]
the undirected graph therefore has $3D$ vertices and $2D+A$ edges.
Figure~\ref{fig:ham-cnf-lift} shows the two local ingredients of the
construction.

\begin{figure}[H]
  \centering
  \begin{tikzpicture}[x=1cm,y=1cm]
    \node[graph vertex,label=below:{\scriptsize $u_{i}$}] (u0) at (0,0) {};
    \node[graph vertex,label=below:{\scriptsize $u_{i+1}$}] (u1) at (2.2,0) {};
    \node[graph vertex,label=above:{\scriptsize clause $q$}] (q) at (1.1,1.15) {};
    \draw[diagram arrow] ([yshift=1.5pt]u0.east) -- ([yshift=1.5pt]u1.west);
    \draw[diagram arrow] ([yshift=-1.5pt]u1.west) -- ([yshift=-1.5pt]u0.east);
    \draw[diagram arrow] (u0) -- (q);
    \draw[diagram arrow] (q) -- (u1);
    \node[font=\small,align=center] at (1.1,-0.9)
      {one position of a positive literal};

    \node[graph vertex,label=below:{\scriptsize $u_{\rm out}$}] (uo) at (4.6,0) {};
    \node[graph vertex,label=below:{\scriptsize $v_{\rm in}$}] (vi) at (6.5,0) {};
    \draw[diagram arrow] (uo) -- node[above,font=\scriptsize] {$u\to v$} (vi);
    \node[font=\large] at (7.4,0) {$\Longrightarrow$};
    \node[graph vertex,label=below:{\scriptsize $u_{\rm out}$}] (uo2) at (8.5,0) {};
    \node[graph vertex,label=below:{\scriptsize $v_{\rm in}$}] (vin) at (10.0,0) {};
    \node[graph vertex,label=below:{\scriptsize $v_{\rm mid}$}] (vm) at (11.2,0) {};
    \node[graph vertex,label=below:{\scriptsize $v_{\rm out}$}] (vout) at (12.4,0) {};
    \draw[diagram edge] (uo2) -- (vin) -- (vm) -- (vout);
    \node[font=\small,align=center] at (10.45,-0.9)
      {the undirected forced triple for $v$};
  \end{tikzpicture}
  \caption{Local pieces of the common CNF construction.  Left: a clause vertex
  can be inserted while traversing an occurrence in its satisfying direction.
  Right: replacing each directed vertex by a degree-two three-vertex chain
  preserves the directed order inside an undirected Hamilton cycle.}
  \label{fig:ham-cnf-lift}
\end{figure}

\paragraph{C2.6.1 Scaled randomly relabeled wheel.}
Before relabeling, the graph is the 1000-vertex wheel obtained from a cycle
$C_{999}$ by adding one universal hub.  The hub has label zero, and the rim
vertices $1,\ldots,999$ occur in that cyclic order.  Thus $n=1000$,
$|E|=1998$, and $C_0=(0,1,\ldots,999)$ is a Hamilton cycle.

\paragraph{C2.6.2 Scaled balanced complete bipartite graph.}
The semantic graph is $K_{128,128}$ with partite sets
$A=\{0,\ldots,127\}$ and $B=\{128,\ldots,255\}$.  Its edges are all pairs
$\{i,128+j\}$ for $0\leq i,j<128$.  It has $n=256$ and
$|E|=16{,}384$, and
$C_0=(0,128,1,129,\ldots,127,255)$ is a Hamilton cycle.

\paragraph{C2.6.3 Quiet-planted satisfiability lift.}
First sample 64 independent uniform Boolean values as the planted assignment.
For each candidate clause, sample an ordered triple of distinct variable
indices uniformly without replacement.  Independently choose one of the four
truth patterns
\[
 (1,1,1),\qquad(1,0,0),\qquad(0,1,0),\qquad(0,0,1)
\]
uniformly, and set each literal's sign so that its value under the planted
assignment is the corresponding bit.  Clauses are identified canonically by
sorting their signed literals by variable index; a duplicate is discarded and
resampled until 272 distinct clauses have been retained.  Thus every literal
position has marginal truth probability $1/2$; this is the precise sense in
which the planting has no one-position sign bias.  Here $V=64$, $M=272$, and
$L=816$, so the common construction gives $D=1{,}154$ directed vertices and an
undirected graph with 3,462 vertices and 5,829 edges.

\paragraph{C2.6.4 Row- and column-permuted affine-parity lift.}
Start with the $80\times80$ binary cyclic matrix $A$ whose row $r$ has support
\[
 \{r,r+1,r+2\}\pmod {80}.
\]
Its kernel is trivial: $z_r+z_{r+1}+z_{r+2}=0$ for every $r$ implies
$z_r=z_{r+3}$, and $\gcd(3,80)=1$ then makes every coordinate equal, whence
the equations force that common value to zero.  Sample uniform planted values
$z\in\mathbb F_2^{80}$.  Next sample a uniform column permutation $\tau$ and
then an independent uniform row permutation $\sigma$, and let equation $i$ contain
the three variables indexed by
$\tau(\sigma(i)),\tau(\sigma(i)+1),\tau(\sigma(i)+2)$ modulo 80, written in
increasing variable order.  Set its right-hand side to the parity of these
three coordinates of $z$.  The permuted matrix remains full rank, so $z$ is
the unique solution.

For each equation, enumerate the eight local Boolean assignments in binary
order.  For each of the four assignments having the wrong parity, append the
unique three-literal clause that is false on exactly that assignment.  This
gives $M=320$ clauses and $L=960$ literal occurrences.  The common construction
therefore has $D=1{,}362$ directed vertices and produces an undirected graph
with 4,086 vertices and 6,885 edges.

\paragraph{C2.6.5 Weak-signal planted satisfiability lift.}
Sample a uniform planted assignment on 40 variables.  Each candidate clause
uses an ordered uniform sample of three distinct variables.  With probability
$7/8$, choose a uniformly random two-element subset of its three literal
positions to be true under the planted assignment; with probability $1/8$,
choose a uniformly random singleton.  Literal signs are set accordingly.
After sorting signed literals by variable index, discard a duplicate clause
and resample; stop after retaining 320 distinct clauses.  Each literal position
therefore has marginal truth probability
$(7/8)(2/3)+(1/8)(1/3)=5/8$; ``weak-signal'' refers only to this per-position
bias.  With $V=40$, $M=320$, and $L=960$, the common construction has
$D=1{,}322$ directed vertices and yields an undirected graph with 3,966
vertices and 6,645 edges.

\paragraph{C2.6.6 Anchored equality/inequality ring lift.}
Sample a uniform planted assignment on variables $x_0,\ldots,x_{159}$ and an
independent uniform permutation $\sigma$ of their indices.  Regard
$x_{\sigma(0)},\ldots,x_{\sigma(159)}$ as a cyclic order, including the pair
from position 159 back to position zero.  For neighboring variables $x,y$
with equal planted values, append the pair of clauses
\[
  (\neg x\lor y)\land(x\lor\neg y)
\]
enforces $x=y$; for unequal planted values,
\[
  (x\lor y)\land(\neg x\lor\neg y)
\]
enforces $x\ne y$.  Process the 160 adjacent pairs in cyclic order, giving 320
binary clauses, and then append the unit literal $x_{\sigma(0)}$ or
$\neg x_{\sigma(0)}$ according to its planted value.  The ring relations and
this anchor determine the assignment uniquely.  Here $V=160$, $M=321$, and
$L=641$, so the common construction forms $D=1{,}124$ directed vertices and an
undirected graph with 3,372 vertices and 5,453 edges.

\paragraph{C2.6.7 Chained-auxiliary affine Hamilton lift.}
Let $A$ be the $23\times23$ binary cyclic matrix whose row $r$ has support
$\{r,r+1,r+2,r+3,r+4\}\pmod {23}$.  This matrix is nonsingular: subtracting
two consecutive homogeneous equations gives $z_r=z_{r+5}$; since
$\gcd(5,23)=1$, all coordinates are equal, and a weight-five equation then
forces their common value to zero.  Independently sample uniform column and row
permutations $\tau$ and $\sigma$, in that order.  Equation $i$ uses the ordered
tuple
\[
 \bigl(\tau(\sigma(i)),\tau(\sigma(i)+1),\ldots,
       \tau(\sigma(i)+4)\bigr)\pmod {23}.
\]
Next sample a uniform assignment to the 23 original variables and evaluate
these rows on it to obtain the right-hand sides.

To encode one equation
\[
  x_a\oplus x_b\oplus x_c\oplus x_d\oplus x_e=r,
\]
the generator introduces three fresh auxiliary variables, used only by that
equation, and the chain
\[
 y_1=x_a\oplus x_b,\qquad
 y_2=y_1\oplus x_c,\qquad
 y_3=y_2\oplus x_d,\qquad
 y_3\oplus x_e=r.
\]
For each ternary relation, enumerate the eight assignments to its three
variables and append the four three-literal clauses that individually exclude
the assignments violating the XOR equality.  Encode the final binary equality
or inequality analogously with the two clauses excluding its two incorrect
assignments.  Thus every equation contributes 14 clauses and 40 literal
occurrences.  Across all 23 equations this introduces 69 auxiliary variables,
numbered consecutively in groups of three after the 23 original variables as
the equations are processed.  Hence $V=92$, $M=322$, and $L=920$.  The common
construction has $D=1{,}336$
directed vertices and yields an undirected graph with 4,008 vertices and 6,721
edges.

\clearpage
\section{Tier-3: Real-World Tasks}
\label{app:tier3}

Tier-3 evaluates the production of standard, independently checkable evidence
artifacts from three established verification ecosystems.  Merlin submits a
Lean proof, a Dafny program, or a DRAT proof log; Arthur invokes the
corresponding pinned native checker on the exact digest-bound public instance.
All three tasks are deterministic and one-shot.  The names Arthur and Merlin
therefore describe the separation between expensive evidence construction and
comparatively direct verification, rather than an interactive
Arthur--Merlin protocol in the complexity-theoretic sense.  Every run has one
600-second wall-clock budget, including local computation, artifact upload,
and Arthur's terminal check.  Each task accepts exactly one raw artifact, not
an archive: the Lean source is limited to 262,144 bytes, the Dafny source to
1,048,576 bytes, and the DRAT proof to 16,777,216 bytes.

\subsection{Machine-Checked Mathematics}
\label{sec:tier3-machine-checked-mathematics}

Machine-Checked Mathematics asks Merlin to turn a finite mathematical
certificate into a proof term accepted by Lean.  The public instance contains
a common prelude, a case-specific module \texttt{Challenge.lean}, their
SHA-256 digests, and a structured copy of the instance data.  The authoritative
module defines transparent predicates
\texttt{Challenge.valid : List Nat -> Prop} and the closed proposition
\texttt{Challenge.MathematicalClaim : Prop}; the latter records
instance-level well-formedness independently of the witness.  The target is
\[
 \begin{aligned}
 \texttt{Challenge.Target}
 &:=\exists z:\texttt{List Nat},\quad \texttt{Challenge.valid}(z)\land {}\\
 &\hspace{6em}\texttt{Challenge.MathematicalClaim}.
 \end{aligned}
\]
Merlin submits one UTF-8 Lean source file whose only import is
\texttt{Challenge} and which exports
\texttt{AMBenchSolution.proof : Challenge.Target}.  Arthur checks the source
envelope and public hashes, compiles the public modules and submission with
Lean 4.19.0, checks the declaration at the exact target type, and rejects any
reported axiom dependency.  After compiling the participant module, an
Arthur-owned wrapper checks
\texttt{example : Challenge.Target := AMBenchSolution.proof} and executes
\texttt{\#print axioms AMBenchSolution.proof}; acceptance requires Lean's
report that this declaration does not depend on any axioms.  Constructs that
can bypass the intended kernel
judgment, including \texttt{sorry}, additional axioms, unsafe or native evaluation,
custom elaborators, host I/O, and additional imports, are forbidden.  Lean's
elaborator may construct the term, but acceptance rests on kernel type checking
of the final proof \cite{lean4-cade-2021}.  The list $z$ is embedded as the
existential witness in that proof term; no separate certificate file is
submitted.

The public Lean source is the canonical certificate specification.  The
descriptions below state the same relations mathematically.  A little-endian
radix-$B$ word with digits $d_0,\ldots,d_{k-1}$ denotes
$\sum_{j=0}^{k-1}d_jB^j$; a final partial word contains only its remaining
digits.  Matrices are serialized in row-major order, finite-field elements use
their standard representatives in $[0,p)$, and a packed permutation lists
$\pi(0),\pi(1),\ldots$ in that order.  Bit $i$ occupies bit
$i\bmod w$ of word $\lfloor i/w\rfloor$.  Conditions involving the witness
form \texttt{valid}; closed identities involving only published constants form
\texttt{MathematicalClaim}.  The phrase ``Lean checks'' may refer to either
part, whereas a stated generator invariant is not an additional acceptance
condition unless explicitly identified as such.  Random choices use a
deterministic case-seeded SHA-256 stream;
uniform and conditioned samples are taken over the stated finite sets.  The
generator constructs and verifies a private reference solution, which is not
disclosed to Merlin.

\begin{lemma}
For every case below, Arthur accepts exactly when the submitted source obeys
the source policy and the Lean 4.19.0 kernel derives an axiom-free term of the
published type \texttt{Challenge.Target}.
\end{lemma}
\begin{proof}
The Arthur-owned modules fix the transparent target and certificate predicate.
The submitted declaration is checked at that exact type, the axiom audit
excludes additional assumptions, and the public digests bind the judgment to
the published instance.
\end{proof}

\paragraph{C3.1.1 Dense binary linear certificate.}
Set $n=80$.  Over $\F_2$, independently generate a unit lower-triangular
matrix $L$ and a unit upper-triangular matrix $U$, with every off-diagonal
entry an independent uniform bit.  Form $A=LU$ and apply independent uniform
row and column permutations, replacing $A$ by
$P_{\mathrm{row}}AP_{\mathrm{col}}$.  Draw $x$ by rejection from uniform bit
vectors until $26\leq |x|\leq53$, define
$S_i=\{j:A_{ij}=1\}$, and publish these row supports together with $b=Ax$.
A certificate is a list
$z\in\{0,1\}^{80}$, and Lean checks
\[
 \sum_{j\in S_i}z_j\equiv b_i\pmod2
 \qquad(0\leq i<80).
\]
Both triangular factors are nonsingular, so the permutations preserve
invertibility and the accepted vector is unique.

\paragraph{C3.1.2 Shuffled CRT reconstruction certificate.}
Set the search floor $F=10^6$, window width $W=200{,}000$, and
$\Delta=\max\{2,\lfloor W/12\rfloor\}$.  Sample the initial cursor uniformly
from $\{F,\ldots,F+W-1\}$.  If $m_i$ is the least prime not below the current
cursor, update the cursor to
$m_i+1+D_i$, where $D_i$ is uniform on $\{0,\ldots,\Delta-1\}$; the cursor is
not required to remain in the initial window.  Repeat until twelve distinct
primes have been obtained.  Let $M=\prod_i m_i$ and sample
$x\in[\lfloor M/3\rfloor,\lfloor2M/3\rfloor]$ uniformly.  The pairs
$(m_i,r_i)$, where $r_i=x\bmod m_i$, are uniformly shuffled and reindexed
as $i=0,\ldots,11$ before publication; $M$ is also public.  In this order,
Merlin supplies
\[
 z=[x,q_0,\ldots,q_{11}],\qquad q_i=(x-r_i)/m_i.
\]
Lean checks $0<x<M$, the list lengths, $M=\prod_i m_i$, and every identity
$x=r_i+q_im_i$.  Primality, residue bounds, and pairwise coprimality are
generator invariants rather than conjuncts of either target predicate.  Under
these stated generator invariants, the Chinese remainder theorem makes the
accepted value unique in $[0,M)$; the Lean certificate itself establishes the
published divisibility identities, not primality or coprimality.

\paragraph{C3.1.3 Hidden line-graph root certificate.}
For a graph $X$, its line graph $L(X)$ has one vertex per edge of $X$, with two
vertices adjacent precisely when the corresponding edges meet.  Construct a
connected cubic bipartite graph $R$ on two sets of 32 vertices as the union of
three perfect matchings with cyclic shifts $0$, $1$, and a seed-selected
$s\in\{2,\ldots,31\}$, sampled uniformly.  Matching $t$ joins left vertex
$i$ to right vertex $i+t\bmod32$.  Let $H=L(R)$ and publish a uniformly
relabeled copy of $G=L(H)$.  Thus $H$ has 96 vertices and 192 edges, while
$G$ has 192 vertices, 576 edges, and degree six.

For each public vertex $v$ of $G$, the certificate assigns a candidate root
edge $\{a_v,b_v\}$ with $0\leq a_v<b_v<96$.  Encode
$c_v=96a_v+b_v$ and pack each consecutive group of eight codes as
$w_j=\sum_{t=0}^{7}c_{8j+t}(96^2)^t$, producing 24 natural numbers.  Lean
checks all bounds, that every recovered root vertex occurs exactly four times,
and that the two recovered pairs associated with the endpoints of every edge
of $G$ intersect.  The 96 four-cliques formed by the stars of the four-regular
graph $H$ give a reference reconstruction.  Since
$96\binom42=576=|E(G)|$, the checked local incidences account for every edge;
pairwise distinctness and the disjointness of pairs assigned to nonadjacent
vertices need not be separate predicates.  Indeed, the recovered four-regular
multigraph contributes exactly 576 endpoint-sharing incidences counted with
multiplicity, while the 576 distinct public edges must all occur among them.
A repeated recovered pair, a pair sharing both endpoints, or an additional
intersection on a public nonedge would reduce the number of available distinct
incidences below 576.  Hence the checked containment and count imply that two
public vertices are adjacent exactly when their recovered pairs intersect.
This is the incidence form of line-graph inversion
\cite{whitney-1932}.

\paragraph{C3.1.4 Anonymous affine-plane decoding certificate.}
Generate $AG(2,19)$, containing 361 points and 380 lines of 19 points each,
and independently relabel the points, the lines, and the order within each
line.  After shuffling, move one fixed horizontal reference line to the first
serialized position; this identifies one otherwise arbitrary line but exposes
neither an origin nor a coordinate parametrization.  Its parallel class is,
of course, recoverable from the published incidence relation as the set of
lines disjoint from it.  In hidden coordinates choose
\[
 f(X,Y)=\sum_{i+j\leq8}c_{ij}X^iY^j
\]
of total degree exactly eight.  The coefficients are serialized by increasing
$Y$-exponent $j=0,\ldots,8$ and, for fixed $j$, increasing $X$-exponent
$i=0,\ldots,8-j$.  Choose two horizontal fibres uniformly without replacement.
Corrupt all 19 readings on those fibres; on each other fibre, choose two
$X$-coordinates uniformly without replacement.  Every selected reading
receives an independent uniform error from $\F_{19}^{*}$.  Thus the published
value at public point $P$ is $y_P=f(P)+e_P$, serialized by public point index,
and the word contains exactly
$2\cdot19+17\cdot2=72$ errors.

The certificate contains a coordinate code $x+19y\in[0,361)$ for every public
point and an inverse table mapping each coordinate code back to its public
point index, each packed 19 digits per base-361 word; it then
contains the 45 triangularly ordered coefficients $c_{ij}$ and the 361 error
bits packed 19 per binary word.  These four portions contain 19, 19, 45, and
19 natural numbers.  Lean checks that the coordinate tables are inverse
permutations, every published block becomes a complete affine line, the
polynomial has exact total degree eight, and the weight-72 indicator marks
exactly the disagreements.  The evaluation code has distance
$19(19-8)=209$; since $2\cdot72<209$, the accepted evaluation word is unique
within the stated radius \cite{macwilliams-sloane-1977}.

\paragraph{C3.1.5 Anonymous modular column equivalence.}
Work over $\F_{1009}$.  Select 80 points uniformly without replacement from
$\mathbb P^1(\F_{1009})$ and map them by the quadratic Veronese embedding
\[
 (u:v)\longmapsto(v^2:uv:u^2)\in\mathbb P^2(\F_{1009}).
\]
For the right-hand configuration, apply pointwise a seed-derived element of
$\operatorname{PGL}_2(\F_{1009})$ obtained by starting at $I_2$ and applying
twelve elementary row operations.  Each operation independently and uniformly chooses one
of: swap two distinct uniform rows; add a uniform nonzero multiple of one row
to another; or multiply one row by a uniform nonzero scalar.  On each side,
independently construct a $3\times3$ transformation from $I_3$ by eighteen
operations with the same rule, apply independent uniform nonzero column
scalings, and uniformly permute the columns.  Publish the two
full-row-rank matrices $L,R\in\F_{1009}^{3\times80}$, omitting all canonical
parameters and hiding maps.

Merlin supplies $A,A^{-1}\in\F_{1009}^{3\times3}$, nonzero scales
$s_0,\ldots,s_{79}$, a column permutation $\pi$, and its inverse.  Each
permutation is packed ten base-80 digits per word, so the complete certificate
has $9+9+80+8+8=114$ natural numbers.  Lean checks both matrix-inverse
identities, all bounds, the permutation inverses, and
\[
 s_j(AL_{\bullet j})_r=R_{r,\pi(j)}\pmod {1009}
 \quad(0\leq j<80,\ 0\leq r<3).
\]
An accepted certificate is therefore an exact monomial equivalence of the two
anonymous projective column configurations.

\paragraph{C3.1.6 Composite cube-root certificate.}
Sample $k$ uniformly from $[2^{127},2^{128})$.  Starting from
$k^3+17k+29$ and $k^3+997k^2+19k+43$, respectively, advance to the least
integer not below the starting value and congruent to $5\bmod6$, then test
candidates at increments of six until one passes the strong probable-prime
screen for bases
\[
 2,3,5,7,11,13,17,19,23,29,31,37,
 41,43,47,53,59,61,67,71,73,79,83,89;
\]
Call the results $p$ and $q$, and set $n=pq$.  Let
$W=\lceil\operatorname{bitlen}(n)/8\rceil$.  Encode $p$ and $q$ as their
minimal big-endian byte strings, each preceded by its two-byte big-endian
length.  Append SHA-256 digests of a fixed domain separator, these two encoded
factors, and four-byte big-endian counters $0,1,\ldots$ until the material has
at least $W+16$ bytes.  If $H$ is the big-endian integer represented by the
complete concatenation and $L=\lfloor n/3\rfloor$, set
\[
 m=L+\bigl(H\bmod(n-L)\bigr).
\]
Publish $n$, exponent $3$, and $c=m^3\bmod n$.  The screen controls only the
generated profile and is not trusted by the acceptance relation.

The certificate is $[p,q,m]$.  Lean checks
\[
 1<p<n,\quad1<q<n,\quad pq=n,\quad m<n,
 \quad m^3\equiv c\pmod n.
\]
Acceptance therefore certifies a nontrivial factorization and an exact modular
cube root, irrespective of whether the factors are prime.

\paragraph{C3.1.7 Anonymous modular subset certificate.}
Let $h(\ell,i)$ be the big-endian integer represented by SHA-256 of a fixed
family domain separator, label $\ell$, and four-byte big-endian index $i$.
Starting with $S_{-1}=0$, set
\[
 s_i=S_{i-1}+1+\bigl(h(\mathtt{increment},i)\bmod1024\bigr),
 \qquad S_i=S_{i-1}+s_i.
\]
Thus $s_i>\sum_{j<i}s_j$.  Set
$M=S_{255}+1+(h(\mathtt{modulus\mbox{-}gap},0)\bmod2^{20})$.  Direct
evaluation with the generator's fixed domain separator gives
$2^{265}\leq M<2^{266}$; this bit-length profile is not asserted as a
consequence of superincreasingness alone.  Starting from
$2+\bigl(h(\mathtt{multiplier},0)\bmod(M-3)\bigr)$, scan cyclically through
$\{2,\ldots,M-1\}$ and take the first value $a$ with $\gcd(a,M)=1$.
The case seed selects a uniform permutation of the 256 residues
$as_i\bmod M$ and a uniformly sampled 128-subset of their public positions.
Publish the permuted weights $b_i$, $M$, and
$t=\sum_i b_ix_i\bmod M$.

The first eight certificate words encode all 256 selection bits in 32-bit
blocks; the next sixteen entries are
\[
 c_k=\sum_{i=16k}^{16k+15}b_ix_i\bmod M
 \qquad(0\leq k<16).
\]
Lean checks all bounds, recomputes every $c_k$, and requires their sum modulo
$M$ to equal $t$.  The chunk residues split the exact dot-product check into
sixteen bounded reductions; they do not add independent witness information.
Hamming weight 128 is a generation property, not an
acceptance condition.  Superincreasingness makes hidden integer subset sums
distinct and smaller than $M$, while multiplication by $a$ is a bijection;
hence the public modular relation has a unique bit-vector solution.

\paragraph{C3.1.8 Anonymous binary syndrome certificate.}
Use the primitive narrow-sense binary BCH code of length $2^9-1=511$ and
designed distance 57 over
$\F_{2^9}=\F_2[X]/(X^9+X^4+1)$.  Let $\alpha$ be the residue class of $X$,
with polynomial-basis coefficients stored from low to high bit.  For
coordinate $i$, concatenate the 9-bit representations of
\[
 \alpha^i,\alpha^{3i},\ldots,\alpha^{55i}
\]
to form its canonical 252-bit column.  The 28 odd syndromes suffice because
binary Frobenius gives $S_{2j}=S_j^2$.  For operation index
$r=0,\ldots,3023$, let
\[
 H_\ell(r)=\operatorname{int}_{\mathrm{be}}
   \bigl(\operatorname{SHA256}(D\mathbin\Vert\ell
          \mathbin\Vert\operatorname{be}_{8}(r))\bigr),
\]
where $D$ is a fixed case-specific ASCII domain separator, defined by the
released generator and followed by one zero byte, and
$\operatorname{be}_k(r)$ is the $k$-byte big-endian encoding of $r$.  The
label satisfies
$\ell\in\{\mathtt{row\mbox{-}left},\mathtt{row\mbox{-}right},
\mathtt{row\mbox{-}kind}\}$.  Set
$i_r=H_{\mathtt{row\mbox{-}left}}(r)\bmod252$.  Reduce
$H_{\mathtt{row\mbox{-}right}}(r)$ modulo 251 and increment the result when
it is at least $i_r$, thereby obtaining the distinct source row $j_r$.  Swap
rows $i_r,j_r$ exactly when
$H_{\mathtt{row\mbox{-}kind}}(r)\equiv0\pmod8$; otherwise XOR row $j_r$ into
row $i_r$.  Apply these operations in increasing $r$.  The fixed column
permutation sorts coordinate $i$ by
$\operatorname{SHA256}(D\mathbin\Vert\mathtt{column\mbox{-}order}
\mathbin\Vert\operatorname{be}_{4}(i))$, breaking ties by $i$.  The
case-seeded generator stream then selects a uniform 28-subset of canonical
coordinates.  Publish the transformed
columns in this public order and the transformed XOR syndrome.  Every column
and the syndrome are stored as eight little-endian 32-bit words, with the
unused top four bits zero.

Merlin's first sixteen words encode the 511 error bits, with 31 active bits in
the last word and its unused top bit required to be zero; sixteen further
entries give the corresponding population counts (popcounts).  Lean checks
these bounds and counts, total weight 28, and all 252 parity equations.  The
count entries support word-by-word checking and carry no independent witness
information.  Two accepted vectors would differ by a nonzero codeword of
weight at most 56, contradicting the BCH distance bound, so the accepted error
vector is unique \cite{macwilliams-sloane-1977}.

\subsection{Dafny-Verified Programs}
\label{sec:tier3-dafny}

Dafny-Verified Programs asks Merlin to implement a small software component.
Preconditions delimit legal calls; postconditions use side-effect-free
specification functions and predicates to characterize the required outputs.
The public instance
contains a module \texttt{Challenge}, which declares the data types and an
abstract \texttt{Candidate} trait with its contracts, and a module
\texttt{ArthurHarness}, which executes deterministic case-seeded examples.
Merlin submits one UTF-8 file defining exactly one module \texttt{Solution},
importing only \texttt{Challenge}, and providing a concrete class
\texttt{Candidate} that extends the public trait and implements every abstract
method.

Arthur first rejects any source that violates the policy below.  It then
concatenates the three modules, invokes Dafny 4.11.0 with its bundled Z3
backend to verify all proof obligations, compiles the same verified program to
Java, and runs the fixed harness.  Acceptance requires policy compliance, zero
verification errors and exactly one case-specific success marker.  The harness
constructs the submission through the public trait type, so the inherited
contract remains an obligation of every concrete override.  Assumptions,
axioms, external
implementations, verification-altering attributes, wildcard termination,
additional includes, I/O, and a participant-defined main method are forbidden.
The accepted artifact is checked by the standard Dafny verification pipeline
\cite{leino2010dafny}.

Arthur verifies \texttt{Combined.dfy} with a 120-second verification limit,
two verifier cores, warnings treated as errors, and Dafny's determinism
enforcement enabled.  Arthur then invokes the Java backend on the unchanged file
with verification disabled, because that same source has already passed the
preceding verifier call, and gives the execution subprocess a separate
120-second limit; warnings are again treated as errors.  The
evaluator-provided Java runtime executes only the generated harness; Dafny and
its bundled Z3 discharge the proof obligations.

The API and its quantified specification are fixed for each case.  A
deterministic SHA-256 counter stream only changes the concrete values embedded
in \texttt{ArthurHarness}; these examples bind executable behavior but do not
replace verification of the universal contracts.  In particular, changing the
case seed changes the runtime replay but not the method signature, contract, or
universal proof obligation.  Integers sampled from a
stated interval are independent and uniform unless a later conditioning rule
is given.

\begin{lemma}
If Arthur accepts a Dafny submission, every implemented method satisfies the
public inherited contract under Dafny's verification semantics, and the
verified Java program produces the required outputs on the digest-bound
harness instance.
\end{lemma}
\begin{proof}
The public trait fixes the method signatures and contracts used at every call
site.  Dafny verifies the concrete overrides against those inherited
obligations before compilation.  Arthur then executes the same merged source
and accepts only the unique expected marker, while the source policy excludes
constructs that would replace a proof obligation by an assumption or external
implementation.
\end{proof}

\paragraph{C3.2.1 Stable partition preservation.}
For an integer sequence $a$, let $N(a)$ be its negative entries in original
order and $P(a)$ its nonnegative entries in original order.  The required
method returns exactly
\[
 \operatorname{StablePartition}(a)=N(a)\mathbin{\Vert}P(a),
\]
where $\Vert$ denotes sequence concatenation.
The public module defines $N$ and $P$ recursively on the head and tail of the
sequence.  They are ordinary public Dafny functions and may be called by the
submitted implementation; the acceptance relation does not prescribe a
particular implementation strategy.  Equality to this specification fixes
both membership and relative order.  The harness independently samples twelve integers from $[-20,20]$ and
compares the executable output with the two stable subsequences.  Verification
must establish the equality for every finite input sequence; equality of
multisets alone is insufficient.

\paragraph{C3.2.2 Boundary-safe lower-bound search.}
For a nondecreasing sequence $a$ and key $k$, the method returns an index
$r\in[0,|a|]$ satisfying
\[
 \forall i\,(0\leq i<r\Longrightarrow a_i<k),\qquad
 \forall i\,(r\leq i<|a|\Longrightarrow k\leq a_i).
\]
These two quantified conditions also cover the boundary answers $0$ and
$|a|$.  The harness samples eighteen integers from $[-30,30]$, sorts them,
samples $k$ from $[-25,25]$, and compares the result with the first index whose
entry is at least $k$, or $18$ if no such entry exists.

\paragraph{C3.2.3 Capacity-constrained reservation update.}
Let $C\geq0$, let $u$ satisfy $0\leq u_i\leq C$ for every
$0\leq i<|u|$, and let $d\geq0$.  The returned Boolean
$\mathit{accepted}$ must satisfy
\[
 \mathit{accepted}\quad\Longleftrightarrow\quad
 \forall i\,(0\leq i<|u|\Longrightarrow u_i+d\leq C).
\]
If accepted, the returned sequence has entries $u_i+d$; otherwise it is
exactly the original sequence.  Hence no partial update is permitted.  The
harness samples $C\in[12,24]$, ten entries independently from $[0,C-3]$, and
$d\in[1,5]$, then computes both expected outputs from the displayed relation.

\paragraph{C3.2.4 Precedence-constrained earliest schedule.}
For any equal-length sequences $d$ and $p$, the legal-input predicate requires
$d_i>0$ and either $p_i=-1$ or $0\leq p_i<i$ for every index $i$.  The output
has $|s|=|d|$ and is the unique sequence
\[
 s_i=
 \begin{cases}
 0,&p_i=-1,\\
 s_{p_i}+d_{p_i},&0\leq p_i<i.
 \end{cases}
\]
for all $0\leq i<|d|$.  For the harness, sample fourteen durations
$d_0,\ldots,d_{13}$ independently from $[1,9]$.  Set $p_0=-1$; for each
$i=1,\ldots,13$, set $p_i=-1$ with probability $1/4$ and otherwise sample
$p_i$ uniformly from $\{0,\ldots,i-1\}$.  The strict index condition makes the
predecessor relation acyclic in the published order; the contract is therefore
earliest-start propagation on a parent forest.

\paragraph{C3.2.5 Dependency-order validation.}
The public specification defines a recursive
$\operatorname{Position}(\pi,v)$ and requires the input order $\pi$ to be a
duplicate-free permutation of $\{0,\ldots,n-1\}$; every edge endpoint must
also lie in this range.  The method returns true exactly when
\[
 \operatorname{Position}(\pi,u)<\operatorname{Position}(\pi,v)
 \qquad\text{for every published edge }(u,v).
\]
For the harness, set $n=10$ and $\pi=(0,1,\ldots,9)$.  For each
$v=1,\ldots,9$, add one edge $(u,v)$ with $u$ uniform in
$\{0,\ldots,v-1\}$.  When $v>2$, independently with probability $1/2$, draw
another source $u'$ uniformly from the same set and append $(u',v)$; the two
edges may coincide.  The resulting example is topologically ordered, while
the fixed runtime harness therefore exercises the true branch.  The verified
method must nevertheless return the correct Boolean for every well-formed
order and edge sequence, including sequences with backward edges, for which
the result is false.

\paragraph{C3.2.6 Audited priority-firewall batch evaluation.}
This case uses an abstract first-match rule table: a packet is already encoded
as a Boolean vector recording which rules match it.  For an arbitrary action
sequence and a packet matrix whose rows all have the same width as that
sequence, the contract applies the ordered first-match rule to every row and
audits the resulting decisions.  In the harness there are
sixteen ordered rules, each action being independently true with probability
$1/2$, and eleven packets represented by sixteen-bit match rows.  Before
conditioning, each match bit is independently true with probability $1/7$;
the first row is then forced to all false, and each later all-false row is
repaired by setting one uniform position true.
For packet $p$, the decision is $(-1,\mathrm{false})$ when no rule matches and
otherwise records the least matching index $j$ and action $a_j$.  In addition
to all eleven decisions, the method returns a sixteen-entry audit vector
\[
 h_j=|\{p:0\leq p<|\mathit{decisions}|,
          \ \text{decision }p\text{ uses rule }j\}|.
\]
The public recursive function \texttt{CountPrefix} defines these counts from
the decision sequence, coupling the functional result and its audit log.

\paragraph{C3.2.7 Reference-counted handle-table preservation.}
A state consists of an $m\times r$ Boolean matrix $R$ and counts $c_j$.  It is
valid exactly when every row has width $r$ and
\[
 c_j=|\{i:R_{ij}=\mathrm{true}\}|\qquad(0\leq j<r).
\]
Calls require $0\leq i<m$ and $0\leq j<r$.  Acquiring an unheld pair $(i,j)$
changes only $R_{ij}$ to true; acquiring an already held pair is rejected
without changing the state.  Releasing a held pair changes only $R_{ij}$ to
false, whereas releasing an unheld pair is rejected unchanged.  Every accepted
update must re-establish the complete count invariant.  The harness uses
$m=8,r=10$ and samples each matrix entry true with probability $1/4$.  It then
selects an acquire cell uniformly from all $mr$ cells, forces that cell false,
and computes the initial counts from the resulting matrix.  After the acquire,
it releases the first true cell in row-major order.  Expected matrices and
column counts are recomputed after each exact update.

\paragraph{C3.2.8 Free-list memory-pool preservation.}
A pool state is $(C,f,O)$, where $C\geq0$, $f$ is a duplicate-free sequence of free
block IDs and $O$ is the set of owned IDs.  Every ID lies in $[0,C)$ and
\[
 b\in O\quad\Longleftrightarrow\quad b\notin f
 \qquad(0\leq b<C).
\]
Allocation returns an acceptance flag, a block integer, and a next state.  If
$f$ is empty it rejects and preserves the state; the block result is
unconstrained by the public contract, and Arthur makes no rejected-branch
comparison involving that field.  Otherwise it accepts, returns $f_0$,
removes that entry, and adds the block to $O$.  Release is legal only under the
precondition $b\in O$; it removes $b$ from $O$ and prepends it to $f$.  The harness fixes
$C=6$, $f=[0,1,2]$, and $O=\{3,4,5\}$, allocates block zero, and then releases
a uniformly selected block from $\{3,4,5\}$.  Both operations must preserve
the quantified partition invariant, not merely return the harness values.

\subsection{DRAT-Certified Bounded Hardware Safety}
\label{sec:tier3-drat-hardware}

This task asks Merlin to certify that a sequential Boolean circuit cannot
reach its designated bad output within a public time horizon.  The circuit is
published as a standard sequential ASCII AIGER file: its combinational network
uses literals, two-input AND gates, and complemented edges, while latches carry
state between frames \cite{biere-2007-aiger}.  It has exactly one ordinary
AIGER output, named \texttt{bad}, and no other output or property section.  For
horizon $H$, the public
bounded-model-checking formula is
\[
 F_H=\operatorname{Init}(s_0)\land
 \bigwedge_{t=0}^{H-1}\operatorname{Trans}(s_t,i_t,s_{t+1})\land
 \bigvee_{t=0}^{H}\operatorname{Bad}(s_t,i_t).
\]
Thus the unrolling contains $H+1$ independently unconstrained input vectors
$i_0,\ldots,i_H$; the final vector is used only if the bad output depends
combinationally on an input.
All eight circuits are safe, so $F_H$ is unsatisfiable.  Merlin submits one
standard ASCII or binary DRAT proof.  Arthur writes the digest-bound public
DIMACS and the submitted proof into a fresh directory and accepts only when a
pinned \texttt{drat-trim} executable at commit
\nolinkurl{2e3b2dc0ecf938addbd779d42877b6ed69d9a985} derives the empty clause.
The SAT solver used to discover a proof is not part of the trusted boundary
\cite{Wetzler2014DRATTrim}.

The submitted proof is a nonempty, uncompressed byte string of at most
16,777,216 bytes.  Arthur imposes no separate line-count or proof-step limit
and does not pre-parse the artifact as text: the same raw bytes are passed to
\texttt{drat-trim}, which recognizes standard ASCII and binary DRAT and
supports ordinary deletion records.  In the fresh directory Arthur runs
\texttt{drat-trim bound.cnf proof.drat -w} with a 180-second subprocess
timeout.  The terminal check accepts exactly when the process exits with code
zero and its combined output contains \texttt{s VERIFIED}.  This checker call
runs inside the task-wide envelope of two CPU cores, 2,048 MiB of memory, and
2,048 MiB of disk.

The DIMACS file is produced by one fixed compiler.  If $M$ is the greatest
AIG variable index, write $x_{t,v}$ for DIMACS variable $tM+v$ and map an
AIGER literal $\ell$ by
\[
 L_t(0)=\bot,\qquad L_t(1)=\top,\qquad
 L_t(2v)=x_{t,v},\qquad L_t(2v+1)=\neg x_{t,v}.
\]
Every latch has a specified initial value in $\{0,1\}$.  For every AIGER AND
whose output and input literals are $z,a,b$, respectively, and every frame
$0\leq t\leq H$, the compiler adds
\[
 (\neg L_t(z)\lor L_t(a)),\qquad
 (\neg L_t(z)\lor L_t(b)),\qquad
 (L_t(z)\lor\neg L_t(a)\lor\neg L_t(b)).
\]
For every latch equation $q'=e$ and $t<H$, it adds the two clauses for
$L_{t+1}(q)\leftrightarrow L_t(e)$; frame zero receives one unit clause per
initial latch.  Finally, one clause contains $L_t(\mathit{bad})$ for every
frame.  During emission, a clause containing $\top$ or a complementary literal
pair is omitted, $\bot$ and repeated literals are deleted, and no other
simplification or variable reindexing is performed.  Hence the DIMACS header
declares $(H+1)M$ variables, including unused frame copies.

The public instance binds both the AIGER and DIMACS bytes by SHA-256 and
publishes the exact variable and clause counts.  To sample a parameter below
$B$, the generator hashes a fixed domain separator, the case seed, and an
eight-byte big-endian counter.  A 256-bit digest is accepted only below the
largest multiple of $B$ smaller than $2^{256}$, and its residue modulo $B$ is
returned; otherwise the counter advances and sampling repeats.

\begin{lemma}[Bounded-safety meaning of an accepted proof]
Assuming the correctness of the AIGER parser and fixed compiler, the SHA-256
bindings, and the pinned DRAT checker, an accepted artifact proves that the
published AIGER circuit has no initialized trace of length at most $H$ whose
bad output is true.
\end{lemma}
\begin{proof}
The literal map preserves constants and complementation.  A topological
induction over the AND gates in each frame shows that their three Tseitin
clauses reproduce the combinational AIG semantics, and the two latch clauses
reproduce each next-state equation.  Induction over frames then gives a
correspondence between satisfying assignments of $F_H$ and initialized AIG
traces reaching bad by frame $H$.  An accepted DRAT derivation proves the
digest-bound DIMACS formula unsatisfiable, excluding every such trace.
\end{proof}

\paragraph{C3.3.1 FIFO occupancy safety.}
Sample the FIFO depth $D$ uniformly from $\{4,5\}$ and the horizon $H$ from
$\{20,21,22,23\}$.  Latches $o_0,\ldots,o_{D-1}$ initially equal zero and
encode occupancy $k$ as the unary prefix $1^k0^{D-k}$.  With public inputs
$p=\mathit{push}$ and $r=\mathit{pop}$, define
\[
 P=p\land\neg r\land\neg o_{D-1},\qquad
 R=r\land\neg p\land o_0.
\]
If $P$ holds, set $o'_0=1$ and $o'_j=o_{j-1}$ for $j>0$; if $R$ holds, set
$o'_j=o_{j+1}$ for $j<D-1$ and $o'_{D-1}=0$; otherwise hold every latch.  The
simultaneous request $p=r=1$ therefore leaves the abstract occupancy unchanged.
The bad output is
\[
 \bigvee_{j=1}^{D-1}(o_j\land\neg o_{j-1}).
\]
Both shifts preserve the prefix form, so the initialized circuit never raises
bad.

\paragraph{C3.3.2 FIFO underflow safety.}
Use the same unary FIFO transition with independent parameters
$D$ and $H$ sampled uniformly from $\{10,11,12\}$ and
$\{28,29,30,31,32\}$, respectively.  The controller's firing
signal and bad output are
\[
 \mathit{pop\_fire}=\mathit{pop}\land o_0,\qquad
 \mathit{bad}=\mathit{pop\_fire}\land\neg o_0.
\]
Here $\mathit{pop\_fire}$ is the controller guard, rather than the pop-only
state-update enable $R$; a simultaneous push and pop may therefore raise the
guard while leaving the occupancy unchanged.  The guard includes $o_0$, so
the bad expression is identically false.

\paragraph{C3.3.3 Round-robin mutual exclusion.}
Sample $N$ and $H$ uniformly from $\{6,7\}$ and $\{20,21,22\}$,
respectively.  The token latches are
initialized to $(1,0,\ldots,0)$.  With public requests $r_j$ and an
unconstrained input $a=\mathit{advance}$, set
\[
 t'_j=\begin{cases}t_{(j-1)\bmod N},&a=1,\\t_j,&a=0,
 \end{cases}
 \qquad g_j=r_j\land t_j.
\]
The bad output is $\bigvee_{j<k}(g_j\land g_k)$.  Rotation and hold both
preserve the one-hot token invariant, hence at most one grant can be true in
every frame.

\paragraph{C3.3.4 Round-robin bounded fairness.}
Sample $N$ and $H$ uniformly from $\{7,8\}$ and $\{22,23,24\}$,
respectively, and use the same one-hot token,
requests, grants, and nondeterministic advance input as in C3.3.3.  Introduce
$N$ zero-initialized wait latches and define
\[
 c=r_0\land a\land\neg g_0,\qquad
 w'_0=c,\qquad w'_j=c\land w_{j-1}\ (1\leq j<N),\qquad
 \mathit{bad}=c\land w_{N-1}.
\]
The wait vector records consecutive clock frames in which $c$ has remained
true, so bad would require $N+1$ such frames.  In fact, no run of $N$
consecutive $c$-frames exists: each one advances the one-hot token, which must
visit client zero within $N$ advances and then makes $g_0$ true.  This monitor
concerns consecutive active clock frames; it does not impose progress across
idle frames or assert an unbounded fairness condition.

\paragraph{C3.3.5 Credit-router conservation.}
Sample capacity $C$ and horizon $H$ uniformly from $\{6,7\}$ and
$\{22,23,24\}$, respectively.  Two unary vectors
$c_0,\ldots,c_{C-1}$ and $f_0,\ldots,f_{C-1}$ represent available credits and
in-flight flits: with $k$ flits in flight, they have forms
$c=1^{C-k}0^k$ and $f=1^k0^{C-k}$.  Initially $c_j=1$ and $f_j=0$.  Define
\[
 S=\mathit{send}\land c_0,\qquad
 R=\mathit{credit\_return}\land f_0,\qquad
 S_o=S\land\neg R,\quad R_o=R\land\neg S.
\]
On $S_o$, set
$c'_j=c_{j+1}$ for $j<C-1$, $c'_{C-1}=0$, $f'_0=1$, and
$f'_j=f_{j-1}$ for $j>0$.  On $R_o$, set
$c'_0=1$, $c'_j=c_{j-1}$ for $j>0$, $f'_j=f_{j+1}$ for $j<C-1$, and
$f'_{C-1}=0$; otherwise hold both vectors.  In particular, when both $S$ and
$R$ are true, the two events cancel and the state is held.  The bad output is
\[
 \bigvee_{j=0}^{C-1}\bigl(c_j\oplus\neg f_{C-1-j}\bigr).
\]
The complementary mirror relation $c_j=\neg f_{C-1-j}$ holds initially and is
preserved by every branch, expressing exact credit conservation.

\paragraph{C3.3.6 Credit-router forwarding safety.}
Sample $C$ and $H$ uniformly from $\{12,13,14\}$ and
$\{36,37,38,39,40\}$, respectively.  Construct two disjoint latch families
implementing the credit/flight
tracker from C3.3.5, named actual and shadow, with identical initial values and
driven by the same \textit{send} and \textit{credit-return} inputs.  Let
\[
 \begin{aligned}
 \mathit{actual\_forward}
   &=\mathit{send}\land c^{\mathrm{actual}}_0,\\
 \mathit{shadow\_forward}
   &=\mathit{send}\land c^{\mathrm{shadow}}_0,\\
 \mathit{bad}
   &=\mathit{shadow\_forward}\land\neg\mathit{actual\_forward}
     =\mathit{shadow\_forward}\land\neg c^{\mathrm{actual}}_0.
 \end{aligned}
\]
Thus bad is the one-directional mismatch in which the shadow tracker permits
forwarding but the actual tracker does not.  The two latch families obey
identical deterministic recurrences, so equality of corresponding latches is
inductive and excludes this mismatch.

\paragraph{C3.3.7 Pipeline flush-stall safety.}
Sample the number of stages $S$ and horizon $H$ uniformly from
$\{9,10,11\}$ and $\{30,31,32,33,34\}$, respectively.  The circuit has valid
latches $v_j$, a pending-flush latch $p$, and an $S$-bit progress register
$w_j$, all initialized to zero, with
inputs \textit{issue}, \textit{stall}, and \textit{flush}.  Write
\[
 a=p\lor\mathit{flush},\qquad
 d=a\land\neg\mathit{stall},\qquad q=w_{S-1}.
\]
The pending latch uses set-dominant priority,
\[
 p'=\begin{cases}
 1,&\mathit{flush},\\
 0,&\neg\mathit{flush}\land q,\\
 p,&\text{otherwise}.
 \end{cases}
\]
The valid latches obey
\[
 v'_j=\begin{cases}
 v_j,&\mathit{stall},\\
 \mathit{issue}\land\neg a,&\neg\mathit{stall}\land j=0,\\
 v_{j-1},&\neg\mathit{stall}\land j>0.
 \end{cases}
\]
Thus no new valid entry is inserted while a flush is active.  The
progress recurrence uses
$\sigma_0=1$ and $\sigma_j=w_{j-1}$ for $1\leq j<S$, and is
\[
 w'_j=
 \begin{cases}
 [j=0],&\mathit{flush}\land\neg\mathit{stall},\\
 0,&\mathit{flush}\land\mathit{stall},\\
 \sigma_j,&\neg\mathit{flush}\land d,\\
 w_j,&\neg\mathit{flush}\land\neg d\land p,\\
 0,&\text{otherwise},
 \end{cases}
\]
where $[j=0]$ denotes the Boolean indicator of $j=0$.  The bad output is
$q\land\bigvee_j v_j$.  Whenever $q$ is true, the active flush has accumulated
$S$ nonstalling shifts since the progress register was reset; no valid entry
was inserted during those shifts, so the pipeline is empty.

\paragraph{C3.3.8 Pipeline writeback consistency.}
Sample $S$ and $H$ uniformly from $\{14,15,16\}$ and
$\{44,45,46,47,48\}$, respectively.  Three zero-initialized length-$S$
pipelines store valid bits $v_j$, one-bit tags $t_j$, and independently
represented one-bit shadow tags $\widehat t_j$.  The control priority is:
flush first, stall second, and shift otherwise.  A flush sets all three
next-state vectors to zero, a stall holds them, and a shift sets
\[
 v'_0=\mathit{issue},\qquad
 t'_0=\widehat t'_0=\mathit{input\_tag},\qquad
 v'_j=v_{j-1},\quad t'_j=t_{j-1},\quad
 \widehat t'_j=\widehat t_{j-1}\quad(1\leq j<S).
\]
The bad output is
$v_{S-1}\land(t_{S-1}\oplus\widehat t_{S-1})$.  Corresponding tag and shadow
latches are equal initially and receive equal updates in every control branch,
so an enabled writeback cannot contain mismatched tags.

\clearpage
\section{Tool Use and Problem Complexity}
\label{app:tool-use}

\subsection{Auxiliary polynomial-time tasks}
The auxiliary Tier-0 contains 18 cases: the six Max-Flow / Min-Cut cases
from C1.1, six 2-SAT cases, and six bipartite minimum vertex-cover cases.
The latter two tasks correspond to restricted forms of the satisfiability
and set-cover problems in Tier-1. They use a single deterministic certificate
submission. We give their task-level definitions below; no additional
case-specific construction is introduced here.

\subsubsection{Max-Flow / Min-Cut}
The public input is a capacitated directed multigraph $G=(V,E)$, nonnegative
integer capacities $(c_e)_{e\in E}$ and distinct terminals $s,t$.
Parallel edge occurrences have distinct identifiers. Merlin submits a flow
$(f_e)_{e\in E}$, its claimed value $F$, and a source-side cut $S\subseteq V$.
Arthur checks $s\in S$, $t\notin S$, capacity constraints
$0\leq f_e\leq c_e$, conservation at every nonterminal vertex, and
\[
 \sum_{e:u_e=s}f_e-\sum_{e:v_e=s}f_e
 =F
 =\sum_{e:v_e=t}f_e-\sum_{e:u_e=t}f_e
 =\sum_{e:u_e\in S,\,v_e\notin S}c_e.
\]
Feasibility gives a lower bound and the cut gives the same upper bound, so
acceptance certifies both a maximum flow and a minimum cut. The task and six
cases are the same as C1.1.
\subsubsection{2-SAT}
The public instance is a Boolean formula in 2-CNF,
\[
  \varphi(x_1,\ldots,x_n)
  =\bigwedge_{j=1}^{m} C_j,
  \qquad
  C_j=\ell_{j,1}\lor\ell_{j,2},
  \qquad
  \ell_{j,k}\in\{x_i,\neg x_i:i\in[n]\}.
\]
Every clause contains exactly two literal entries.  The task is a decision
problem rather than a weighted optimization problem: all clauses are hard and
there are no soft-clause weights.

Let \(D_\varphi=(V_\varphi,A_\varphi)\) denote the implication graph, where
\(
  V_\varphi=\{x_i,\neg x_i:i\in[n]\}
\)
and each clause \(\ell_{j,1}\lor\ell_{j,2}\) contributes the two arcs
\(
  \overline{\ell_{j,1}}\longrightarrow\ell_{j,2},
  \overline{\ell_{j,2}}\longrightarrow\ell_{j,1}.
\)
Here \(\overline{\ell}\) denotes the complementary literal.  Write
\(\mathsf{Path}_{D_\varphi}(u,v)\) for the set of directed paths from \(u\)
to \(v\) in this graph.

Merlin submits one of two tagged certificates.  A SAT certificate contains a
complete Boolean assignment \(a\in\{0,1\}^n\).  An UNSAT certificate contains
an index \(i\in[n]\) and two explicit implication paths \(P^+\) and \(P^-\).
Arthur accepts exactly when
\[
\begin{aligned}
  &\bigl[
    \pi=(\mathsf{SAT},a)
    \ \land\ 
    a\in\{0,1\}^n
    \ \land\
    \bigwedge_{j=1}^{m}C_j(a)=1
  \bigr]
  \\
  {}\lor{}&
  \bigl[
    \pi=(\mathsf{UNSAT},i,P^+,P^-)
    \ \land\
    i\in[n]
    \ \land\
    P^+\in\mathsf{Path}_{D_\varphi}(x_i,\neg x_i)
    \ \land\
    P^-\in\mathsf{Path}_{D_\varphi}(\neg x_i,x_i)
  \bigr].
\end{aligned}
\]
For a SAT certificate, the verifier evaluates every public clause directly.
For an UNSAT certificate, it checks the endpoints of both paths and verifies
that every consecutive pair of literals is an implication induced by a
public clause.  The two paths place \(x_i\) and \(\neg x_i\) in the same
strongly connected component and therefore certify unsatisfiability.
Conversely, every unsatisfiable 2-CNF formula has such a complementary pair.

The current generator supplies both kinds of instances: four of the six case
profiles are satisfiable and two are unsatisfiable.  A certificate can be
constructed in polynomial, indeed linear, time using the implication graph.
Arthur nevertheless checks only the submitted assignment or paths and does
not compare the claimed status with a hidden reference solution.

\subsubsection{Bipartite Graph Minimum Vertex Cover}
Let
\[
  G=(L\mathbin{\dot\cup}R,E)
\]
be a finite simple bipartite graph whose bipartition is public.  Vertices have
opaque identifiers, and every public edge has a distinct opaque edge
identifier.  A vertex cover is a set \(S\subseteq L\mathbin{\dot\cup}R\)
such that
\[
  \forall\,\{u,v\}\in E:\qquad u\in S\ \lor\ v\in S.
\]
A matching is a set \(M\subseteq E\) whose edges have pairwise disjoint
endpoints.

Merlin submits a pair \((S,M)\), represented by distinct side-tagged vertex
identifiers for \(S\) and distinct public edge identifiers for \(M\).  Arthur
accepts exactly when
\[
\begin{aligned}
&S\subseteq L\mathbin{\dot\cup}R,
  \qquad M\subseteq E,
  \\
  &\forall\,\{u,v\}\in E:\quad u\in S\lor v\in S,
  \\
  &\forall\,e,f\in M,\ e\neq f:\quad e\cap f=\varnothing,
  \\
  &|S|=|M|.
\end{aligned}
\]
The concrete verifier additionally checks that all identifiers are valid,
that every side tag is correct, and that neither list contains duplicates.

By Kőnig's theorem, the minimum vertex-cover cardinality of a bipartite
graph equals its maximum-matching cardinality.  Therefore, a valid vertex
cover \(S\) and a valid matching \(M\) satisfying \(|S|=|M|\) certify that
\(S\) is a minimum vertex cover.  Arthur checks only coverage, matching
validity, and cardinality equality; it does not compute an optimum during
verification.

The current generator constructs balanced graphs with
\(|L|=|R|=q\) and plants a perfect matching.  Since either complete side is
a vertex cover of size \(q\), every generated instance satisfies
\[
  \nu(G)=\tau(G)=q.
\]
This planted structure determines the instance family, but acceptance still
depends solely on the submitted primal--dual certificate.

\subsection{Evaluation with and without tools}
The tool-enabled auxiliary experiment evaluates Luna, Terra and Sol on the
18 Tier-0 cases. The no-tools experiment evaluates Terra and Sol on three
Tier-1 families: Max-Flow / Min-Cut, Weighted Partial MaxSAT, and Structured
Set Cover, with six cases in each family. These are distinct experimental
sets: 2-SAT and bipartite vertex cover belong to the auxiliary set, while
the no-tools comparison uses the original Tier-1 satisfiability and
set-cover tasks.

In the no-tools condition, the model receives the public specification,
protocol and full instance as text. It cannot call a shell, read a file,
execute code, search the web, or use another agent. The trusted host parses
the final JSON object and forwards it to Arthur without changing its
contents. The relay fixes sequence zero and empty artifact references.
There is no repair pass or retry after the model's submission.

Large inputs may be split into consecutive text messages within the same
fresh session. Each nonfinal input message requests only an acknowledgment;
after all parts have arrived, a final message requests the certificate.
The input and acknowledgment turns remain within the common 600-second
budget. A model-side event-stream audit checks that no tool was called.

\begin{table}[htbp]
\centering\small
\caption{Task-specific fields in the no-tools submission prompt.}
\label{tab:no-tools-fields}
\begin{tabularx}{\linewidth}{@{}l>{\raggedright\arraybackslash}X>{\raggedright\arraybackslash}X@{}}
\toprule
Task & Required JSON fields & Event kind \\
\midrule
Max-Flow / Min-Cut & \code{flow_value}, \code{edge_flows}, \code{source_side_cut} & \code{maxflow.certificate} \\
Weighted Partial MaxSAT & \code{assignment} & \code{maxsat.assignment} \\
Structured Set Cover & \code{set_ids} & \code{set_cover.selection} \\
\bottomrule
\end{tabularx}
\end{table}

\promptfile{pr:no-tools}{Direct no-tools task message}{prompts/publication/no_tools_direct.txt}
\promptfile{pr:no-tools-parts}{Multipart input and final-answer messages}{prompts/publication/no_tools_multipart.txt}

\clearpage
\section{Prompts for Different Levels of User Expertise}
\label{app:user-expertise}

\subsection{Conditions and common instructions}
The user-prompt experiment fixes the evaluated model to GPT-5.6-Terra and
compares three kinds of assistance on all 40 Tier-2 cases. Engineer provides
execution guidance; Student provides a task-family method; Expert adds
knowledge of the instance construction and an associated strategy.
None of these prompts supplies the required witness, proof messages or
Arthur's challenge answers.

The instructions are organized as a common base followed by the selected
role's advice. Let $P_0$ denote the base task prompt, $R$ the common resource
instructions, $E$ the Engineer prompt, $S_f$ the Student prompt for family
$f$, and $X_{f,c}$ the Expert addition for case $c$. Concatenation gives
\[
 \begin{aligned}
 P_{\rm Engineer}&=P_0\Vert R\Vert E,\\
 P_{\rm Student}(f)&=P_0\Vert R\Vert S_f,\\
 P_{\rm Expert}(f,c)&=P_0\Vert R\Vert S_f\Vert X_{f,c}.
 \end{aligned}
\]
Thus the Student advice printed below is also part of the corresponding
Expert prompt. The common instructions and family blocks are printed once
to avoid repetition.

\promptfile{pr:role-common}{Common execution constraints}{prompts/publication/role_controls.txt}

\subsection{Engineer: execution guidance}
Engineer asks the model to maintain a durable protocol state, validate
messages before sending, and reserve time for verification and cleanup.
The advice concerns execution rather than the mathematical solution.
\promptfile{pr:engineer}{Consolidated Engineer instructions}{prompts/publication/engineer.txt}

\subsection{Student: task-family guidance}
The following six prompts provide general methods for the corresponding
families without identifying the case-specific construction.
\shortpromptfile{pr:student-gni}{Graph Non-Isomorphism}{prompts/publication/student/gni.txt}

\shortpromptfile{pr:student-sumcheck}{Sum-Check}{prompts/publication/student/sumcheck.txt}

\shortpromptfile{pr:student-gkr}{GKR}{prompts/publication/student/gkr.txt}

\shortpromptfile{pr:student-image}{Circuit Image Size}{prompts/publication/student/image.txt}

\shortpromptfile{pr:student-tqbf}{True Quantified Boolean Formulas}{prompts/publication/student/tqbf.txt}

\shortpromptfile{pr:student-hamilton}{Hamiltonicity}{prompts/publication/student/hamilton.txt}

\subsection{Expert: case-specific guidance}
For each case below, the Expert prompt appends the displayed addition after
the corresponding Student prompt. Case identifiers follow the task
descriptions. The advice explains the construction and a strategy, while
leaving the model to recover the structure from the public instance,
implement the method and complete the proof protocol.
\Needspace{13\baselineskip}
\subsubsection{Graph Non-Isomorphism}

\shortpromptfile{pr:expert-gni-c01}{Case C2.1.1}{prompts/publication/expert/gni_c01.txt}

\shortpromptfile{pr:expert-gni-c02}{Case C2.1.2}{prompts/publication/expert/gni_c02.txt}

\shortpromptfile{pr:expert-gni-c03}{Case C2.1.3}{prompts/publication/expert/gni_c03.txt}

\shortpromptfile{pr:expert-gni-c04}{Case C2.1.4}{prompts/publication/expert/gni_c04.txt}

\shortpromptfile{pr:expert-gni-c05}{Case C2.1.5}{prompts/publication/expert/gni_c05.txt}

\shortpromptfile{pr:expert-gni-c06}{Case C2.1.6}{prompts/publication/expert/gni_c06.txt}

\shortpromptfile{pr:expert-gni-c07}{Case C2.1.7}{prompts/publication/expert/gni_c07.txt}

\Needspace{13\baselineskip}
\subsubsection{Sum-Check}

\shortpromptfile{pr:expert-sumcheck-c01}{Case C2.2.1}{prompts/publication/expert/sumcheck_c01.txt}

\shortpromptfile{pr:expert-sumcheck-c02}{Case C2.2.2}{prompts/publication/expert/sumcheck_c02.txt}

\shortpromptfile{pr:expert-sumcheck-c03}{Case C2.2.3}{prompts/publication/expert/sumcheck_c03.txt}

\shortpromptfile{pr:expert-sumcheck-c04}{Case C2.2.4}{prompts/publication/expert/sumcheck_c04.txt}

\shortpromptfile{pr:expert-sumcheck-c05}{Case C2.2.5}{prompts/publication/expert/sumcheck_c05.txt}

\shortpromptfile{pr:expert-sumcheck-c06}{Case C2.2.6}{prompts/publication/expert/sumcheck_c06.txt}

\shortpromptfile{pr:expert-sumcheck-c07}{Case C2.2.7}{prompts/publication/expert/sumcheck_c07.txt}

\Needspace{13\baselineskip}
\subsubsection{GKR}

\shortpromptfile{pr:expert-gkr-c01}{Case C2.3.1}{prompts/publication/expert/gkr_c01.txt}

\shortpromptfile{pr:expert-gkr-c02}{Case C2.3.2}{prompts/publication/expert/gkr_c02.txt}

\shortpromptfile{pr:expert-gkr-c03}{Case C2.3.3}{prompts/publication/expert/gkr_c03.txt}

\shortpromptfile{pr:expert-gkr-c04}{Case C2.3.4}{prompts/publication/expert/gkr_c04.txt}

\shortpromptfile{pr:expert-gkr-c05}{Case C2.3.5}{prompts/publication/expert/gkr_c05.txt}

\shortpromptfile{pr:expert-gkr-c06}{Case C2.3.6}{prompts/publication/expert/gkr_c06.txt}

\shortpromptfile{pr:expert-gkr-c07}{Case C2.3.7}{prompts/publication/expert/gkr_c07.txt}

\Needspace{13\baselineskip}
\subsubsection{Circuit Image Size}

\shortpromptfile{pr:expert-image-c01}{Case C2.4.1}{prompts/publication/expert/image_c01.txt}

\shortpromptfile{pr:expert-image-c02}{Case C2.4.2}{prompts/publication/expert/image_c02.txt}

\shortpromptfile{pr:expert-image-c03}{Case C2.4.3}{prompts/publication/expert/image_c03.txt}

\shortpromptfile{pr:expert-image-c04}{Case C2.4.4}{prompts/publication/expert/image_c04.txt}

\shortpromptfile{pr:expert-image-c05}{Case C2.4.5}{prompts/publication/expert/image_c05.txt}

\shortpromptfile{pr:expert-image-c06}{Case C2.4.6}{prompts/publication/expert/image_c06.txt}

\Needspace{13\baselineskip}
\subsubsection{True Quantified Boolean Formulas}

\shortpromptfile{pr:expert-tqbf-c01}{Case C2.5.1}{prompts/publication/expert/tqbf_c01.txt}

\shortpromptfile{pr:expert-tqbf-c02}{Case C2.5.2}{prompts/publication/expert/tqbf_c02.txt}

\shortpromptfile{pr:expert-tqbf-c03}{Case C2.5.3}{prompts/publication/expert/tqbf_c03.txt}

\shortpromptfile{pr:expert-tqbf-c04}{Case C2.5.4}{prompts/publication/expert/tqbf_c04.txt}

\shortpromptfile{pr:expert-tqbf-c05}{Case C2.5.5}{prompts/publication/expert/tqbf_c05.txt}

\shortpromptfile{pr:expert-tqbf-c06}{Case C2.5.6}{prompts/publication/expert/tqbf_c06.txt}

\Needspace{13\baselineskip}
\subsubsection{Hamiltonicity}

\shortpromptfile{pr:expert-hamilton-c01}{Case C2.6.1}{prompts/publication/expert/hamilton_c01.txt}

\shortpromptfile{pr:expert-hamilton-c02}{Case C2.6.2}{prompts/publication/expert/hamilton_c02.txt}

\shortpromptfile{pr:expert-hamilton-c03}{Case C2.6.3}{prompts/publication/expert/hamilton_c03.txt}

\shortpromptfile{pr:expert-hamilton-c04}{Case C2.6.4}{prompts/publication/expert/hamilton_c04.txt}

\shortpromptfile{pr:expert-hamilton-c05}{Case C2.6.5}{prompts/publication/expert/hamilton_c05.txt}

\shortpromptfile{pr:expert-hamilton-c06}{Case C2.6.6}{prompts/publication/expert/hamilton_c06.txt}

\shortpromptfile{pr:expert-hamilton-c07}{Case C2.6.7}{prompts/publication/expert/hamilton_c07.txt}

\clearpage
\section{Experience Transfer to Native Verification Tasks}
\label{app:transfer}

\subsection{Teacher and learner procedure}
The transfer experiment asks whether experience from synthetic Tier-1 and
Tier-2 tasks can help a fresh model construct evidence for Tier-3 tasks.
A teacher reads source execution records and writes a short English lesson.
The learner studies that lesson before receiving the downstream problem.
The downstream tasks are machine-checked mathematics in Lean, program
verification in Dafny, and bounded hardware safety with DRAT certificates.
Each family contains eight cases.

The teacher is instructed to inspect both successful and failed records,
extract source-supported operational advice, and keep source citations and
coverage information outside the short learner-facing entry. The lesson
must not contain downstream answers. The two delivered entries contain 621
and 598 words, respectively, and are reproduced below.

For every downstream case, a fresh learner receives the lesson during a
300-second study phase. The task is then released with an independent
600-second evaluation budget. Study is ungraded: there is no quiz, required
summary or coverage score. The full lesson package remains available during
evaluation. The learner chooses how to apply it, and the current task's
contract determines acceptance.

The evaluation condition prohibits acquiring or reusing third-party
specialized solving software. Standard runtimes, compilers, standard
libraries and the provided client remain available. The learner may write
its own construction and checking code and consult public documentation
read-only. Arthur invokes the task's native checker on the submitted
artifact. These restrictions are stated in the evaluation message, rather
than inferred from the teacher's general advice about tool use.

\subsection{Qualitative Analysis of a Zero-Trace Control}

We recognize that an important control for separating the contribution of source-trace experience from general guidance and additional preparation is to give the learner a guide generated without source traces, while matching the 300-second study phase, guide length, and downstream evaluation conditions. To examine this control, we asked GPT-5.6-Terra and GPT-5.6-Sol to generate a guide under the same guide-generation requirements but \textbf{without any source traces}, as provided in \ref{app:zero_trace_guide_sol} and \ref{app:zero_trace_guide_terra}. 

The resulting guide explicitly states that, in the absence of observable source records, its recommendations are extrapolations. Its content primarily concerns structured problem solving, tool selection, candidate validation, state preservation, and time management. These recommendations may have practical value, but they provide no evidence from actual source-task executions about which strategies succeeded or failed under particular conditions. For example, the guide illustrates its advice using complementary labels at symmetric positions and explicitly acknowledges that this example does not originate from a source record. The guide is therefore limited in the traceable source-task experience it provides. This assessment concerns the provenance and specificity of its experiential content, rather than its downstream effectiveness.

Some recommendations also illustrate operational heuristics that warrant caution without support from source records, including performing two independent checks, constructing valid and invalid miniature examples for a newly written checker, and reserving approximately one fifth of the time budget. Mechanically applying these recommendations could introduce redundant work or reduce productive solving time when existing validation already covers the relevant risks or submission costs are low. 

\subsection{Transfer prompts}
The teacher prompt fixes the source boundary and the format of the portable
lesson. The study prompt is delivered before the learner sees its target
task. The evaluation-release message and task prelude state the evaluation
boundary and point to the public task instructions. Uppercase braced fields
are filled with the corresponding model and workspace paths.

\promptfile{pr:teacher}{Teacher: extract a portable lesson}{prompts/publication/transfer_teacher.txt}
\promptfile{pr:study}{Learner: study before task release}{prompts/publication/transfer_study.txt}
\promptfile{pr:release}{Learner: release the downstream task}{prompts/publication/transfer_evaluation_release.txt}
\promptfile{pr:transfer-prelude}{Prelude to the downstream task instructions}{prompts/publication/transfer_task_prelude.txt}

\subsection{Lesson written by Sol}
The following is the complete learner-facing entry, with its headings
typeset as paragraphs.
The current contract defines success: construction, required encoding, and terminal acceptance may be separate obligations. Choose a route that can deliver all three within its time and resource limits.

\paragraph{Construct or search}

When the visible instance exposes independent pieces, sparse pivots, a distinguishing invariant, or a small interface between pieces, seek a complete witness through that structure before expanding global search. The useful intermediate result is an explicit correspondence or boundary state that accounts for every constraint, not merely a suggestive pattern. For example, on disjoint paths with allowed labels, retain feasible labels at the current endpoint and predecessor choices; extend only through compatible adjacent labels, then reconstruct each path. This turns local feasibility into an actual assignment. Check that no edges connect the supposedly independent pieces. Stop pursuing the reduction if unexplained constraints remain or the boundary state grows beyond a practical size; a smaller search over unresolved choices may then be preferable.

When no cheap construction is apparent, an already permitted solver or numerical capability can be useful if its representation fits the remaining constraints. Seek a complete feasible candidate, not merely a promising objective value or progress counter. Use approximations to propose candidates only when exact validation can establish the required property. If a capability is unavailable, acquiring it is conditional on actual permissions and credible setup savings; do not assume acquisition is necessary. Prefer direct construction when setup would consume the time needed to finish it. Stop setup or search when it no longer offers a plausible path to a qualifying, exportable candidate within the remaining budget.

\paragraph{Check the uncertainty}

When a new construction or checker has a specific uncertainty, select a small independent check aimed at that uncertainty: recompute the required constraints and cost from the candidate, compare a classifier with a different invariant implementation, or test the first transition between two kinds of state. For a newly written checker, tiny known-valid and known-invalid cases may expose checker errors before blaming the construction, when such cases are readily available and useful. Carry fixed contributions across boundaries; a correct recurrence inside each region can still fail at their junction. Seek agreement on the property that acceptance depends on. Once that uncertainty is resolved, stop adding optional checks unless new evidence warrants them. Complete every validation required by the contract; a successful local rehearsal does not replace live acceptance.

\paragraph{Preserve usable state}

When a complete candidate first satisfies the acceptance conditions, preserve the full object before further improvement. If only a threshold is required, move toward packaging instead of chasing an unnecessary optimum. For an interactive exchange, retain the entire returned response together with the outgoing message, session identity, sequence, and continuation state before changing clients or advancing. Parse the saved response, not a shortened display. If acknowledgment is uncertain, use only an allowed state-recovery mechanism; do not replay or guess when an accepted message or its challenge may have been lost. For asynchronous results, also retain running status, the continuation handle, and final return code. Resume or inspect existing work before launching it again; a shortened output string does not establish completion.

\paragraph{Finish the exchange}

When setup, computation, or optional rehearsal threatens delivery time, estimate the remaining serialization and required exchanges, including their observed latency. Seek a schedule that still reaches the terminal result, not just the first submission. As a heuristic, reserve a meaningful share of the budget for packaging, required checks, and completion, enlarging it for long serial exchanges; no fixed fraction is established by these records. Stop optional refinement when that reserve is reached and deliver a qualifying candidate through the required procedure. If none exists, preserve useful work and report the shortfall rather than submit a known-invalid object or call partial progress success.

\subsection{Lesson written by Terra}
The following is the complete learner-facing entry; the title is represented
by this subsection heading.
\paragraph{Construct locally}

Treat a verifier-backed task as a request for an accepted object, not merely a plausible algorithm. Begin by looking for a small correspondence that makes the witness local: an invariant separating alternatives, a boundary condition that fixes a recurrence, or a factorization that turns a global object into independent pieces. The trigger is that the public structure repeats, has small local arity, or exposes a distinguishable signature. Seek a construction whose every submitted component is generated from the same checked representation; stop this route when the proposed correspondence is ambiguous on even one required lane or when its local identity fails. In the retained records, a table on two binary coordinates was factored into two unary contributions, then each variable's small polynomial was built from those contributions; the recorded successful artifact checked the first boundary identity before continuing. That is an example of structure, not a claim that factorization will exist elsewhere.

\paragraph{Choose mechanics}

When the task already permits local code or a supplied client, use it only when it reduces the dominant uncertainty--such as exact enumeration of a small domain, deterministic serialization, or preservation of a multi-step state. The useful next result is a candidate plus the state needed to reproduce or resume it. Setup is not free: if a direct constructive derivation is shorter than preparing unfamiliar machinery, prefer the direct route; stop investing in setup when it cannot produce a checked candidate soon enough for the task's terminal exchange. The archived records include compact exact enumerators and state/payload files; other records reached their time limit without terminal acceptance. This is a conditional tradeoff, not an instruction to acquire tools.

\paragraph{Check proportionately}

Use small independent checks that target the verifier's likely rejection surface. Before sending, if the actual candidate or interface makes a condition relevant, check the corresponding local boundary identity, count or coverage threshold, item correspondence, or message shape and sequence. When one invariant drives many answers and a cheap differently organized calculation is available, use it to catch a shared orientation, indexing, or encoding error; do not require a second implementation otherwise. If a newly written checker will determine whether to submit, optional tiny known-valid and known-invalid controls can establish that it distinguishes the intended cases; stop when that confidence is sufficient for the live candidate. Seek agreement between the candidate, its serialized form, and the current response state; stop checking once the applicable acceptance conditions are established, rather than turning validation into a new project. A recorded signature-based response was rejected; the conditional separation check remains a preventive recommendation, not an attributed diagnosis of that rejection.

\paragraph{Preserve the handoff}

Preserve the complete candidate early: the generated witness, parsed challenge, accepted sequence number, and enough provenance to rebuild the exact payload. If the current interface exposes a raw response, asynchronous status, continuation handle, or return code, retain it with that state; this is a bounded preventive recommendation, not a requirement to create new tracking. The trigger is any stateful or sequenced interface, or a witness expensive to recompute. The useful next result is safe continuation or a ready final package; stop exploring and package once a complete, checked candidate exists. As a heuristic--not an observed universal law--reserve a visible final slice of the budget for serialization, the required terminal round trip, and recovery from one ordinary formatting error. Some clean observations reached the time limit without terminal acceptance, whereas accepted records reached a terminal acknowledgement. Never treat local program output as acceptance: the current contract's required validation and terminal result remain decisive.

\subsection{Lesson written by Sol with zero trace}
\label{app:zero_trace_guide_sol}
Evidence status: no observable source records were present, so every decision below is a recommended extrapolation rather than a lesson demonstrated by the archive. The task contract remains controlling, and none of these routes should replace its required validation or output format.

Begin by identifying the smallest fact that would turn the requested object from a search problem into a construction. The trigger is a constraint that repeats, a boundary case that fixes neighboring choices, a correspondence between two representations, or a subproblem that can be reused without modification. Seek one locally checkable invariant and use it to generate a complete candidate, not merely fragments. Stop this route when the invariant no longer determines legal extensions, when its proof costs more than direct enumeration, or when it conflicts with an explicit requirement. A generic mathematical example is a symmetric structure whose paired positions must have complementary labels: fixing one representative per pair can reduce the choices while preserving a simple pairwise check. This example is illustrative only; it is not source-native because no retained source artifact was available.

Consider an already permitted capability only when the next uncertainty is genuinely cheaper for it to resolve. The trigger is a repetitive calculation, a finite search with a crisp predicate, or a format transformation that is tedious but mechanically testable. Seek either a valid candidate or a decisive counterexample to the current idea. Stop if setup, acquisition, or adaptation consumes the time needed to construct and package the answer; then prefer a direct construction. Never assume that a capability may be acquired: actual permissions and the current contract decide. Output from an unverified procedure is evidence to inspect, not acceptance.

Once a candidate exists, apply two small independent checks that target different failure modes. The trigger is the first complete candidate or a change that could invalidate an earlier one. Seek, first, a structural check of the governing constraints and, second, a representation check such as size, ordering, boundary values, or round-trip interpretation. If a checker is written, test it on one deliberately valid miniature and one deliberately broken miniature before trusting it. Stop checking when required validation passes and the remaining checks merely restate the same predicate. Optional checks should protect acceptance, not create a new objective or displace the deliverable.

Preserve continuity as soon as a complete candidate appears. The trigger is any candidate that might pass even if a better route remains open. Save the entire object together with the response state needed to resume: assumptions, unresolved risks, and the exact validation status. Seek a recoverable baseline from which local improvements can be made without reconstructing prior work. Stop exploratory edits when they endanger that baseline or leave insufficient time for required serialization, packaging, and terminal acceptance. Never submit a known-invalid object merely to preserve activity.

As a heuristic, reserve roughly the final fifth of the available budget for assembling the exact response, rerunning required checks, and obtaining the terminal acceptance signal. This proportion is not an observed universal law and should shrink or grow with packaging complexity. The trigger to switch is either a validated complete candidate or arrival at that reserve. Seek a contract-compliant submission with traceable validation. Stop all optional optimization once it threatens that outcome. If no valid candidate exists at the switch point, spend the reserve repairing the closest complete construction rather than beginning an unrelated search, unless the contract itself makes that route impossible.

\subsection{Lesson written by Terra with zero trace}
\label{app:zero_trace_guide_terra}

No source records, retained artifacts, or manifest rows were inspected for this control entry. Consequently, none of the following is presented as an observed lesson, a proven transfer, or a substitute for the current task contract. It is a compact operating hypothesis for a later solver when its own permitted evidence supports the choice.

If the task exposes a small invariant, boundary condition, correspondence, or independently reusable subproblem, first ask whether it fixes most of the witness while leaving only a local choice. Seek a complete candidate obtained by extending that structure, rather than beginning a broad search. For example, when objects can be paired by a conserved relation, construct and retain those pairs before choosing the remaining compatible links. Stop this route when the relation does not determine a valid continuation, when its residual cases remain global, or when checking the extension is more expensive than an allowed direct method. This is a recommendation, not a recorded action.

If a capability is already permitted and immediately reduces a clearly identified bottleneck, use it only after estimating whether its setup can pay back within the remaining budget. Seek a small intermediate result that directly advances the candidate or acceptance package, not an open-ended exploration. A direct construction is often the more viable route when configuration, input conversion, or learning the interface would consume the time needed to finish and preserve a witness. Stop the capability route if its first useful output requires substantial additional setup, if its permissions do not cover the actual task, or if it cannot produce evidence that the checker accepts. Tool acquisition is conditional on the later task's permissions and is not assumed here.

Once a candidate exists, run the smallest independent checks that target the most failure-prone obligations: coverage of required elements, uniqueness or consistency of assignments, boundary cases, and the exact shape of the deliverable. Seek a concrete discrepancy or a concise confirmation tied to the stated contract. Stop adding optional checks when the required conditions are covered, the candidate is internally coherent, and another check would displace packaging or terminal acceptance. An unverified program result is not success; where validation is required, keep it required. Conversely, optional checking should aid acceptance rather than become a separate score to optimize.

Maintain one complete, recoverable representation of the candidate and the response state needed to continue: assumptions still in force, decided components, unresolved components, and the format expected at submission. After each meaningful advance, preserve the full object rather than only a summary of why it might work. Seek a final pass that converts that object into the required response without omissions. Stop revising once the object has passed the necessary checks and sufficient time remains for required packaging and terminal acceptance.

A budget split can be used only as a heuristic: reserve a visible final portion for validation, formatting, and submission, then spend the earlier portion on the route that yields the next checkable construction most quickly. Reallocate when a route produces no such object. This is not an observed universal law, and the current task's contract overrides every suggestion above.


\end{document}